\documentclass[13pt]{article}

\usepackage{latex/PRIMEarxiv}

\usepackage[utf8]{inputenc} 
\usepackage[T1]{fontenc}    
\usepackage{hyperref}       
\hypersetup{
    colorlinks=true, 
    linkcolor=blue,  
    citecolor=blue,  
    urlcolor=blue    
}
\usepackage{url}            
\usepackage{booktabs}       
\usepackage{amsfonts}       
\usepackage{amsmath}
\usepackage{nicefrac}       
\usepackage{microtype}      
\usepackage{lipsum}
\usepackage{fancyhdr}       
\usepackage{graphicx}       
\graphicspath{{media/}}     

\usepackage{enumitem}
\usepackage{amssymb}

\usepackage{multirow}
\usepackage{booktabs}

\usepackage{tcolorbox}
\tcbuselibrary{breakable} 

\usepackage{listings}

\lstdefinestyle{promptlisting}{
    basicstyle=\ttfamily\small, 
    breaklines=true,            
    columns=fullflexible,       
    keepspaces=true             
}

\title{
  Infinity-Parser2 Technical Report
}

\author{
  INF Team
}

\begin{document}
\maketitle

\begin{center}
  \vspace{-1.2cm}
  \normalsize
  \raisebox{-0.1em}{\includegraphics[height=1em]{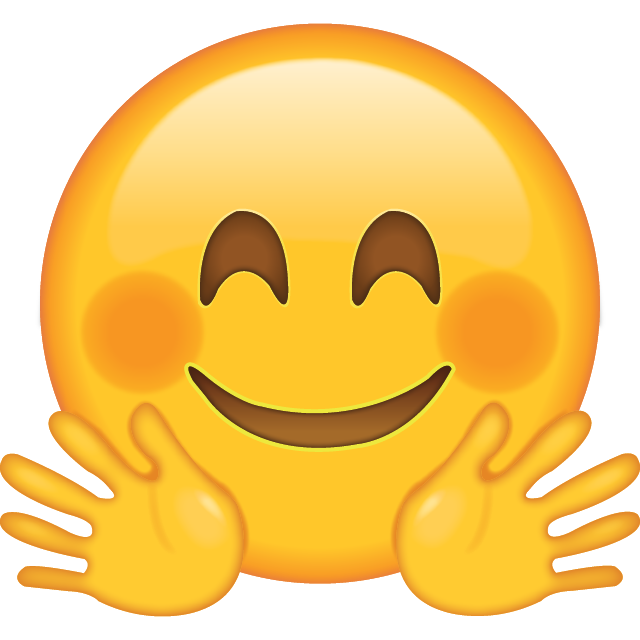}}\ \href{https://huggingface.co/collections/infly/infinity-parser2}{Models}\quad
  \raisebox{-0.1em}{\includegraphics[height=1em]{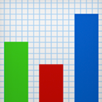}}\ \href{https://huggingface.co/datasets/infly/Infinity-Doc2-5M}{Dataset}\quad
  \raisebox{-0.1em}{\includegraphics[height=1em]{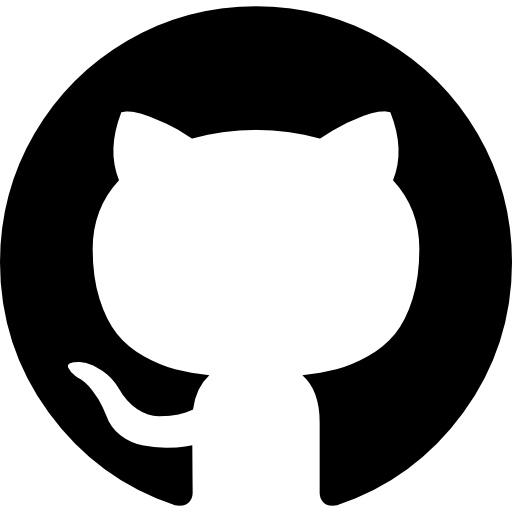}}\ \href{https://github.com/infly-ai/INF-MLLM}{Code}\quad
  \raisebox{-0.1em}{\includegraphics[height=1em]{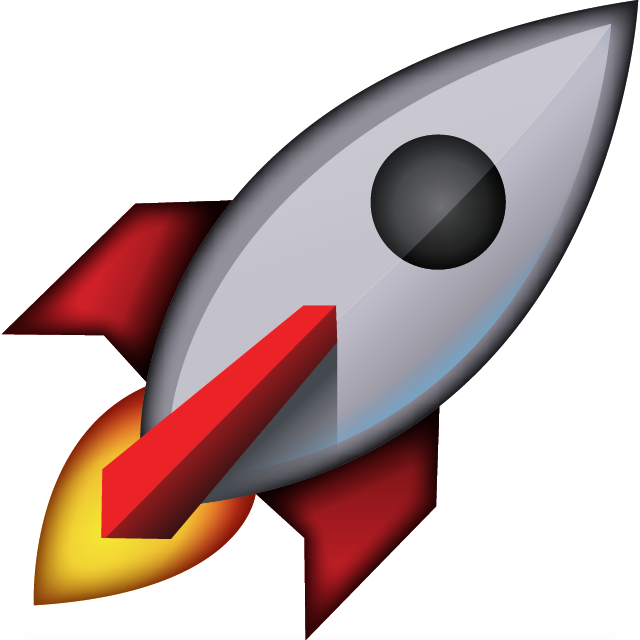}}\ \href{https://huggingface.co/spaces/infly/Infinity-Parser2-Demo}{Demo}\quad
  \vspace{0.5cm}
\end{center}

\begin{abstract}
We present \textbf{Infinity-Parser2}, a large multimodal model that couples a controllable data-synthesis pipeline with multi-task reinforcement learning for end-to-end document parsing, addressing the persistent scarcity of faithfully annotated parsing corpora. Our contributions are threefold. First, we build a scalable synthesis engine, pairing a controllable rendering framework with an iterative refinement loop, and use it to construct and open-source \textbf{Infinity-Doc2-5M}: a 5-million-sample bilingual (Chinese/English) corpus spanning diverse document types, annotated with element bounding boxes, canonical content forms (Markdown, HTML, LaTeX, SMILES, structured charts), and full-page reading order. Second, we introduce a verifiable, multi-task reward system that enables Joint Reinforcement Learning across eight co-trained objectives (document parsing, layout analysis, table parsing, math formula parsing, chart parsing, chemical formula parsing, document VQA, and general multimodal understanding), unifying perception, structure, and reasoning in a single optimization signal. Third, we release two variants under a shared architecture: \textbf{Infinity-Parser2-Flash}, optimized for low-latency inference with a \textbf{3.68$\times$} throughput gain over Infinity-Parser-7B, and \textbf{Infinity-Parser2-Pro}, engineered for precision-critical settings. Infinity-Parser2-Pro reaches state-of-the-art \textbf{87.6\%} on olmOCR-Bench and \textbf{74.3\%} on ParseBench, surpassing DeepSeek-OCR-2, PaddleOCR-VL-1.5, and MinerU2.5, with strong generalization to charts, chemical formulas, and document VQA.
\end{abstract}

\begin{figure*}[htbp]
  \centering
  \includegraphics[width=\textwidth]{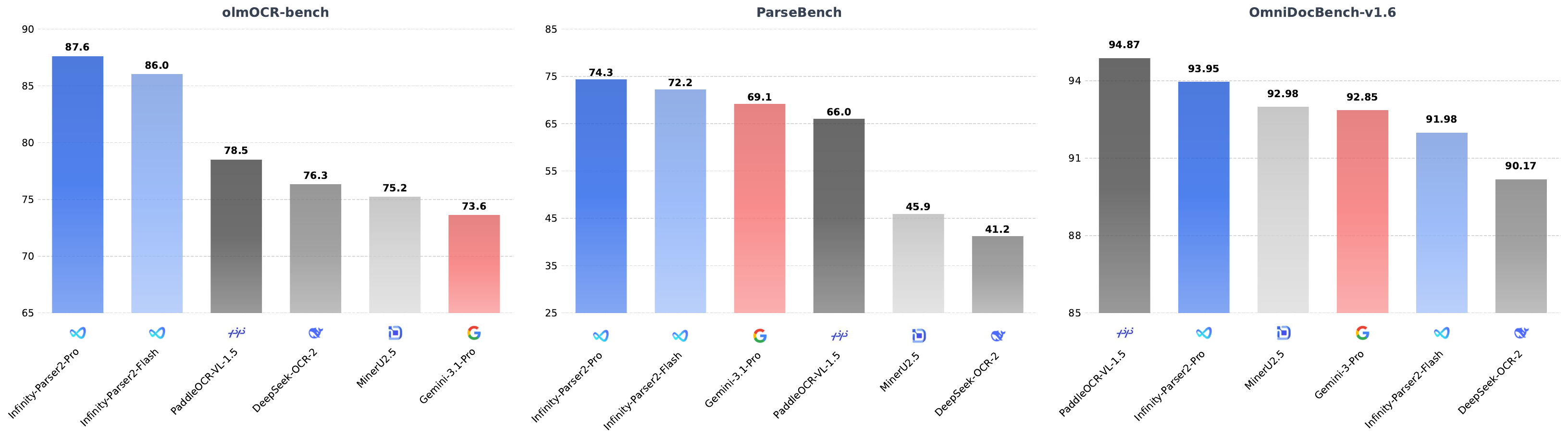}\\[4pt]
  \includegraphics[width=\textwidth]{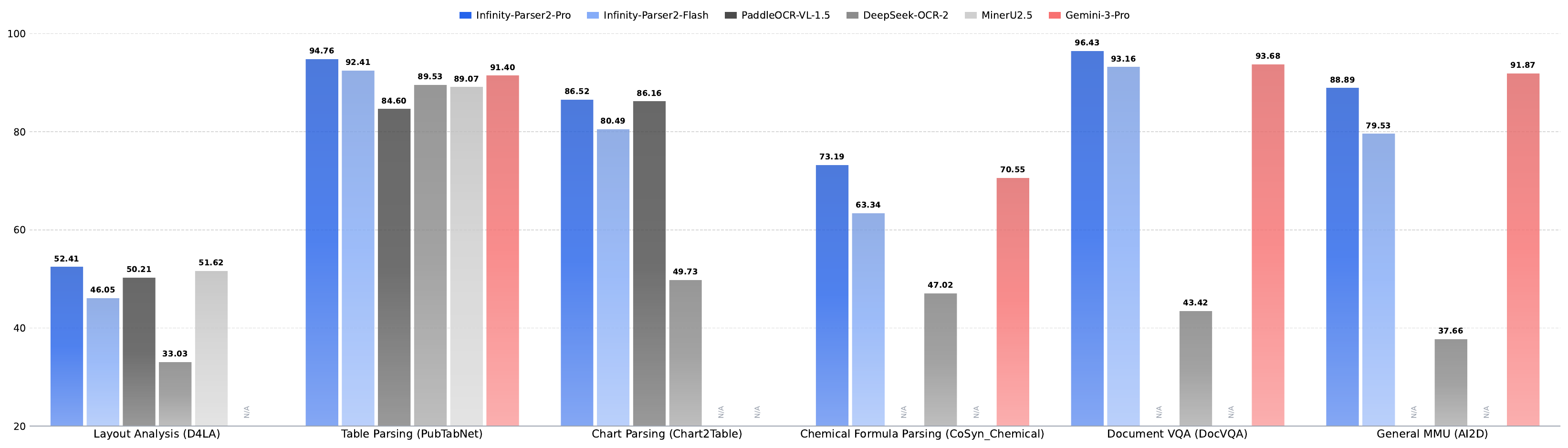}
  \caption{Performance evaluation. Top: document parsing benchmarks. Bottom: cross-domain multi-task capabilities.}
  \label{fig:perf}
\end{figure*}

\newpage
{
\setcounter{tocdepth}{2}
\tableofcontents
}

\newpage
\section{Introduction}

Document parsing has emerged as a pivotal frontier in multimodal understanding, serving as the structural bridge that transforms unstructured visual content into machine-readable, semantically grounded representations. As Large Language Models (LLMs) evolve from passive responders into autonomous agents that ingest, reason over, and act upon real-world artifacts, the ability to faithfully interpret heterogeneous documents (from scientific papers and financial filings to invoices, slides, and chemistry notes) has become a prerequisite for downstream reasoning and decision-making. Consequently, the scope of the field has long outgrown classical Optical Character Recognition (OCR): it now demands holistic, page-level understanding that jointly handles layout analysis, fine-grained element parsing, globally consistent reading order, and structured information extraction. Yet attaining reliable, near-human accuracy across this stack remains formidable, owing to the combinatorial diversity of layouts, the density of cross-modal cues, and the inherently structured nature of the targets.

Despite rapid progress, two structural bottlenecks continue to constrain the field. First, on the data side, prevailing Supervised Fine-Tuning (SFT) recipes built on Vision-Language Models (VLMs)~\cite{niu2025mineru2, cui2025paddleocr, wei2025deepseek} are bounded by the scarcity of large-scale, faithfully annotated parsing corpora that simultaneously cover heterogeneous layouts, fine-grained element semantics, and globally consistent reading order. As a result, SFT models that perform competitively in-distribution often degrade sharply on out-of-domain (OOD) templates, low-resource languages, or specialized verticals such as charts and chemical formulas. Second, on the optimization side, document parsing is intrinsically a multi-task problem (text recognition, layout grounding, table and formula structuring, chart decoding, and document-level reasoning are tightly coupled), yet existing pipelines typically treat these objectives as isolated heads or sequential stages. Even recent Reinforcement Learning (RL) approaches~\cite{wang2025infinity, chen2025logics, poznanski2025olmocr2} predominantly optimize a single, narrow textual signal, leaving structural fidelity, spatial alignment, and cross-task transfer under-exploited.

To surmount these limitations, we introduce \textbf{Infinity-Parser2}, a large multimodal model that couples a controllable data-synthesis pipeline with multi-task reinforcement learning to deliver comprehensive, end-to-end document parsing. To break the data bottleneck, we build a scalable synthesis engine featuring a controllable rendering framework paired with an iterative refinement loop, and use it to construct \textbf{Infinity-Doc2-5M}, a 5-million-sample bilingual (Chinese/English) corpus spanning a broad spectrum of document types and layouts (including academic papers, research reports and financial reports, newspapers, textbooks, exam papers, and magazines), covering single- and multi-column layouts, mixed text–table–figure compositions, and embedded mathematical, tabular, and molecular content. Each sample is richly annotated with element bounding boxes, canonical content forms (Markdown, HTML, LaTeX, SMILES, structured charts), and full-page reading order, providing the structural priors that downstream training requires. To consolidate the fragmented optimization landscape, we further advance the post-training recipe with a verifiable, multi-task reward system that enables Joint Reinforcement Learning across eight co-trained objectives (document parsing, layout analysis, table parsing, math formula parsing, chart parsing, chemical formula parsing, document VQA, and general multimodal understanding), unifying perception, structure, and reasoning under a single, end-to-end optimization signal. Finally, to meet the disparate demands of real-world deployment, we release two model variants under a shared architecture: \textbf{Infinity-Parser2-Flash}, tuned for low-latency inference, and \textbf{Infinity-Parser2-Pro}, engineered for maximum accuracy in precision-critical settings.

Extensive experiments across a broad suite of public benchmarks validate the effectiveness of Infinity-Parser2 (Figure~\ref{fig:perf}). Our system establishes new state-of-the-art results on end-to-end document parsing, with Infinity-Parser2-Pro reaching \textbf{87.6\%} on olmOCR-Bench and \textbf{74.3\%} on ParseBench, surpassing strong contemporary systems including DeepSeek-OCR-2, PaddleOCR-VL-1.5, and MinerU2.5. The Flash variant delivers a \textbf{3.68$\times$} throughput improvement (441 $\rightarrow$ 1{,}624 tokens/s) over our prior Infinity-Parser-7B while remaining competitive in accuracy, enabling efficient large-scale deployment. Beyond document-centric metrics, Infinity-Parser2 generalizes strongly to charts, chemical formulas, and document VQA, and retains robust general multimodal reasoning, evidencing the breadth of capabilities unlocked by joint multi-task RL.

Our primary contributions are summarized as follows:

\begin{itemize}[
  leftmargin=1.0em, labelsep=0.5em
]
\item We design a controllable rendering framework coupled with an iterative refinement pipeline to overcome the parsing-data bottleneck, and use it to construct and open-source \textbf{Infinity-Doc2-5M}, a 5-million-sample bilingual corpus spanning academic papers, research reports and financial reports, newspapers, textbooks, exam papers, and magazines, annotated with element-level bounding boxes, canonical content forms (Markdown, HTML, LaTeX, SMILES, structured charts), and full-page reading order.

\item We propose a verifiable, multi-task reward system that drives Joint Reinforcement Learning across eight co-trained objectives (document parsing, layout analysis, table parsing, math formula parsing, chart parsing, chemical formula parsing, document VQA, and general multimodal understanding), unifying perception, structure, and reasoning within a single optimization signal.

\item We release two deployment-aware variants under a shared architecture: \textbf{Infinity-Parser2-Flash}, achieving a 3.68$\times$ throughput improvement over Infinity-Parser-7B, and \textbf{Infinity-Parser2-Pro}, attaining \textbf{87.6\%} on olmOCR-Bench and \textbf{74.3\%} on ParseBench, surpassing DeepSeek-OCR-2, PaddleOCR-VL-1.5, and MinerU2.5.

\item To foster community innovation, we release the complete suite of assets, including the Infinity-Doc2-5M dataset, the source code, and both model variants (Infinity-Parser2-Flash and Infinity-Parser2-Pro).

\end{itemize}

\section{Related Work}

Document parsing has rapidly evolved into a core research direction at the intersection of vision, language, and structure understanding, with the goal of converting heterogeneous digital and scanned pages into faithful, machine-readable representations. Based on how the parsing capability is trained and assembled, prevailing approaches can be broadly grouped into three families: pipeline-based methods, end-to-end methods, and reinforcement learning (RL)-based methods.

\subsection{Pipeline-based Methods}

A long-standing line of work follows a pipeline-based paradigm. Systems such as MinerU2.5~\cite{niu2025mineru2} and PaddleOCR-VL~\cite{cui2025paddleocr} decompose document parsing into a sequence of specialized stages: a layout analyzer first detects and classifies semantic regions (text blocks, tables, formulas, and figures), after which task-specific expert models recognize the content of each region in parallel, and an optional reading-order module reassembles the page into a structured format such as Markdown or HTML. This modular design offers favorable engineering ergonomics, supports targeted optimization of individual components, and enables high-throughput inference on standardized layouts. However, the very staging that grants modularity also introduces error accumulation along the pipeline: imperfect region proposals or misclassified blocks propagate to downstream recognizers and reading-order predictors, where they are difficult to recover. The reliance on hand-engineered interfaces between stages further limits adaptability to documents whose layouts deviate from the assumptions baked into each module.

\subsection{End-to-end Methods}

End-to-end approaches instead seek to subsume the full parsing stack within a single model. Representative systems such as the dots.ocr series~\cite{dotsocr2025,zheng2026multimodalocrparsedocuments} and the DeepSeek-OCR series~\cite{wei2025deepseek,wei2026deepseek2} fine-tune large Vision-Language Models (VLMs) via supervised fine-tuning (SFT) to map a document image directly to its structured representation, jointly handling text recognition, layout grounding, table and formula structuring, and reading-order reconstruction. By learning holistic visual--textual representations, these models avoid hand-crafted inter-stage interfaces and achieve strong in-distribution accuracy. Their effectiveness, however, is tightly coupled to the diversity and fidelity of the training corpus: the scarcity of large-scale, faithfully annotated parsing data (covering heterogeneous layouts, fine-grained element semantics, and globally consistent reading order) remains a fundamental bottleneck, and SFT-only models often degrade noticeably on out-of-domain (OOD) templates, low-resource languages, or specialized verticals such as charts and chemical formulas.

\subsection{RL-based Methods}

To move beyond the limits of token-level imitation, a growing body of work explores reinforcement learning as a post-training strategy for document parsing. Approaches such as~\cite{wang2025infinity, chen2025logics, poznanski2025olmocr2, ling2025table2latex} optimize VLMs against outcome-oriented rewards that reflect the quality of the parsed output, and have demonstrated encouraging gains across document types. Nevertheless, most existing RL pipelines remain narrowly scoped: they optimize a single, predominantly textual reward signal and treat related sub-capabilities (element parsing, table and chart decoding, chemical formula recognition, and document-level question answering) as isolated heads or sequential stages, leaving structural fidelity, spatial alignment, and cross-task transfer under-exploited. In contrast, our Infinity-Parser2 introduces a verifiable, multi-task reward system that drives Joint Reinforcement Learning across eight co-trained objectives (document parsing, layout analysis, table parsing, math formula parsing, chart parsing, chemical formula parsing, document VQA, and general multimodal understanding), unifying perception, structure, and reasoning within a single end-to-end optimization signal and thereby promoting robust generalization across heterogeneous documents and downstream tasks.

\section{Data Curation}


Our method centers on three pillars that jointly drive the construction of Infinity-Parser2. First, we propose a \textbf{data iteration flywheel} (Sec.~\ref{sec:flywheel}), a closed-loop methodology in which model evaluation, bad-case mining, data construction, and fine-tuning continuously reinforce one another. Second, to support the data construction step at scale, we develop a dedicated \textbf{document data synthesis engine} (Sec.~\ref{sec:synthesis}) that produces layout-faithful documents through a three-stage pipeline. Third, we describe the resulting \textbf{task-specific datasets} (Sec.~\ref{sec:dataset}) used to train the model across structure analysis, element recognition, and document reasoning tasks. In short, Sec.~\ref{sec:flywheel} formalizes \emph{how} data evolves with the model, Sec.~\ref{sec:synthesis} details \emph{what tool} produces this data at scale, and Sec.~\ref{sec:dataset} describes \emph{what data} the model is ultimately trained on.

\subsection{Data Iteration Flywheel}
\label{sec:flywheel}

\textbf{Overview.} Data quality and data construction strategy play a critical role in multimodal document understanding. While raw document data are abundant, transforming them into high-quality, task-relevant training data under limited annotation and compute budgets remains a major challenge. Static data pipelines, which fix the dataset before training begins, cannot respond to the model's evolving weaknesses and therefore tend to over-invest in already-mastered patterns while under-covering rare or difficult cases.

To address this, we propose a model-driven data iteration flywheel that tightly couples data construction with model behavior. Rather than statically defining sample importance, the flywheel dynamically estimates sample value based on the current model's inference results and iteratively refines the dataset.

\begin{figure*}[htbp]
    \centering
    \includegraphics[width=\textwidth]{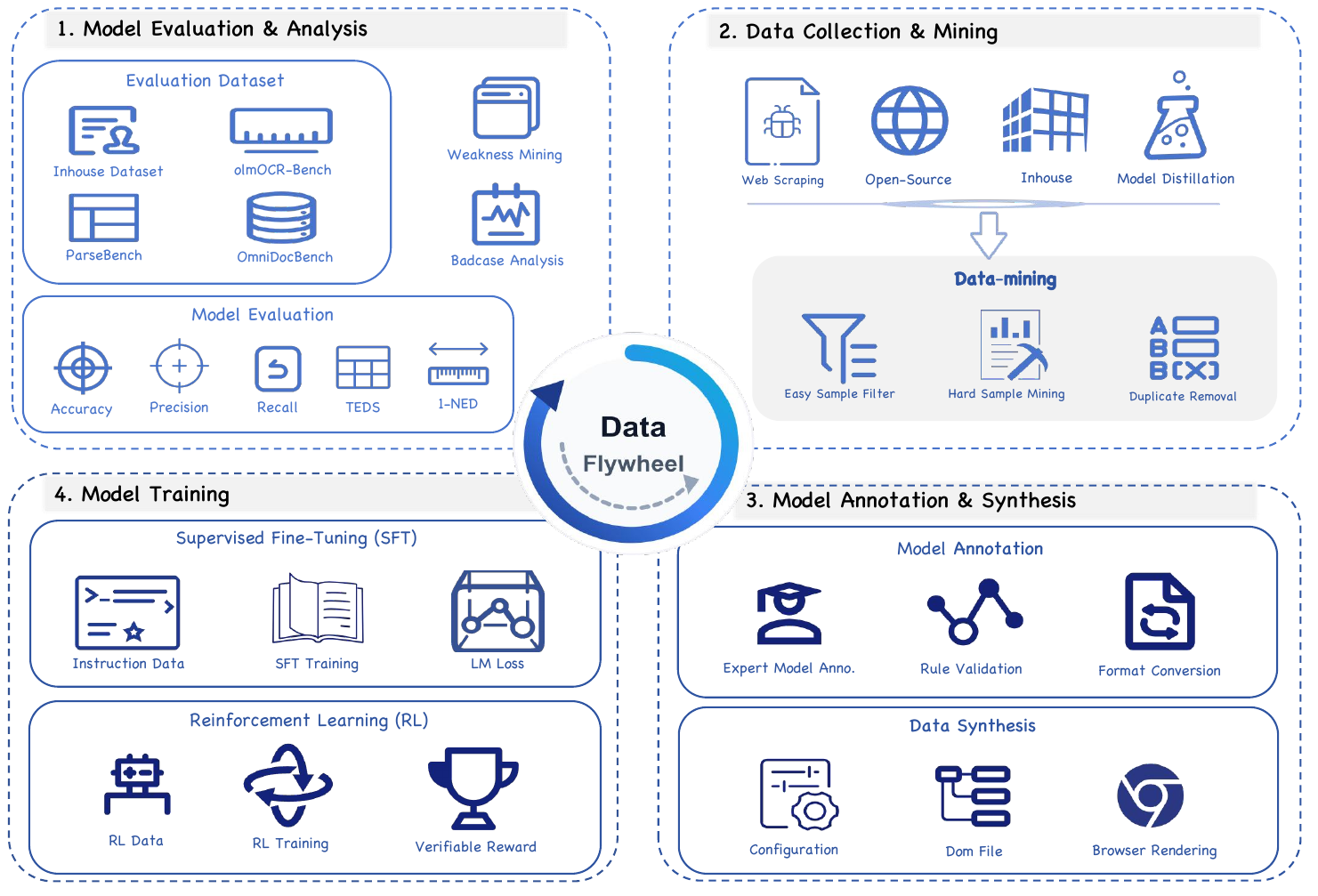}
    \caption{Overall pipeline of our data iteration flywheel, a closed loop of four stages. Stage 1 evaluates the current model and consolidates its failures into a weakness taxonomy, Stage 2 collects raw documents matching these weaknesses, Stage 3 annotates and synthesizes them into training data, and Stage 4 fine-tunes the model and hands it back to Stage 1 as the new baseline. Each turn thus converts the model's residual weaknesses into targeted training data for the next round, until benchmark gains saturate.}
    \label{fig:data_flywheel}
\end{figure*}

As illustrated in Figure~\ref{fig:data_flywheel}, each turn of the flywheel comprises four stages:
(1) \emph{Model Evaluation and Bad-case Analysis},
(2) \emph{Data Collection and Mining},
(3) \emph{Model Annotation and Data Synthesis}, and
(4) \emph{Model Fine-tuning and Iteration}.
We describe the details of each stage in the following sections.

\textbf{Stage 1: Model Evaluation and Bad-case Analysis.} At each flywheel iteration, the current parsing model~\footnote{The first iteration is bootstrapped from off-the-shelf open-source Qwen3.5-2B and Qwen3.5-35B-A3B. Subsequent iterations use the most recently fine-tuned model as the new baseline.} is evaluated on a multi-task benchmark suite covering end-to-end document parsing, layout analysis, and element-level parsing.  Each benchmark is scored by its official script at two levels: a per-sample accuracy on each page and a per-subcategory accuracy aggregated along axes such as document type, layout complexity, and element class. The subcategory breakdown is essential because a single overall score can hide systematic failures on narrow slices, e.g., a model with strong average accuracy may still collapse on financial reports or multi-column layouts.

Bad cases are then mined through a ranked diagnostic pipeline. Samples and sub-categories are first sorted by accuracy, with the lowest-scoring tail retained as the candidate pool so that diagnostic effort focuses on the model's empirically weakest regions. These candidates are inspected via a visualization workflow that overlays predictions onto the source page, surfacing document-level errors such as handwriting misrecognition, scan-induced degradation, and mis-handled domain-specific notation, element-level errors such as missed glyphs, fragmented formulas, and mis-parsed tables, along with layout-level errors such as reading-order flips, omissions in dense text, and mis-grouped regions. Diagnosed cases are consolidated into a three-axis weakness taxonomy along document type, element type, and layout pattern, with every case tagged on all three axes to yield a compact multi-label description of current weaknesses.

We emphasize that this pool serves a strictly \emph{diagnostic} purpose: because its samples originate from held-out evaluation benchmarks, they are never reused as training data, which would conflate measurement with optimization and contaminate subsequent evaluations. Instead, the accumulated weakness tags act as a demand signal that flows into Stage~2, where targeted acquisition is performed on disjoint, training-safe sources matching the same weakness profile.

\textbf{Stage 2: Data Collection and Mining.} Given the weakness tags from Stage~1, this stage samples raw, unlabeled documents from source domains according to those tags. Each tag is first translated into one or more acquisition queries: document-type tags map to domain and source-channel filters, element-type tags to content-density signals that surface documents densely populated by the target structure, and layout tags to page-level structural filters. The resulting query set is dispatched in parallel across three complementary sources (targeted web crawling, public document datasets, and internal proprietary corpora), with each tag routed to whichever subset best matches its profile. After harvesting, the pool is re-projected onto the Stage~1 taxonomy. Tags falling below a target count trigger additional harvesting, and persistent gaps escalate to the next flywheel iteration.

This stage serves a strictly \emph{sourcing} purpose, with no annotation or training-time use occurring here, decoupling raw-data supply from labeling so that each can be tuned independently. In the initialization round, roughly 2{,}000 samples are collected per tag. Subsequent rounds adaptively rebalance the budget toward tags that remain under-resolved.

\textbf{Stage 3: Model Annotation and Data Synthesis.} Given the raw, unlabeled pool from Stage~2, this stage converts it into training-ready supervision through two complementary paths: one anchored on harvested real documents, the other on targeted synthesis. The dual design reflects two constraints: per-sample human annotation does not scale to the flywheel cadence, and real-document harvesting alone leaves residual gaps on rare cells of the weakness taxonomy such as low-resource glyphs, atypical layouts, and structurally complex elements.

On the real-data path, each page first undergoes layout analysis into semantically coherent regions (text blocks, formulas, and tables), which are then routed to domain-specific expert models. We use dots.ocr~\cite{dotsocr2025} for layout analysis, PaddleOCR-VL~\cite{cui2025paddleocr} for text recognition, MinerU2.5~\cite{niu2025mineru2} for formula parsing, and Infinity-Parser~\cite{wang2025infinity} for table parsing. Label quality is ensured by low-confidence-score filtering and handcrafted rule-based filtering.

On the synthesis path, we generate targeted synthetic documents for the rare layouts, low-resource languages, and under-represented elements that real-data mining cannot economically supply. The synthesis engine (whose corpus selection, layout templates, and rendering parameters are conditioned directly on the Stage~1 tags) is detailed in Sec.~\ref{sec:synthesis}. Outputs from both paths converge into a newly annotated batch, which is mixed with historical training data accumulated from prior iterations and forwarded to Stage~4.

\textbf{Stage 4: Model Fine-tuning and Iteration.} Given the merged training corpus from Stage~3, this stage fine-tunes the current parsing model on the expanded dataset. Each round's newly annotated data is appended to the existing training corpus rather than replacing it, so that fine-tuning at every iteration still includes the targeted examples from all prior rounds. This prevents the model from regressing on long-tail weaknesses that earlier iterations have already addressed. The fine-tuned model is then handed back to Stage~1 of the subsequent iteration as the new baseline, where the diagnostic cycle resumes on a fresh evaluation of the same multi-task benchmark suite.

Because the benchmark suite remains fixed across iterations, per-iteration scores are directly comparable and serve as the convergence signal. Looping through Stages~1--4 round after round, the flywheel progressively converts each iteration's residual weaknesses into the next iteration's targeted training data, tightening coverage on the long tail of document types, element classes, and layout patterns flagged in earlier rounds. Iterations continue until benchmark gains across consecutive rounds become negligible, at which point we consider the flywheel converged.

\subsection{DOM-Based Document Synthesis Engine}
\label{sec:synthesis}

Human annotation is prohibitively expensive, while labels produced by expert models inevitably introduce noise. Data synthesis thus offers a promising alternative for generating document data at scale. However, no existing synthesis paradigm fully satisfies the requirements of document data generation. Pixel-level, diffusion-based generation~\cite{textdiffuser,anyword3m} is well suited to scene-text images yet fails to provide reliable structural labels for documents. Block-level engines~\cite{tableformer,cosyn,decimer} synthesize individual elements (such as tables, charts, and chemical formulas) but cannot assemble complete page-level documents. LaTeX-source synthesis~\cite{nougat,vary} is structurally rich but offers no fine-grained bounding-box annotations. HTML-based generation~\cite{wang2025infinity,poznanski2025olmocr2} yields page-level images but relies on rigid, hand-authored templates with limited layout diversity. These gaps call for a synthesis engine that produces diverse, page-level documents while preserving fine-grained structural annotations.

\begin{figure*}[htbp]
    \centering
    \includegraphics[width=\textwidth]{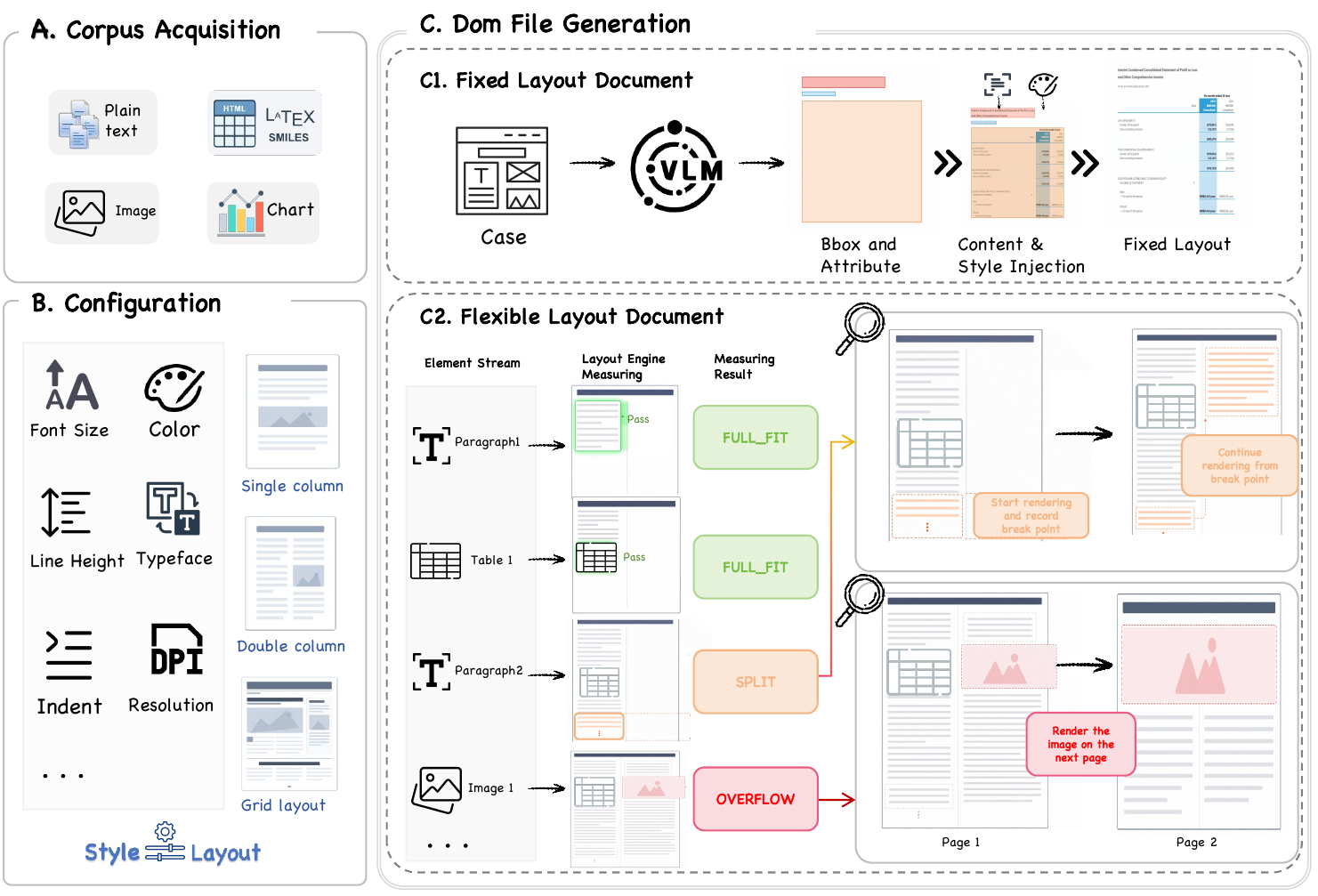}
    \caption{Overall framework of our DOM-based document data synthesis engine. \textbf{(A)}~Corpus acquisition gathers plain text, structured elements, and visual assets. \textbf{(B)}~Configuration sets the page layout and element styles of a document template. \textbf{(C)}~DOM file generation instantiates documents along two paths. The fixed-layout path reproduces the layout of a real exemplar by vision-language parsing followed by content and style injection, while the flexible-layout path streams elements through a layout engine that measures each element and resolves it as \texttt{FULL\_FIT}, \texttt{SPLIT}, or \texttt{OVERFLOW} to produce content-elastic, multi-page layouts. Fine-grained labels are read directly off the laid-out DOM.}
    \label{fig:data_synth_engine}
\end{figure*}

We therefore design a novel DOM-based (Document Object Model) synthesis engine that meets both requirements. As illustrated in Figure~\ref{fig:data_synth_engine}, the engine operates in three stages: corpus acquisition, configuration, and DOM file generation. Its core idea is to adopt the DOM as a unified intermediate representation that binds content, hierarchical structure, and geometric coordinates within a single tree. Conditioned on the acquired corpus and a sampled template, the DOM file generation stage instantiates documents along two complementary paths: a fixed-layout path that reproduces the layout of real document exemplars, and a flexible-layout path that composes diverse, content-elastic page layouts. Because the coordinates are carried by the DOM itself, the engine extracts fine-grained, multi-task labels automatically at rendering time, without any post-hoc annotation.

\textbf{Corpus Acquisition.} This stage gathers the raw content (plain text, structured elements, and visual assets) that populates the synthetic documents. \textit{General text} comprises multilingual paragraphs drawn from public sources~\cite{wikimedia_dumps}, internal databases, and model distillation, filtered by length, language (English, Simplified Chinese, Traditional Chinese, etc.), and domain (academic, financial, educational, etc.) to match downstream document types. \textit{Structured elements} include tables, mathematical formulas, charts, and chemical formulas, represented as HTML, LaTeX, CSV, and SMILES strings collected from arXiv and public datasets~\cite{pubtabnet,unimernet,decimer}. \textit{Visual assets} consist of figures and images~\cite{ng2021understandingguidedimagecaptioning} embedded as document illustrations.

\textbf{Configuration.} A template defines the global appearance of a document through two complementary groups of parameters:

\begin{itemize}[leftmargin=1.0em, labelsep=0.5em]
    \item \textbf{Page-level parameters} govern the canvas and its functional regions: page size and margins, rendering resolution, the writing mode (horizontal or vertical) and text direction (left-to-right or right-to-left), the layout mode of the main body (single-column, multi-column, and grid layout), and the placement of auxiliary regions such as headers, footers, page numbers, and rotated side bars.
    \item \textbf{Element-level parameters} specify the visual style of each content type (body text, multi-level headings, ordered and unordered lists, tables, figures, inline and block equations, code, algorithms, and references), covering typography (font family and size, color, line height, alignment, indentation, and spacing) together with a block-flow attribute that determines how each block wraps and breaks across columns and pages.
\end{itemize}

Rather than fixing these parameters, each template stores a preset default value together with a small perturbation range for every parameter. At generation time, we jitter the defaults (font sizes, margins, line spacing, colors, and region geometry) within these bounds, so that a single template instantiates into a family of documents that share the same logical structure yet differ in fine-grained appearance. This turns every template into a source of controlled data augmentation, substantially enlarging visual diversity while keeping each layout typographically valid. Moreover, the perturbation ranges need not be uniform: when the engine is driven by our data flywheel (Sec.~\ref{sec:flywheel}), they can be biased toward the layout and document-type weaknesses recorded in its weakness taxonomy, so that augmentation concentrates on the regions where the parser currently underperforms. Maintaining a library of such templates across common document types (academic papers, financial reports, books, letters, and the like) lets the engine cover a broad space of real-world layouts.

\textbf{DOM File Generation.} Given the acquired corpus and a sampled template, this stage instantiates documents along two complementary paths. The \emph{fixed-layout} path targets document types whose visual form must closely match real specimens, such as financial reports and forms. It takes a real page as a layout exemplar, uses a vision-language model to recover its region bounding boxes and attributes, and injects sampled content and template styles into that layout, producing documents that preserve the exemplar's structure while varying its content. The \emph{flexible-layout} path instead composes each document from scratch, letting page count and element placement emerge from the content so as to cover a broad space of multi-column, multi-page layouts.

In the flexible-layout path, the engine first instantiates a complete DOM tree that serves as the document's logical intermediate representation. Every node is typed against a single element schema (paragraph, multi-level heading, list, table, figure, inline or block formula, code, header, footer, and side bar), and this type later drives renderer dispatch. The schema is open, so new modalities such as charts and chemical formulas are introduced by registering a node type and its renderer rather than by modifying the pipeline. Each node carries three pieces of information:
\begin{itemize}[leftmargin=1.0em, labelsep=0.5em]
    \item its \textbf{content}, retained in the native source form of each modality (plain text, LaTeX for formulas, HTML for tables, SMILES for molecules, etc.), so that nothing is lost to a lossy intermediate encoding.
    \item its \textbf{structural role} (heading level, reading order, and parent--child containment), which makes the document hierarchy explicit.
    \item a \textbf{style descriptor} sampled from the template (typography, alignment, spacing), together with a \emph{block-flow constraint} that declares how the node may be paginated: \emph{flexible} (freely splittable, e.g.\ body text, headings, and code), \emph{continuous} (splittable but order-preserving, e.g.\ tables and lists), or \emph{atomic} (indivisible, e.g.\ a figure and its caption).
\end{itemize}

Crucially, the DOM emitted at this point is purely \emph{logical}: it fixes content, hierarchy, and appearance but leaves all geometry unresolved, deferring coordinate computation to rendering, which repopulates the \emph{same} tree with exact positions. The DOM is thus a single source of truth from which one traversal yields ground-truth annotations for multiple downstream tasks: reading order, element-type and hierarchy labels, and structured text (Markdown/LaTeX/HTML) reconstructed directly from node content, without any additional labeling step. This explicit encoding of structural relations is precisely why we adopt a DOM rather than a flatter intermediate such as a Markdown or LaTeX string, from which hierarchy, reading order, and element boundaries would otherwise have to be recovered post hoc.

The flexible path then resolves this logical DOM into a concrete, paginated document, rasterizes it, and reads every label directly off the laid-out tree. Rather than \emph{predicting} positions, the engine delegates typesetting to a real browser layout engine and then \emph{reads back} exact coordinates, so that all geometric labels are correct \emph{by construction}. Three tightly coupled components carry out this rendering.

\textit{Layout planning.} The page canvas is first divided into functional regions (a main body together with headers, footers, and side bars), and the body is partitioned into a configurable arrangement of columns, rows, or grid cells, with cell extents taken from template ratios or jittered for diversity. The resulting regions are linearized into a reading sequence consistent with the document's writing mode (horizontal or vertical) and text direction (left-to-right or right-to-left), so that reading-order labels follow directly from the plan rather than being inferred from pixels.

\textit{Content-elastic pagination.} Elements are streamed into the planned regions one at a time, so that the page count emerges from the content itself rather than being fixed in advance. Before being committed, each element is rendered off-screen on a hidden measurement canvas and measured at its true rendered size, so that placement never relies on size estimates. Comparing this measured extent against the space left in the current region yields one of three outcomes (Figure~\ref{fig:data_synth_engine}). An element that fits entirely is placed in position (\texttt{FULL\_FIT}). An element that fits only partially is broken at a measured point and continues at the top of the next region, for instance the second column of the same page (\texttt{SPLIT}). An element that cannot be accommodated in the remaining space is deferred to the next region or page (\texttt{OVERFLOW}). How a block may be broken follows its block-flow constraint. \emph{Flexible} blocks break line by line across regions, \emph{continuous} blocks break while preserving order with table headers repeated on every continuation, and \emph{atomic} blocks are never broken. When an atomic block overflows, the engine performs a \emph{backward reflow}, either re-laying the most recently placed blocks to reclaim room on the current page or deferring the block to the next page, which removes orphaned fragments while preserving the global reading order.

\textit{Element pre-rendering.} Complex elements pass through dedicated sub-renderers \emph{before} measurement, so that their true footprint, rather than an approximation, drives layout. Formulas are typeset to SVG by a LaTeX engine and synchronized on web-font readiness, with display equations assigned document-global numbering. Tables are rendered from HTML/CSS inside an isolated shadow DOM, their logical cell grid (including row and column spans) recovered, and cross-page cut points located by binary search over measured geometry within the boundary cell, under typographic safeguards that forbid mid-word breaks, line-leading punctuation, and orphaned sub/superscripts. Images and custom fonts (CJK, Latin, and mixed scripts) are fully loaded before measurement to prevent reflow. Dispatch is keyed on element type and routed through a shared library manager, so the engine extends to additional notation renderers, such as charts and chemical formulas, by registering a renderer rather than altering the pipeline.

Finally, each laid-out page is rasterized at high resolution by device-pixel supersampling (optionally composited over template-controlled background textures for visual variability), and, in lockstep with the image, per-task labels are exported by traversing the laid-out DOM: character-, line-, and cell-level bounding boxes obtained from the browser's native geometry queries and offset into absolute page coordinates, alongside reading-order, element-type, hierarchy, and structured-text labels. Because the pixels and the annotations are read from one and the same DOM, they are exact and mutually consistent by construction, requiring no post-hoc detection or OCR.

\subsection{Dataset Preparation}
\label{sec:dataset}

\begin{table*}[t]
\centering
\caption{Overview of the final training data composition of Infinity-Parser2. Sources span public datasets, flywheel-mined real documents, and synthesis-engine data. The original size denotes the total number of available samples, while the sampled size indicates the number retained after balanced sampling.}
\label{tab:dataset_summary}
\resizebox{\textwidth}{!}{%
\begin{tabular}{l|c|l|c|c}
\toprule
\textbf{Task} & \textbf{Sub-Task} & \textbf{Dataset} & \textbf{Original Size} & \textbf{Sampled Size} \\
\midrule
\multicolumn{5}{l}{\textit{Document Structure Tasks}} \\
\midrule
\multirow{5}{*}{Document Parsing}
  & \multirow{3}{*}{doc2json} & Pseudo-Labeled Web Documents & 776K & 776K\\
  &                           & Manually Labeled Newspaper               & 2.5K & 2.5K \\
  &                           & Synthesized Documents                             & 251K & 251K \\
\cmidrule(l){2-5}
  & \multirow{2}{*}{doc2md}   & Infinity-Doc-400K~\cite{wang2025infinity}    & 57K  & 57K \\
  &                           & Web-Crawled Handwritten Docs           & 62K  & 62K \\
\midrule
\multirow{2}{*}{Layout Analysis}
  & \multirow{2}{*}{layout\_analysis}       & M$^{6}$Doc~\cite{cheng2023m6doclargescalemultiformatmultitype} & 4K & 4K \\
  &                           & Synthesized Documents                             & 30K  & 30K \\
\midrule
\multicolumn{5}{l}{\textit{Element-level Parsing Tasks}} \\
\midrule
\multirow{3}{*}{Table Parsing}
  & table2html                & PubTabNet~\cite{pubtabnet}, FinTabNet~\cite{fintabnet}, MMTab-HTML~\cite{mmtab}             & 676K & 676K \\
\cmidrule(l){2-5}
  & \multirow{2}{*}{table2md} & MMTab-MD~\cite{mmtab}                                      & 27K  & 27K \\
  &                           & Synthesized Tables                             & 309K & 309K \\
\midrule
Math Formula Parsing & formula2latex & im2latex~\cite{im2latex}, UniMER~\cite{unimernet}, HME~\cite{hme100k}, CROHME~\cite{crohme}                       & 1.1M & 631K \\
\midrule
\multirow{3}{*}{Chart Parsing}
  & chart2table & ChartSFT~\cite{chartsft}, Chart-MoE~\cite{chartmoe}, UniChart~\cite{unichart}                             & 1.5M & 315K \\
  & chart2json  & Chart-MoE~\cite{chartmoe}, ChartQA~\cite{chartqa}                                        & 900K & 315K \\
  & chart2code  & ChartGen~\cite{chartgen}, Chart2Code~\cite{chart2code}, Chart-MoE~\cite{chartmoe}                           & 1.2M & 315K \\
\midrule
Chemical Formula Parsing & chem2smiles & Synthesized Chemical Formula Images & 5.9M & 315K \\
\midrule
\multicolumn{5}{l}{\textit{Reasoning and Generalization Tasks}} \\
\midrule
\multirow{2}{*}{Document VQA} & \multirow{2}{*}{docvqa} & DocVQA~\cite{docvqa}, ChartQA~\cite{chartqa}, AI2D~\cite{ai2d}, DocReason25K~\cite{docreason25k},& \multirow{2}{*}{1.6M} & \multirow{2}{*}{315K} \\
& & InfoVQA~\cite{infovqa}, DT-VQA~\cite{dtvqa}, TinyChart~\cite{tinychart} & & \\
\midrule
\multirow{2}{*}{General Multimodal Understanding} & \multirow{2}{*}{general\_mmu} & M4-Instruct~\cite{m4instruct}, LLaVA-v1.5~\cite{llava15}, ShareGPT-4V~\cite{sharegpt4v}, ShareGPT-4o~\cite{sharegpt4o}, CogVLM~\cite{cogvlm} & \multirow{2}{*}{3.2M} & \multirow{2}{*}{631K} \\
& & ALLaVA-Instruct~\cite{allava}, LVIS-Instruct~\cite{lvisinstruct4v}, TextVQA~\cite{textvqa}, OCRVQA~\cite{ocrvqa}, AnyWord-3M~\cite{anyword3m} & & \\
\midrule
\multicolumn{5}{l}{\textit{Others}} \\
\midrule
Blank-page Handling & - & Synthesized Documents & 5K & 5K \\
\bottomrule
\end{tabular}%
}
\end{table*}

This section describes the final training data composition of Infinity-Parser2, assembled after the data iteration flywheel (Sec.~\ref{sec:flywheel}) has converged. Each task draws from three sources: public datasets, real documents mined by the flywheel, and synthetic documents produced by the synthesis engine (Sec.~\ref{sec:synthesis}). For each task we report its definition, data sources, scale, annotation or synthesis strategy, and output format. A consolidated overview is given in Table~\ref{tab:dataset_summary}.

\textbf{Document Structure Tasks.} These tasks operate on the document as a whole and produce structural representations.

\begin{itemize}[leftmargin=1.0em, labelsep=0.5em]
    \item \textbf{Document Parsing.} Given an input document image, the model transforms it into a structured representation by identifying and ordering document elements according to the natural reading order, annotating each detected region with its element type, content, and spatial location, thereby converting unstructured pages into structured data. For the doc2json task, the training data combine real documents mined by the flywheel with synthetic documents produced by the synthesis engine: we mine 776K documents from the web spanning nine categories (exam papers, slides, academic papers, books, textbooks, magazines, notes, newspapers, and financial reports) and obtain high-quality pseudo labels for them through Stage 3 of our flywheel, manually annotate 2.5K complex newspaper images, and synthesize an additional 251K documents with our synthesis engine. For the doc2md task, we sample 57K high-quality examples from Infinity-Doc-400K~\cite{wang2025infinity} and mine 62K scanned handwritten image–text pairs from the Library of Congress digital archives.

    \item \textbf{Layout Analysis.} Layout analysis detects document elements in reading order and predicts their types and bounding boxes without recognizing their content. The training data comprise 4K samples from the publicly available M$^{6}$Doc dataset~\cite{cheng2023m6doclargescalemultiformatmultitype}, supplemented with 30K samples from our synthesis engine, for which layout labels are obtained directly from the DOM at near-zero cost. The outputs follow a standard layout-element schema covering titles, paragraphs, formulas, tables, figures, captions, headers, and footers.
\end{itemize}

\textbf{Element-level Parsing Tasks.} These tasks operate on individual document elements and produce element-specific content.

\begin{itemize}[leftmargin=1.0em, labelsep=0.5em]
    \item \textbf{Table Parsing.} Table parsing focuses on recognizing the content of table elements and producing structured sequences such as HTML and Markdown. For the table2html task, we collect 676K samples from PubTabNet~\cite{pubtabnet}, FinTabNet~\cite{fintabnet}, and MMTab-HTML~\cite{mmtab}. For the table2md task, we obtain 27K samples from MMTab-MD~\cite{mmtab} and generate an additional 309K samples with our synthesis engine.

    \item \textbf{Math Formula Parsing.} Math formula parsing recognizes the content of mathematical formula elements and produces structured LaTeX sequences. We collect 1.1M samples from im2latex~\cite{im2latex}, UniMER~\cite{unimernet}, HME~\cite{hme100k}, and CROHME~\cite{crohme}.

    \item \textbf{Chart Parsing.} Chart parsing converts visual charts (bar, line, pie, scatter, etc.) into structured data records and Python code. For the chart2table task, we collect 1.5M samples from ChartSFT~\cite{chartsft}, Chart-MoE~\cite{chartmoe}, and UniChart~\cite{unichart}. For the chart2json task, we obtain 900K samples from Chart-MoE~\cite{chartmoe} and ChartQA~\cite{chartqa}. For the chart2code task, we collect 1.2M samples from ChartGen~\cite{chartgen}, Chart2Code~\cite{chart2code}, and Chart-MoE~\cite{chartmoe}.

    \item \textbf{Chemical Formula Parsing.} Chemical formula parsing recognizes 2D chemical formula diagrams and outputs their SMILES representations. Because real-world annotated data are scarce, this task relies heavily on our synthesis engine, which renders chemical formulas from SMILES corpora into diagram images with controlled visual styles. Using four rendering backends (CDK~\cite{cdk}, RDKit~\cite{rdkit}, OpenChemLib~\cite{openchemlib}, and Indigo~\cite{indigo}) and a corpus drawn from DECIMER~\cite{decimer}, we generate 5.9M samples.
\end{itemize}

\textbf{Reasoning and Generalization Tasks.} These tasks go beyond per-element recognition and require document-level understanding or cross-domain generalization.

\begin{itemize}[leftmargin=1.0em, labelsep=0.5em]
    \item \textbf{Document VQA.} Document VQA requires answering natural-language questions grounded in a document image. We collect 1.6M samples from DocVQA~\cite{docvqa}, ChartQA~\cite{chartqa}, AI2D~\cite{ai2d}, DocReason25K~\cite{docreason25k}, InfoVQA~\cite{infovqa}, DT-VQA~\cite{dtvqa}, and TinyChart~\cite{tinychart}.

    \item \textbf{General Multimodal Understanding.} To preserve the open-domain vision-language alignment of the underlying VLM and mitigate catastrophic forgetting during document-heavy fine-tuning, we additionally include general image–text pairs and instruction data. In total, we collect 3.2M samples from public multimodal corpora, including M4-Instruct~\cite{m4instruct}, LLaVA-v1.5~\cite{llava15}, ShareGPT-4V~\cite{sharegpt4v}, ShareGPT-4o~\cite{sharegpt4o}, CogVLM~\cite{cogvlm}, ALLaVA-Instruct~\cite{allava}, LVIS-Instruct~\cite{lvisinstruct4v}, TextVQA~\cite{textvqa}, OCRVQA~\cite{ocrvqa}, and AnyWord-3M~\cite{anyword3m}.
\end{itemize}

\textbf{Blank-page Handling.} Beyond the task-specific data above, we additionally curate 5K blank-page samples spanning varying resolutions, background colors, and background textures. For inputs containing no valid content, the model is trained to emit task-specific empty outputs (e.g., empty Markdown, empty JSON array). This explicitly suppresses hallucinations on degenerate inputs and improves system robustness.

\textbf{Dataset Summary.} Overall, the final training set (Table~\ref{tab:dataset_summary}) combines public, flywheel-mined, and synthesis-engine data across structure-level, element-level, and reasoning tasks, comprising approximately 5M samples after balanced sampling. The sampling ratios are designed to ensure balanced coverage along the multi-attribute schema introduced in Sec.~\ref{sec:flywheel}, and to remain consistent with the bad-case distribution observed at flywheel convergence. To foster further progress in document parsing, after removing business-sensitive and privacy-related samples, we open-source the \textbf{Infinity-Doc2-5M} dataset at~\url{https://huggingface.co/datasets/infly/Infinity-Doc2-5M}.

\section{Model Training}

\begin{figure*}[htbp]
    \centering
    \includegraphics[width=\textwidth]{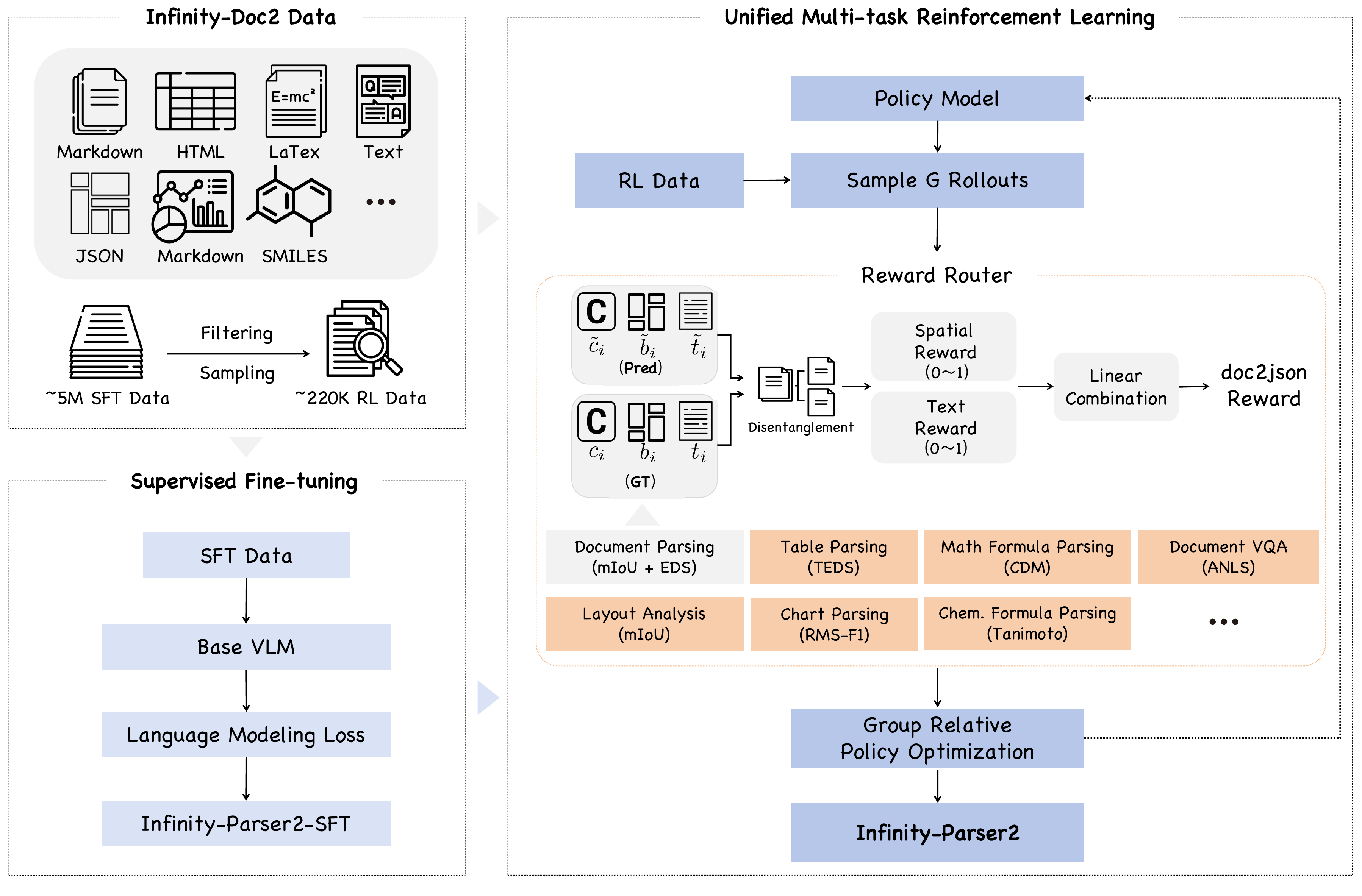}
    \caption{Overall framework of Infinity-Parser2. We cast document parsing as a unified image-to-sequence problem: a single model maps a document image and a task instruction to a structured output (Markdown, HTML, LaTeX, JSON, etc.), with the target task selected solely by the instruction. Training proceeds in two stages, supervised fine-tuning (SFT) for broad multi-task parsing capability followed by joint reinforcement learning with verifiable rewards (RLVR) that co-trains all tasks under one policy, each rewarded by its own native evaluation metric.}
    \label{fig:overall_framework}
\end{figure*}

\subsection{Overall Training Strategy}

We formulate document parsing as a unified image-to-sequence generation problem. Given a document image $I$ and a task instruction $c$ that specifies the target task and output format, the model $\pi_\theta$ autoregressively generates a token sequence $y=(y_1,\dots,y_T)\sim\pi_\theta(\cdot\mid I,c)$ that decodes into a structured representation $[e_1,\dots,e_N]$, an ordered list of document elements rendered in Markdown, HTML, LaTeX, JSON, etc., where each $e_i$ carries the $i$-th element's type, content, and spatial location (where applicable). A single parameter set $\theta$ thus serves the entire parsing stack (document-, element-, and reasoning-level tasks) under one autoregressive interface, with the task selected solely through $c$.

As illustrated in Figure~\ref{fig:overall_framework}, training proceeds in two stages. We first perform supervised fine-tuning (SFT) to instill broad parsing capability through next-token prediction, and then conduct joint reinforcement learning with verifiable rewards (RLVR) to align the model with task-native evaluation metrics and to close the gap between teacher-forced training and autoregressive inference. This SFT-then-RLVR recipe follows recent document-parsing systems~\cite{wang2025infinity,poznanski2025olmocr2}. Our contribution lies in the design of the reward system that drives the RL stage rather than in the recipe itself. Concretely, our training design rests on three choices:

\begin{itemize}[leftmargin=1.0em, labelsep=0.5em]
    \item \textbf{Unified multi-task RLVR.} We co-train eight objectives (document parsing, layout analysis, table parsing, math formula parsing, chart parsing, chemical formula parsing, document VQA, and general multimodal understanding) within a single RL stage, rewarding each with a structure-aware metric native to that task (Sec.~\ref{sec:reward_design}).
    \item \textbf{Disentangled spatial--textual reward.} For structured-layout parsing, we introduce a reward that scores element localization and content fidelity separately, a signal absent from prior edit-distance- or unit-test-based rewards (Sec.~\ref{sec:str}).
    \item \textbf{Reward routing with normalization.} All objectives are optimized jointly through per-task reward routing and cross-task reward normalization, so that a single policy is trained over heterogeneous tasks with comparable gradient scales (Sec.~\ref{sec:rl_opt}).
\end{itemize}

Under this recipe we train two deployable variants on a shared architecture (Infinity-Parser2-Flash for low-latency inference and Infinity-Parser2-Pro for precision-critical settings), whose configurations are detailed in our experimental setup.

\subsection{Supervised Fine-Tuning}
\label{sec:sft}

We adopt Qwen3.5~\cite{qwen3.5} as our foundation model, a vision--language model pre-trained on large-scale multimodal data with strong inherent document-understanding capability. We fine-tune it on Infinity-Doc2-5M (Sec.~\ref{sec:dataset}) with a supervised next-token-prediction objective:
\begin{equation}
\mathcal{L}_{\mathrm{SFT}}(\theta) = -\,\mathbb{E}_{(I,c,y)\sim\mathcal{D}}\sum_{t=1}^{|y|}\log\pi_\theta\!\left(y_t \mid y_{<t}, I, c\right),
\end{equation}
where $\mathcal{D}$ denotes the training corpus and each triple $(I,c,y)$ pairs an input image and task instruction with its target sequence. The SFT corpus spans all sub-tasks, including the full general-multimodal-understanding set, which preserves the open-domain vision--language alignment of the base model and mitigates catastrophic forgetting during document-heavy fine-tuning. This stage endows the model with broad multi-task parsing competence and provides a strong initialization for the subsequent RL stage.

\subsection{Joint Reinforcement Learning with Verifiable Rewards}
\label{sec:reinforcement_learning}

\subsubsection{Motivation}

The next-token-prediction objective of SFT, trained with teacher forcing, optimizes the model under a ground-truth-conditioned distribution that differs from the autoregressive distribution encountered at inference. This exposure bias encourages the model to overfit the sequential dependencies and annotation noise of the training data, limiting its generalization to unseen document types. To bridge this train--inference gap, we add a reinforcement learning stage in which the model generates full sequences (rollouts) and is optimized against task-specific reward signals, promoting robust generation strategies that transfer better to out-of-distribution document formats. The effectiveness of this stage hinges entirely on the reward signals, whose design we describe next.

\subsubsection{Reward Design}
\label{sec:reward_design}

All RL signals in Infinity-Parser2 are \emph{verifiable rewards}: each is computed by a deterministic, reference-based metric rather than a learned reward model, which keeps the signal faithful to ground truth and structurally limits reward hacking. We adopt a single guiding principle, \emph{metric-as-reward}, under which every task is rewarded by its own native evaluation metric, so that RL optimizes directly for the quantity by which the task is ultimately judged. All rewards are normalized to $[0,1]$ to keep gradient magnitudes comparable across co-trained tasks (Sec.~\ref{sec:rl_opt}). When an output jointly encodes multiple facets (most notably structured-layout parsing, which couples element localization with textual content), a single scalar metric cannot separate distinct error modes. We therefore factorize the reward into facet-specific terms, as detailed below.

\paragraph{Disentangled Spatial--Textual Reward for Structured Parsing.}
\label{sec:str}
The doc2json task produces a joint detection--recognition structure: an ordered list of elements $E_i=(c_i,b_i,t_i)$, each carrying a category $c_i$, a bounding box $b_i$, and textual content $t_i$. Scoring such an output with a single edit distance over its serialized form is ill-suited on two counts: it conflates \emph{where} an element is with \emph{what} it contains, yielding an ambiguous learning signal, and it measures geometric error through coordinate digit strings, where a few-pixel shift can perturb the reward disproportionately. We therefore disentangle the reward into a spatial term and a textual term, each evaluated in its native space.

\textit{Spatial reward.} Matching predicted boxes to references requires a fragile assignment step (e.g., Hungarian matching) under an IoU threshold, and it further breaks down when the two sides are annotated at inconsistent granularity: a single reference block may be predicted as several finer boxes (one-to-N), or several reference blocks may be covered by one prediction (N-to-one), leaving no unambiguous one-to-one correspondence. We therefore avoid box matching altogether and instead score localization at the level of per-category occupied area. For each category $k$, we merge all predicted boxes of that category into a single region $P^{k}$ by geometric union, and likewise form $G^{k}$ from the references. The spatial reward is the mean area Intersection-over-Union (mIoU)~\cite{chen2018encoder} over the categories that appear in the prediction or the reference, $K=K_{\mathrm{pred}}\cup K_{\mathrm{gt}}$:
\begin{equation}
r_{\mathrm{spatial}} = \frac{1}{\lvert K\rvert}\sum_{k\in K}\frac{\lvert P^{k}\cap G^{k}\rvert}{\lvert P^{k}\cup G^{k}\rvert},
\end{equation}
where $\lvert\cdot\rvert$ denotes area and $\lvert K\rvert$ the number of categories. A category present on only one side has an empty intersection and thus contributes $0$. The unions and intersections are evaluated exactly from the box geometry, so the spatial reward requires no rasterization or resolution parameter. This region-overlap formulation is \emph{assignment-free} and \emph{order-invariant} at the box level, and remains robust to mis-counting: spurious predictions enlarge the union while missed regions shrink the intersection, both lowering the score without any explicit box matching.

\textit{Textual reward.} The element contents are concatenated in the predicted reading order into a single Markdown string $s_{\mathrm{pred}}$ and scored against the reference $s_{\mathrm{gt}}$ by Edit Distance Similarity, $r_{\mathrm{text}}=\mathrm{EDS}(s_{\mathrm{pred}},s_{\mathrm{gt}})$. Because the concatenation follows reading order, this term additionally penalizes reading-order errors, so localization, content fidelity, and reading order are jointly covered by the two terms without a dedicated ordering reward. Using EDS also keeps the content signal consistent with the doc2md task.

\textit{Combination.} The two terms are linearly combined,
\begin{equation}
r_{\mathrm{doc2json}} = \lambda\,r_{\mathrm{spatial}} + (1-\lambda)\,r_{\mathrm{text}},
\end{equation}
with both components in $[0,1]$. We report the weight $\lambda$ in our experimental setup. Predictions that cannot be parsed into the element list receive zero reward. Unlike prior RLVR-based parsers that reward only flat textual output (Infinity-Parser~\cite{wang2025infinity} couples edit distance with paragraph-count and reading-order terms but supplies no localization signal, while olmOCR-2~\cite{poznanski2025olmocr2} scores unit tests on rendered text rather than on coordinate-bearing structured output), this reward acts directly on the structured prediction, separately crediting \emph{where} each element is and \emph{what} it contains.

\paragraph{Reward Suite over the Co-trained Objectives.}
We extend the same metric-as-reward principle across all eight co-trained objectives:
\begin{itemize}[leftmargin=1.0em, labelsep=0.5em]
    \item \textbf{Document parsing.} doc2md is rewarded by EDS and doc2json by the spatial--textual reward above.
    \item \textbf{Layout analysis.} Layout analysis, which predicts typed boxes without content, is rewarded by the spatial term $r_{\mathrm{spatial}}$ alone.
    \item \textbf{Table parsing.} We use the Tree-Edit-Distance-based Similarity (TEDS)~\cite{zhong2020image} for both table2html and table2md, reflecting the hierarchical structure of tables.
    \item \textbf{Math formula parsing.} We use the Character Detection Matching (CDM) metric~\cite{wang2025imagetexttransformingformula} for formula2latex, which compares rendered formulas and is therefore robust to syntactically different but visually equivalent LaTeX.
    \item \textbf{Chart parsing.} We adopt the Relative Mapping Similarity F1 (RMS-F1)~\cite{liu2023deplot} for chart2table, the mean Average Precision from the Structuring Chart-oriented Representation Metric (SCRM)~\cite{xia2023structchart} for chart2json, and EDS for chart2code, treating the generated plotting code as a string.
    \item \textbf{Chemical formula parsing.} For chem2smiles we use the Tanimoto similarity~\cite{tanimoto1958elementary} between the predicted and reference molecular fingerprints.
    \item \textbf{Document VQA.} For docvqa we use the Average Normalized Levenshtein Similarity (ANLS).
    \item \textbf{General multimodal understanding.} We apply RLVR only to the verifiable subset of this objective (samples that carry reference short answers), scored by ANLS. Open-ended samples are excluded from RL and retained only in SFT.
\end{itemize}

\subsubsection{Optimization}
\label{sec:rl_opt}

We optimize the policy with Group Relative Policy Optimization (GRPO)~\cite{shao2024deepseekmathpushinglimitsmathematical}, which dispenses with a learned value model. For each input $(I,c)$, the current policy samples a group of $G$ rollouts $\{y^{(i)}\}_{i=1}^{G}$, each scored by its task reward $r^{(i)}\in[0,1]$. The group-relative advantage is obtained by normalizing rewards within the group,
\begin{equation}
\hat{A}^{(i)} = \frac{r^{(i)} - \mathrm{mean}\big(\{r^{(j)}\}_{j=1}^{G}\big)}{\mathrm{std}\big(\{r^{(j)}\}_{j=1}^{G}\big)}.
\end{equation}
The policy is then updated by maximizing the standard PPO-style clipped surrogate with these advantages, regularized by a KL penalty toward a frozen reference policy. Because every reward is verifiable, the advantage reflects genuine task quality rather than a learned proxy.

To co-train heterogeneous tasks in a single run, a \emph{reward router} dispatches each rollout to the reward function of its task, while the within-group normalization places all rewards on a common $[0,1]$ scale, keeping advantages comparable across tasks within a mixed batch. The RL training set is constructed by randomly sampling 5\% of each task from Infinity-Doc2-5M, yielding roughly 220K examples. Its detailed composition and the optimization hyperparameters are reported in our experimental setup.

\section{Experiments}

\subsection{Implementation Details}

\begin{itemize}[
  leftmargin=1.0em,
  labelsep=0.5em
]
\item \textbf{Experimental Setup.} Our approach builds upon the Qwen3.5 architecture~\cite{qwen3.5}. We train Infinity-Parser2-Pro based on Qwen3.5-35B-A3B, and train Infinity-Parser2-Flash based on Qwen3.5-2B. All experiments are performed on a cluster of 64 NVIDIA H100 GPUs, utilizing PyTorch 2.10.0 and CUDA 12.8. We employ Megatron-LM-0.16.0~\cite{shoeybi2020megatronlmtrainingmultibillionparameter} for distributed model parallel training.

\item \textbf{Supervised Fine-Tuning (SFT) Stage.} Both variants are trained for one epoch with the ms-swift~\cite{zhao2025swift} framework on its Megatron-LM backend. We use a maximum sequence length of 32,768 with multimodal sequence packing, encode each page image into at most 16,384 visual tokens, and keep all modules (the ViT encoder, the vision--language aligner, and the LLM backbone) trainable. We adopt the Adam optimizer with a micro batch size of 1 and a global batch size of 64. The learning rate follows a cosine schedule that warms up over $3\%$ of the steps to a peak of $1 \times 10^{-5}$ and then decays to $1 \times 10^{-6}$. Training is conducted in non-thinking mode. The two variants differ only in parallelism. Infinity-Parser2-Flash uses tensor parallel size 2, pipeline parallel size 1, and context parallel size 1. The MoE-based Infinity-Parser2-Pro uses tensor parallel size 8, expert parallel size 4, pipeline parallel size 1, and context parallel size 1, with grouped-GEMM expert computation, shared-expert overlap, and an MoE auxiliary-loss coefficient of $1 \times 10^{-6}$. Sequence parallelism and full activation recomputation are enabled throughout.

\item \textbf{Reinforcement Learning (RL) Stage.} We refine both SFT checkpoints with GRPO~(Sec.~\ref{sec:rl_opt}) using the VeRL~\cite{sheng2025hybridflow} framework on its Megatron-LM backend, for one epoch with a learning rate of $1 \times 10^{-6}$. We draw $8$ rollouts per prompt with an asynchronous vLLM engine (generation tensor parallel size 4, GPU memory utilization $0.5$), apply a KL loss (low-variance estimator, coefficient $0.01$), and set the entropy coefficient to $0$. For the disentangled doc2json reward (Sec.~\ref{sec:str}), the spatial and textual terms are weighted by $\lambda = 0.3$ and $1-\lambda = 0.7$, respectively. The two variants differ in batch size, sequence budget, and parallelism. Infinity-Parser2-Flash uses a rollout batch size and mini-batch size of 32, maximum prompt and response lengths of 8,192 and 16,384, and tensor parallel size 2 with pipeline and context parallel sizes of 1. Infinity-Parser2-Pro uses a rollout batch size and mini-batch size of 48, maximum prompt and response lengths of 8,192, and tensor parallel size 2, context parallel size 2, expert parallel size 8, and pipeline parallel size 1, with MoE auxiliary- and z-loss coefficients of $1 \times 10^{-2}$ and $1 \times 10^{-3}$. Parameter, gradient, and optimizer-state CPU offloading is enabled throughout.

\item \textbf{Inference and Evaluation.} Our internal evaluation tools are built on top of the official lmms-eval~\cite{zhang2025lmms} toolkit and extend it with the additional document-parsing tasks and capability dimensions studied in this work, while reusing its standardized metric implementations. In the following tables, results marked with `*' are re-evaluated by us under this unified pipeline, while the remaining baseline numbers are cited from their original reports. When re-evaluating a baseline, we follow its officially recommended prompt template and output format, so that the comparisons reflect genuine capability differences rather than prompt or output-format mismatches. The default decoding parameters are set as follows: $temperature = 0.0$, $top\_p = 0.95$, $top\_k = 50$, $num\_beams = 1$, $min\_pixels = 2,048$, $max\_pixels = 16,777,216$, and $max\_new\_tokens = 32,768$.
\end{itemize}

\subsection{Evaluation on Document Structure Tasks}

We evaluate the end-to-end document parsing capability of Infinity-Parser2 on three public benchmarks, olmOCR-Bench~\cite{poznanski2025olmocr}, ParseBench~\cite{zhang2026parsebenchdocumentparsingbenchmark}, and OmniDocBench-v1.6~\cite{ouyang2025omnidocbench}, comparing against three families of systems, namely pipeline-based tools (e.g., Marker~\cite{paruchuri2024marker}), expert VLMs (e.g., Nanonets-OCR-s~\cite{nanonets2025ocrs} and LightOnOCR-2~\cite{taghadouini2026lightonocr}), and general-purpose VLMs (e.g., Qwen2.5-VL~\cite{bai2025qwen25vl}). As summarized in Table~\ref{tab:document_parsing_public}, both of our variants deliver leading accuracy across the suite. On olmOCR-Bench, Infinity-Parser2-Pro attains an overall score of $87.6$ and establishes a new state of the art, exceeding the strongest baseline dots.mocr~\cite{zheng2026multimodalocrparsedocuments} ($83.9$) by $3.7$ points and widely adopted systems such as PaddleOCR-VL-1.5~\cite{cui2025paddleocr} ($78.5$), DeepSeek-OCR-2~\cite{wei2026deepseek2} ($76.3$), and MinerU2.5~\cite{niu2025mineru2} ($75.2$) by even larger margins. On ParseBench, Infinity-Parser2-Pro again ranks first with $74.3$, surpassing the strongest competing expert system Chandra-OCR-2~\cite{datalab2026chandra} ($70.1$) by $4.2$ points and the leading general-purpose VLMs Gemini-3.1-Pro ($69.1$) and GPT-5.5 ($67.8$) by $5.2$ and $6.5$ points, together with other specialized systems such as PaddleOCR-VL-1.5 ($66.0$) and dots.mocr ($55.8$).

\begin{table*}[htbp]
\centering
\caption{Comprehensive evaluation of document parsing algorithms on olmOCR-Bench, ParseBench, and OmniDocBench-v1.6. The ``Release Date'' column reports each model's first public release. `*' denotes results evaluated by our internal evaluation tools.}
\label{tab:document_parsing_public}
\small
\begin{tabular}{l c c c c}
\toprule
\multirow{2}{*}{\textbf{Methods}} & \multirow{2}{*}{\textbf{Release Date}} & \textbf{olmOCR-Bench} & \textbf{ParseBench} & \textbf{OmniDocBench-v1.6} \\
\cmidrule(lr){3-3} \cmidrule(lr){4-4} \cmidrule(lr){5-5}
& & \textbf{Overall$\uparrow$} & \textbf{Overall$\uparrow$} & \textbf{Overall$\uparrow$} \\
\midrule

\multicolumn{5}{c}{\textbf{Based on Pipeline Tools}} \\
\midrule
Marker 1.10.1 & 2025-09-30 & $76.1 \pm 1.1$ & - & - \\
PaddleOCR-VL-1.5 & 2026-01-29 & $78.5 \pm 1.0$* & 66.0 & 94.87 \\
GLM-OCR & 2026-02-03 & $75.2 \pm 1.1$* & 29.6 & \textbf{95.15} \\
\midrule

\multicolumn{5}{c}{\textbf{Based on Expert VLMs}} \\
\midrule
Nanonets-OCR-s & 2025-06-12 & $64.5 \pm 1.1$ & - & - \\
MinerU2.5 & 2025-09-19 & $75.2 \pm 1.1$ & 45.9 & 92.98 \\
Hunyuan-OCR & 2025-11-25 & $74.7 \pm 1.0$* & - & 89.87 \\
olmOCR-2 & 2025-10-22 & $82.4 \pm 1.1$ & - & - \\
LightOnOCR-2 & 2026-01-19 & $83.2 \pm 0.9$ & 48.0 & - \\
DeepSeek-OCR-2 & 2026-01-27 & $76.3 \pm 1.0$ & 41.2 & 90.17 \\
Chandra-OCR-2 & 2026-03-17 & $83.1 \pm 0.9$ & 70.1 & - \\
dots.ocr & 2025-07-30 & - & - & 90.50 \\
dots.mocr & 2026-03-19 & $83.9 \pm 0.9$ & 55.8 & - \\
\midrule

\multicolumn{5}{c}{\textbf{Based on General VLMs}} \\
\midrule
GPT-5.2 & 2025-12-11 & - & - & 86.52 \\
GPT-5.5 & 2026-04-23 & - & 67.8 & - \\
Gemini-3-Pro & 2025-11-18 & - & - & 92.85 \\
Gemini-3.1-Pro & 2026-02-19 & $73.6 \pm 1.1$* & 69.1 & - \\
Qwen2.5-VL-7B & 2025-01-28 & $65.5 \pm 1.2$ & - & - \\
Qwen2.5-VL-72B & 2025-01-28 & $75.8 \pm 1.1$* & - & - \\
Qwen3.5-2B & 2026-03-02 & $77.2 \pm 1.0$* & 27.3 & - \\
Qwen3.6-35B-A3B & 2026-04-16 & - & 44.1 & - \\
\midrule

\multicolumn{5}{c}{\textbf{Ours}} \\
\midrule
Infinity-Parser-7B & 2025-10-20 & $82.5 \pm 1.0$ & - & - \\
Infinity-Parser2-Flash & 2026-05-11 & $86.0  \pm 0.8$ & 72.2 & 91.98 \\
Infinity-Parser2-Pro & 2026-05-11 & $\mathbf{87.6 \pm 0.8}$ & \textbf{74.3} & 93.95 \\
\bottomrule
\end{tabular}
\end{table*}

On OmniDocBench-v1.6, Infinity-Parser2-Pro reaches $93.95$, outperforming leading expert VLMs such as MinerU2.5 ($92.98$), dots.ocr~\cite{dotsocr2025} ($90.50$), and DeepSeek-OCR-2 ($90.17$), as well as every general-purpose VLM, including the strongest one, Gemini-3-Pro ($92.85$). Only the two leading pipeline systems, PaddleOCR-VL-1.5 ($94.87$) and GLM-OCR~\cite{zhipu2026glmocr} ($95.15$), score higher. Notably, although GLM-OCR tops OmniDocBench-v1.6, it scores only $29.6$ on ParseBench. This large gap arises because the two benchmarks target different capabilities: OmniDocBench chiefly rewards faithful full-page text and table reconstruction, whereas ParseBench additionally evaluates chart-data extraction, semantic formatting, and visual grounding, which GLM-OCR does not handle. Such divergences reflect genuine differences in capability coverage rather than evaluation artifacts, and they underscore the importance of assessing parsing quality across complementary benchmarks. The lightweight Infinity-Parser2-Flash is highly competitive across the suite, scoring $86.0$ on olmOCR-Bench and $72.2$ on ParseBench and thereby exceeding all baselines on both, while improving over our prior Infinity-Parser-7B~\cite{wang2025infinity} from $82.5$ to $86.0$ on olmOCR-Bench. These results indicate that the joint multi-task reinforcement learning recipe yields consistent gains in parsing accuracy, and that the Flash variant preserves most of this quality at substantially lower inference cost.

\begin{table*}[htbp]
\centering
\caption{Layout detection evaluation on DocLayNet, OmniDocBench-v1.5 and D$^4$LA. All results are evaluated by our internal evaluation tools.}
\label{tab:layout_analysis_public}
\small
\begin{tabular}{l c c c}
\toprule
\multirow{2}{*}{\textbf{Methods}} & \textbf{DocLayNet} & \textbf{OmniDocBench-v1.5} & \textbf{D$^4$LA} \\
\cmidrule(lr){2-2} \cmidrule(lr){3-3} \cmidrule(lr){4-4}
& \textbf{mIoU$\uparrow$} & \textbf{mIoU$\uparrow$} & \textbf{mIoU$\uparrow$}\\
\midrule
DocLayout-YOLO & 65.80 & 73.11 & 42.44 \\
PP-DocLayoutV2 & 70.89 & 75.17 & 51.10 \\
PP-DocLayoutV3 & 71.05 & 74.80 & 50.21 \\
\midrule
dots.ocr & \textbf{80.16} & \textbf{79.82} & 51.34 \\
MinerU2.5 & 67.74 & 76.28 & 51.62\\
DeepSeek-OCR-2 & 45.62 & 55.28 & 33.03 \\
\midrule
Qwen3.5-2B & 23.08 & 32.82 & 21.57 \\
Qwen3.5-35B-A3B & 53.62 & 64.98 & 50.11 \\
\midrule
Infinity-Parser2-Flash & 64.97 & 73.07 & 46.05 \\
Infinity-Parser2-Pro & 64.93 & 74.56 & \textbf{52.41} \\
\bottomrule
\end{tabular}
\end{table*}

In addition, we evaluate the layout analysis performance of Infinity-Parser2 on DocLayNet~\cite{pfitzmann2022doclaynet}, OmniDocBench-v1.5~\cite{ouyang2025omnidocbench}, and D$^4$LA~\cite{da2023vgt}. Prior work typically reports mean Average Precision (mAP), which matches predicted and reference boxes one-to-one under an IoU threshold. This becomes unreliable across these benchmarks, since they annotate document regions at inconsistent granularity and thus produce the one-to-N and N-to-one matches that motivate our assignment-free spatial reward (Section~\ref{sec:reward_design}). We therefore cast layout evaluation as a semantic image segmentation task and adopt the mean Intersection over Union (mIoU)~\cite{chen2018encoder}, which compares per-category occupied area and is invariant to how each region is partitioned into boxes. To further reconcile discrepancies in category definitions across models and benchmarks, we uniformly map all labels to five broad categories, namely textual, figure, table, formula, and other. Because this protocol departs from the mAP setting used in the original papers, the mIoU scores reported below are not directly comparable to those prior numbers. To keep the comparison fair, we re-evaluate every method (baselines and ours) under this identical protocol with our internal evaluation tools.

As reported in Table~\ref{tab:layout_analysis_public}, Infinity-Parser2 operates as a general-purpose end-to-end parser yet attains layout-detection quality competitive with dedicated detectors, though it still trails the strongest specialist, dots.ocr, on DocLayNet. On the challenging D$^4$LA benchmark, Infinity-Parser2-Pro achieves the best mIoU of $52.41$, surpassing strong baselines such as MinerU2.5~\cite{niu2025mineru2} ($51.62$), dots.ocr~\cite{dotsocr2025} ($51.34$), and PP-DocLayoutV2~\cite{paddle2025ppdoclayoutv2} ($51.10$). Nonetheless, D$^4$LA scores remain markedly lower than those on the other two benchmarks across all methods, as the dataset consists of challenging, blurry scanned documents whose ambiguous appearance makes element types prone to misclassification. On OmniDocBench-v1.5, it reaches $74.56$ mIoU, comparable to PP-DocLayoutV3~\cite{paddle2026ppdoclayoutv3} ($74.80$) and PP-DocLayoutV2 ($75.17$) and approaching MinerU2.5 ($76.28$). On DocLayNet, both variants remain competitive at roughly $65$ mIoU, on par with DocLayout-YOLO~\cite{zhao2024doclayoutyolo} ($65.80$), although dots.ocr retains a clear lead on DocLayNet ($80.16$) and OmniDocBench-v1.5 ($79.82$). Notably, our training recipe yields large gains over the underlying Qwen3.5 backbones~\cite{qwen3.5}. Infinity-Parser2-Flash lifts the Qwen3.5-2B baseline from $23.08$ to $64.97$ on DocLayNet and from $21.57$ to $46.05$ on D$^4$LA, and Infinity-Parser2-Pro improves the Qwen3.5-35B-A3B baseline by $11.31$ and $9.58$ mIoU on DocLayNet and OmniDocBench-v1.5, respectively.

\subsection{Evaluation on Element-level Parsing Tasks}

Element parsing evaluates the fundamental capability of vision-language models to recognize and interpret discrete components within a document, including text blocks, tables, mathematical formulas, charts, and chemical formulas. We benchmark this capability across five element categories: text recognition, table recognition~\cite{zhong2020image, zheng2021global}, mathematical formula recognition~\cite{wang2024unimernet}, chart parsing~\cite{kantharaj2022chart, masry2022chartqa, chen2024onechart}, and chemical formula recognition. For text recognition, we crop all text-related sub-images from the 1,355 images of OmniDocBench-v1.5 according to the layout detection labels and discard those with null annotations, yielding 17,148 block-level images. We compare both Infinity-Parser2-Flash and Infinity-Parser2-Pro against three families of baselines: pipeline OCR systems, general-purpose vision-language models, and task-specialized expert models.

Table~\ref{tab:element_parsing_public} reports text, table, and mathematical formula recognition, measured respectively by edit distance similarity (EDS), tree-edit-distance-based similarity (TEDS), and character detection matching (CDM). Infinity-Parser2-Pro achieves the best overall results across the three tasks. On table recognition it attains state-of-the-art accuracy, reaching 94.76 TEDS on PubTabNet and 98.88 TEDS on FinTabNet, ahead of the second-best Infinity-Parser2-Flash (92.41 and 98.51) and all other systems. On mathematical formula recognition it obtains the highest average CDM of 97.7, with the top scores on the SPE (99.4) and CPE (98.3) subsets. On text recognition it reaches 95.05 EDS, second only to GLM-OCR (95.60) and ahead of strong baselines such as PaddleOCR-VL-1.5 (94.97) and Qwen3.5-35B-A3B (93.30). It further surpasses the proprietary Gemini-3-Pro and GPT-5.2 on both PubTabNet TEDS and average CDM, and improves substantially over the first-generation Infinity-Parser-7B (80.26 EDS and 91.6 average CDM). The lighter Infinity-Parser2-Flash remains highly competitive, delivering 94.31 EDS, 96.5 average CDM, and the second-best TEDS on both table benchmarks.

\begin{table*}[htbp]
\centering
\caption{Element parsing evaluation of text recognition, table recognition and math formula recognition tasks. For each benchmark, the best result is in bold and the second best is underlined. `*' denotes results evaluated by our internal evaluation tools.}
\label{tab:element_parsing_public}
\small
\begin{tabular}{lcccccccc}
\toprule
\multirow{2}{*}{\textbf{Methods}} & \textbf{Text Rec. (EDS$\uparrow$)} & \multicolumn{2}{c}{\textbf{Table Rec. (TEDS$\uparrow$)}} & \multicolumn{5}{c}{\textbf{Math Formula Rec. (CDM$\uparrow$)}} \\
\cmidrule(lr){2-2} \cmidrule(lr){3-4} \cmidrule(lr){5-9}
 & \textbf{OmniDocBench-v1.5} & \textbf{PubTabNet} & \textbf{FinTabNet} & \textbf{SPE} & \textbf{SCE} & \textbf{CPE} & \textbf{HWE} & \textbf{Avg.} \\
\midrule
MinerU2.5 & 86.00 & 89.07 & 95.97 & 98.4 & \textbf{96.4} & 96.6 & 94.4 & 96.5 \\
olmOCR-2 & 91.98* & 84.02* & 88.10* & 94.9* & 84.0* & 88.0* & 86.4* & 88.3* \\
LightOnOCR-2 & 74.42* & 51.42* & 79.58* & 36.9* & 19.7* & 62.8* & 24.2* & 35.9* \\
PaddleOCR-VL-1.5 & 94.97* & 84.60 & 94.71* & \underline{99.1*} & 94.7* & \underline{97.1*} & 92.3* & 95.8* \\
DeepSeek-OCR-2 & 84.13* & 89.53* & 51.84* & 95.3* & 84.7* & 54.1* & 85.2* & 79.8* \\
GLM-OCR & \textbf{95.60*} & 85.20 & 92.56* & - & - & - & - & 96.5 \\
\midrule
GPT-5.2 & - & 84.40 & - & - & - & - & - & 90.5 \\
Gemini-3-Pro & - & 91.40 & - & - & - & - & - & 96.4 \\
Qwen2.5-VL-7B & 87.38* & 81.60 & 82.58 & 97.5* & 94.7* & 86.1* & 91.8* & 92.5* \\
Qwen3.5-2B & 91.67* & 76.33* & 78.95* & 98.8* & 95.7* & 92.8* & 95.3* & 95.7* \\
Qwen3.5-35B-A3B & 93.30* & 90.06* & 90.44* & \textbf{99.4*} & 95.1* & 96.8* & \textbf{97.1*} & \underline{97.1*} \\
\midrule
Infinity-Parser-7B & 80.26* & 91.82 & 95.92 & 97.9* & 87.2* & 89.9* & 91.4* & 91.6* \\
Infinity-Parser2-Flash & 94.31 & \underline{92.41} & \underline{98.51} & 98.2 & \underline{96.2} & 96.1 & 95.3 & 96.5 \\
Infinity-Parser2-Pro & \underline{95.05} & \textbf{94.76} & \textbf{98.88} & \textbf{99.4} & \underline{96.2} & \textbf{98.3} & \underline{96.7} & \textbf{97.7} \\
\bottomrule
\end{tabular}
\end{table*}

Table~\ref{tab:chart_parsing_public} reports chart parsing across three structured sub-tasks: Chart-to-Table on ChartQA (RMS-F1), Chart-to-JSON on ChartX-SE (AP@strict, AP@slight, AP@high), and Chart-to-Code on ChartMimic-direct-600 (execution rate, low-level and high-level similarity). Infinity-Parser2-Pro achieves the best Chart-to-JSON results among all evaluated models, with scores of 61.7, 68.9, and 73.7, surpassing the much larger Qwen3.5-35B-A3B (60.2, 67.8, 73.4). On Chart-to-Code it attains the highest high-level similarity (79.6) and the second-best execution rate (87.1), trailing only the chart-specialized ChartCoder (91.4). On Chart-to-Table it reaches 86.5 RMS-F1, the best among general-purpose parsers and approaching dedicated chart experts such as TinyChart-3B (93.8) and ChartAst (91.6). Infinity-Parser2-Flash offers a favorable accuracy-efficiency balance, delivering 80.5 RMS-F1 together with consistent Chart-to-JSON and Chart-to-Code scores.

\begin{table*}[htbp]
\centering
\caption{Comprehensive evaluation of chart parsing algorithms: Chart-to-Table is evaluated on the ChartQA dataset, Chart-to-JSON on the ChartX-SE dataset, and Chart-to-Code on the ChartMimic-direct-600 dataset. For each benchmark, the best result is in bold and the second best is underlined. `*' denotes results evaluated by our internal evaluation tools.}
\label{tab:chart_parsing_public}
\resizebox{\textwidth}{!}{
\begin{tabular}{lccccccc}
\toprule
\multirow{2}{*}{\textbf{Model}} & \textbf{Chart-to-Table} & \multicolumn{3}{c}{\textbf{Chart-to-JSON}} & \multicolumn{3}{c}{\textbf{Chart-to-Code}}\\
 \cmidrule(lr){2-2} \cmidrule(lr){3-5} \cmidrule(lr){6-8}
 & \textbf{RMS-F1$\uparrow$} & \textbf{AP@strict$\uparrow$} & \textbf{AP@slight$\uparrow$} & \textbf{AP@high$\uparrow$} & \textbf{Exec. Rate} & \textbf{Low-Level} & \textbf{High-Level} \\
\midrule
ChartAst & \underline{91.6} & 18.8 & 36.8 & 47.3 & - & - & - \\
TinyChart-3B & \textbf{93.8} & - & - & - & 42.5 & 26.3 & 25.9 \\
OneChart-0.2B & - & 44.6 & 53.1 & 59.7 & - & - & - \\
ChartCoder & - & - & - & - & \textbf{91.4} & \textbf{77.4} & 74.0 \\
\midrule
Hunyuan-OCR & 77.0* & - & - & - & - & - & - \\
olmOCR-2 & 71.0* & 48.2* & 57.4* & 64.9* & 64.9* & 43.2* & 52.1* \\
PaddleOCR-VL-1.5 & 86.2* & - & - & - & - & - & - \\
DeepSeek-OCR-2 & 49.7* & - & - & - & - & - & - \\
LightOnOCR-2 & 34.6* & - & - & - & - & - & - \\
\midrule
Qwen2.5-VL-7B & 64.3* & 50.3* & 57.4* & 64.3* & 72.1* & 41.9* & 41.6* \\
Qwen3.5-2B & 78.7* & 47.3* & 55.5* & 61.9* & 42.0* & 25.4* & 42.9* \\
Qwen3.5-35B-A3B & 85.2* & \underline{60.2*} & \underline{67.8*} & \underline{73.4*} & 85.8 & \underline{70.6} & \underline{77.8} \\
\midrule
Infinity-Parser-7B & 66.3* & 50.5* & 59.0* & 65.8* & - & - & - \\
Infinity-Parser2-Flash & 80.5 & 53.4 & 62.0 & 67.7 & 62.1 & 45.2 & 60.3 \\
Infinity-Parser2-Pro & 86.5 & \textbf{61.7} & \textbf{68.9} & \textbf{73.7} & \underline{87.1} & 70.1 & \textbf{79.6} \\
\bottomrule
\end{tabular}
}
\end{table*}

Table~\ref{tab:chem_formula_parsing} reports chemical formula parsing on the public CoSyn-Chemical and in-house ChemDraw-Bench benchmarks, evaluated by InChI exact-match accuracy, Tanimoto similarity, and valid SMILES rate. This task remains highly challenging, and the domain-specialized expert Decimer retains a clear lead on both benchmarks, far ahead on InChI and Tanimoto. On CoSyn-Chemical, the proprietary Gemini-3-Pro records the highest InChI of any general-purpose model (64.84, second only to Decimer overall), while Infinity-Parser2-Pro is the strongest open general-purpose model, ahead of holistic OCR systems such as OCRVerse~\cite{doctron2025ocrverse}, posting 53.91 InChI, 73.19 Tanimoto, and 86.72 valid SMILES and surpassing Qwen3.5-35B-A3B on both InChI (50.78) and valid SMILES (85.94) while matching its Tanimoto (73.19). On the in-house ChemDraw-Bench, which better exposes generalization, Infinity-Parser2-Pro is by far the strongest general-purpose model, attaining the second-best InChI (49.95) and Tanimoto (72.35) overall behind only Decimer and even ahead of the chemical specialist Img2Mol~\cite{clevert2021img2mol} (14.28 InChI), while substantially outperforming Qwen3.5-35B-A3B (8.15, 61.98) and DeepSeek-OCR-2 (8.10, 63.80). Infinity-Parser2-Flash also remains robust, particularly on ChemDraw-Bench with 34.30 InChI and 66.21 Tanimoto.

\begin{table*}[htbp]
\centering
\caption{Comprehensive evaluation of chemical formula parsing algorithms: the results are evaluated on public CoSyn-Chemical dataset and in-house ChemDraw-Bench dataset. For each benchmark, the best result is in bold and the second best is underlined. `*' denotes results evaluated by our internal evaluation tools.}
\label{tab:chem_formula_parsing}
\small
\begin{tabular}{lcccccc}
\toprule
\multirow{2}{*}{\textbf{Model}} & \multicolumn{3}{c}{\textbf{CoSyn-Chemical}} & \multicolumn{3}{c}{\textbf{ChemDraw-Bench}}\\
 \cmidrule(lr){2-4} \cmidrule(lr){5-7}
 & \textbf{Inchi.} & \textbf{Tanimoto} & \textbf{Valid smiles} & \textbf{Inchi.} & \textbf{Tanimoto} & \textbf{Valid smiles} \\
\midrule
Img2Mol & 52.34* & 70.65* & \textbf{98.44*} & 14.28* & 67.88* & \underline{98.98*} \\
Decimer & \textbf{86.72*} & \textbf{96.72*} & \underline{97.66*} & 65.32* & \textbf{93.21*} & \textbf{100.00*} \\
\midrule
OCRVerse & - & 54.70 & 89.10 & - & - & - \\
olmOCR-2 & - & 5.58* & 21.09* & - & 2.73* & 8.61* \\
DeepSeek-OCR-2 & 37.50* & 47.02* & 55.47* & 8.10* & 63.80* & 90.38* \\
\midrule
Gemini-3-Pro & \underline{64.84*} & 70.55* & 71.88* & - & - & - \\
Qwen2.5-VL-7B & 3.12* & 14.69* & 35.16* & - & 22.76* & 51.61* \\
Qwen3.5-2B & 15.62* & 41.48* & 69.53* & 0.51* & 16.53* & 30.97* \\
Qwen3.5-35B-A3B & 50.78* & \underline{73.19*} & 85.94* & 8.15* & 61.98* & 81.23* \\
\midrule
Infinity-Parser2-Flash & 39.06 & 63.34 & 83.59 & 34.30 & 66.21 & 74.69 \\
Infinity-Parser2-Pro & 53.91 & \underline{73.19} & 86.72 & \underline{49.95} & \underline{72.35} & 77.28 \\
\bottomrule
\end{tabular}
\end{table*}

\subsection{Evaluation on Reasoning and Generalization Tasks}

We further assess whether the document-parsing specialization of Infinity-Parser2 compromises its general multimodal competence, evaluating on benchmarks for reasoning and perception (AI2D~\cite{kembhavi2016diagram}, MathVista~\cite{lu2023mathvista}, MMBench-EN/CN~\cite{liu2024mmbench}, MMMU~\cite{yue2024mmmu}, MMStar~\cite{chen2024we}) and OCR and document understanding (OCRBench~\cite{liu2024ocrbench}, DocVQA~\cite{docvqa}, InfoVQA~\cite{infovqa}). As reported in Table~\ref{tab:general_mm_public}, many OCR-specialized systems attain parsing accuracy at the expense of general capability: DeepSeek-OCR-2 and Hunyuan-OCR~\cite{tencent2025hunyuanocr} degrade sharply, reaching only 37.66 and 65.38 on AI2D, respectively. In contrast, Infinity-Parser2 preserves the general understanding inherited from its Qwen3.5~\cite{qwen3.5} foundation. Infinity-Parser2-Pro is the strongest among open systems, ranking first or second on seven of the nine benchmarks and trailing only the much larger proprietary Gemini-3-Pro, while surpassing it on MMMU (61.89 vs. 56.00), DocVQA (96.43 vs. 93.68), and InfoVQA (86.26 vs. 85.24).

Relative to its Qwen3.5-35B-A3B base, Infinity-Parser2-Pro improves on six of the nine benchmarks, including MMMU (+8.56), MMBench-EN (+1.80), InfoVQA (+1.75), and DocVQA (+1.09), and is unchanged on OCRBench. The two exceptions are the reasoning-centric MathVista ($-$4.7) and MMStar ($-$3.36), where specialization toward document parsing exacts a measurable cost on pure mathematical and logical reasoning. This trade-off is amplified for the lightweight Infinity-Parser2-Flash, whose smaller backbone leaves less spare capacity: it gains on document-centric and perception benchmarks such as AI2D, MMBench, MMMU, DocVQA, and InfoVQA, but regresses more sharply on the reasoning benchmarks MathVista ($-$10.9) and MMStar ($-$6.87), alongside a smaller drop on OCRBench ($-$2.1). Overall, the joint optimization across eight co-trained objectives equips Infinity-Parser2 with state-of-the-art parsing ability while retaining most of its broad multimodal understanding, with the residual cost concentrated on reasoning-heavy benchmarks rather than causing broad forgetting across capabilities.

\begin{table*}[htbp]
\centering
\caption{Comprehensive evaluation of VLMs on general multimodal understanding benchmarks. All results are evaluated by our internal evaluation tools. Owing to the high API inference cost of Gemini-3-Pro, we set its reasoning effort to \texttt{low}. For each benchmark, the \textbf{best} result is in bold and the \underline{second best} is underlined.}
\label{tab:general_mm_public}
\resizebox{\textwidth}{!}{
\begin{tabular}{lccccccccc}
\toprule
\textbf{Methods} & \textbf{AI2D} & \textbf{MathVista} & \textbf{MMBench-EN} & \textbf{MMBench-CN} & \textbf{MMMU} & \textbf{MMStar} & \textbf{OCRBench} & \textbf{DocVQA} & \textbf{InfoVQA} \\
\midrule
Hunyuan-OCR & 65.38 & 43.4 & 51.80 & 46.13 & 32.22 & 43.21 & 83.0 & 88.35 & 64.76 \\
olmOCR-2 & 82.61 & 69.2 & 83.50 & 80.84 & 51.44 & 61.71 & 84.3 & 94.17 & 79.83\\
DeepSeek-OCR-2 & 37.66 & - & - & - & - & - & 47.2 & 43.42 & 22.07 \\
\midrule
Gemini-3-Pro & \textbf{91.87} & \textbf{81.8} & \textbf{90.29} & \textbf{90.98} & \underline{56.00} & \textbf{83.78} & \textbf{89.3} & 93.68 & \underline{85.24} \\
Qwen2.5-VL-7B & 82.90 & 69.1 & 82.73 & 80.50 & 50.89 & 62.19 & 84.8 & 95.19 & 81.25 \\
Qwen3.5-2B & 75.71 & 70.4 & 71.30 & 71.39 & 39.11 & 64.00 & 83.7 & 92.36 & 72.37 \\
Qwen3.5-35B-A3B & 87.69 & \underline{76.1} & 85.74 & 85.91 & 53.33 & \underline{73.02} & \underline{86.2} & \underline{95.34} & 84.51 \\
\midrule
Infinity-Parser-7B & 82.74 & 69.6 & 82.90 & 80.76 & 51.22 & 62.56 & 84.7 & 95.03 & 80.81 \\
Infinity-Parser2-Flash & 79.53 & 59.5 & 77.92 & 75.77 & 45.89 & 57.13 & 81.6 & 93.16 & 75.94 \\
Infinity-Parser2-Pro & \underline{88.89} & 71.4 & \underline{87.54} & \underline{86.43} & \textbf{61.89} & 69.66 & \underline{86.2} & \textbf{96.43} & \textbf{86.26} \\
\bottomrule
\end{tabular}
}
\end{table*}

\subsection{Ablation Studies}

\subsubsection{Ablation Study on Image Resolutions and Base Models}

\begin{figure*}[htbp]
    \centering
    \includegraphics[width=\textwidth]{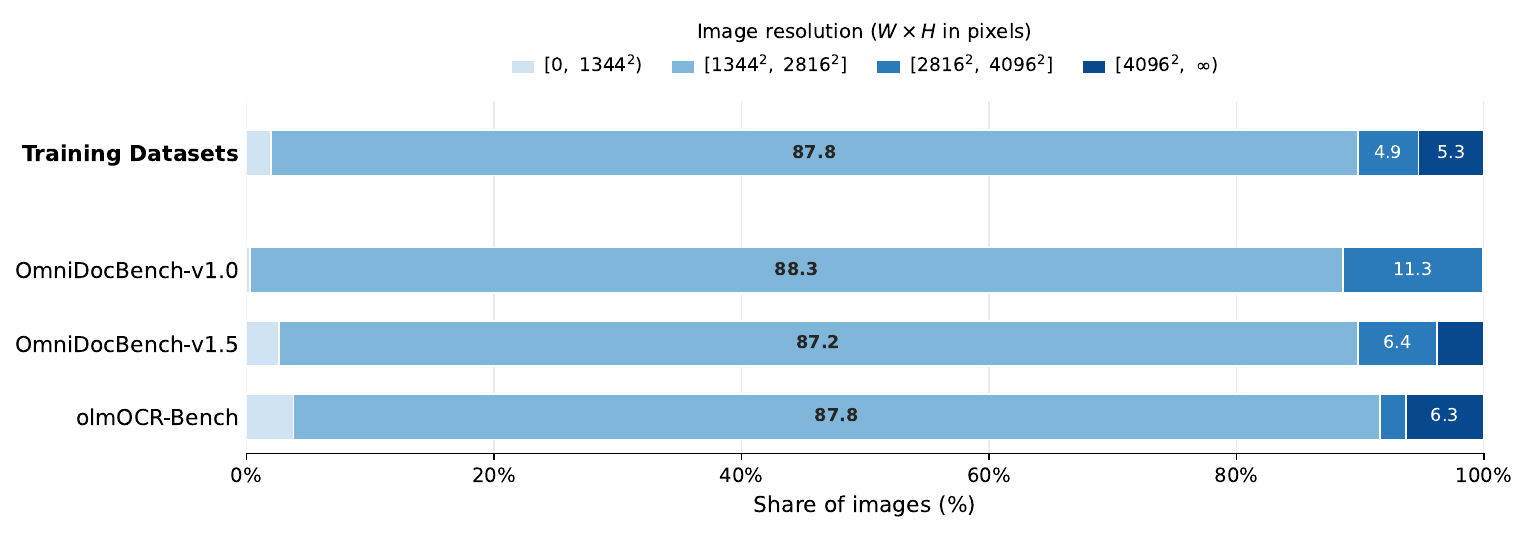}
    \caption{Image resolution distribution across benchmark datasets and training dataset. Across all datasets, roughly $87$--$88\%$ of images concentrate in the $[1344^2, 2816^2]$ band, and the training distribution closely aligns with the three evaluation benchmarks.}
    \label{fig:resolution_distribution}
\end{figure*}

We ablate two design choices for our lightweight parser, the training-time image resolution and the base model. The maximum image resolution permitted during training bounds the fidelity at which fine-grained content such as small fonts, dense tables, and compact formulas is preserved, but a higher ceiling also inflates the number of visual tokens and the context length required to encode a page. We study three resolution ceilings that span this trade-off, $1344 \times 1344$, $2816 \times 2816$, and $4096 \times 4096$, each paired with a correspondingly enlarged context length of $8192$, $16384$, and $32768$ tokens. These ceilings are informed by the image-resolution distribution of our data (Figure~\ref{fig:resolution_distribution}): across the training set and the three evaluation benchmarks, roughly $87$--$88\%$ of images fall within the $[1344^2, 2816^2]$ band and the training and evaluation distributions are closely aligned, yet a non-trivial tail exceeds $2816^2$, including more than $11\%$ of OmniDocBench-v1.0. Raising the ceiling thus preserves these large pages at native resolution rather than downsampling them. The resolution sweep is conducted on the Qwen3-VL-2B~\cite{qwen3vl2025} base model, and at the highest ceiling we further replace it with the stronger Qwen3.5-2B~\cite{qwen3.5} to quantify the gain attributable to the base model alone. To isolate these factors, all configurations are trained on the same corpus of 776K Pseudo-Labeled Web Documents, differing only in the base model, the resolution ceiling, and its paired context length.

Table~\ref{tab:ablation_image_resolutions} reports the results. On olmOCR-Bench, whose verification-based protocol probes localized targets rather than full-page reconstruction, the ceiling has only a marginal effect, with the overall score rising from $78.1$ to $78.4$. The benchmarks that score full-page fidelity benefit more clearly. Raising the ceiling from $1344 \times 1344$ to $4096 \times 4096$ improves the overall score on OmniDocBench-v1.5 from $87.59$ to $88.79$ and lowers the Chinese edit distance on OmniDocBench-v1.0 (Edit-ZH) from $0.181$ to $0.157$, whereas the English edit distance (Edit-EN) is already near saturation and remains essentially unchanged ($0.129$ to $0.128$). The improvement concentrates on Chinese documents, which are typographically denser and therefore most sensitive to the detail lost when high-resolution pages are downsampled to a lower ceiling. With the ceiling fixed at $4096 \times 4096$ and a $32768$-token context, replacing the Qwen3-VL-2B base with Qwen3.5-2B brings a further consistent gain, raising olmOCR-Bench from $78.4$ to $81.4$, lifting the OmniDocBench-v1.5 overall score from $88.79$ to $89.07$, and reducing Edit-EN from $0.128$ to $0.118$, while Edit-ZH remains at $0.157$. Guided by these results, we build on the Qwen3.5-2B base model and adopt the $4096 \times 4096$ resolution ceiling with a $32768$-token context as the default training configuration.

\begin{table}[htbp]
\centering
\caption{Ablation study on image resolutions and base models for Infinity-Parser2-Flash.}
\label{tab:ablation_image_resolutions}
\small
\begin{tabular}{c@{\hspace{5pt}}c@{\hspace{5pt}}c@{\hspace{5pt}}c@{\hspace{5pt}}c@{\hspace{5pt}}c@{\hspace{5pt}}c}
\toprule
\multirow{2}{*}{\textbf{Base Model}} & \multirow{2}{*}{\textbf{Image Resolution}} & \multirow{2}{*}{\textbf{Context Length}} & \textbf{olmOCR-Bench} & \textbf{OmniDocBench-v1.5} & \multicolumn{2}{c}{\textbf{OmniDocBench-v1.0}} \\
\cmidrule(lr){4-4} \cmidrule(lr){5-5} \cmidrule(lr){6-7}
 & & & Overall $\uparrow$ & Overall $\uparrow$ & Edit-EN $\downarrow$ & Edit-ZH $\downarrow$ \\
\midrule
\multirow{3}{*}{Qwen3-VL-2B} & $1344 \times 1344$ & 8192 & 78.1 & 87.59 & 0.129 & 0.181 \\
 & $2816 \times 2816$ & 16384 & 78.3 & 87.97 & 0.128 & 0.177 \\
 & $4096 \times 4096$ & 32768 & 78.4 & 88.79 & 0.128 & \textbf{0.157} \\
\midrule
Qwen3.5-2B & $4096 \times 4096$ & 32768 & \textbf{81.4} & \textbf{89.07} & \textbf{0.118} & \textbf{0.157} \\
\bottomrule
\end{tabular}
\end{table}

\subsubsection{Ablation Study on Output Format and Reward Design}

We further ablate two coupled design choices, the output representation and the RL reward, holding the Qwen3.5-2B base, the corpus of 776K Pseudo-Labeled Web Documents, and the $4096 \times 4096$ resolution ceiling with a $32768$-token context fixed as above. We take doc2json as the target format because it serializes per-element bounding boxes and structured fields together with text, which supports layout grounding and downstream structured extraction in a single pass, whereas doc2md emits text-only Markdown that discards this spatial information. This distinction also shapes the reward, as the disentangled reward of Sec.~\ref{sec:str} depends on the bounding boxes that only doc2json provides, so on top of the doc2json SFT model we compare its textual EDS term alone against the full spatial-and-textual reward.

\begin{table}[htbp]
\centering
\caption{Ablation study on output format (doc2md versus doc2json) and RL reward design for Infinity-Parser2-Flash. On DocLayNet, `-' indicates that the doc2md output contains no layout coordinates and thus cannot be evaluated.}
\label{tab:ablation_output_format_reward}
\small
\begin{tabular}{c@{\hspace{5pt}}c@{\hspace{5pt}}c@{\hspace{5pt}}c@{\hspace{5pt}}c@{\hspace{5pt}}c@{\hspace{5pt}}c@{\hspace{5pt}}c}
\toprule
\multirow{2}{*}{\textbf{Stage}} & \multirow{2}{*}{\textbf{Task}} & \multirow{2}{*}{\textbf{Reward}} & \textbf{DocLayNet} & \textbf{olmOCR-Bench} & \textbf{OmniDocBench-v1.5} & \multicolumn{2}{c}{\textbf{OmniDocBench-v1.0}} \\
\cmidrule(lr){4-4} \cmidrule(lr){5-5} \cmidrule(lr){6-6} \cmidrule(lr){7-8}
 & & & mIoU $\uparrow$ & Overall $\uparrow$ & Overall $\uparrow$ & Edit-EN $\downarrow$ & Edit-ZH $\downarrow$ \\
\midrule
\multirow{2}{*}{SFT} & doc2md & - & - & 81.9 & 90.58 & 0.114 & 0.140 \\
 & doc2json & - & 62.94 & 81.4 & 89.07 & 0.118 & 0.157 \\
\midrule
\multirow{2}{*}{RL} & doc2json & EDS & 64.11 & 83.4 & 90.77 & 0.129 & 0.156 \\
 & doc2json & mIoU + EDS & 65.63 & 83.8 & 90.88 & 0.122 & 0.144 \\
\bottomrule
\end{tabular}
\end{table}

Table~\ref{tab:ablation_output_format_reward} reports the results. At the SFT stage the two formats reach similar text fidelity, with doc2json trailing doc2md only slightly ($81.4$ against $81.9$ on olmOCR-Bench and $89.07$ against $90.58$ on OmniDocBench-v1.5), while doc2json additionally yields $62.94$ mIoU on DocLayNet, a layout capability that doc2md cannot produce at all. We therefore apply RL to doc2json rather than doc2md, since only doc2json admits the spatial reward and produces layout output. The textual EDS reward alone already lifts doc2json to $83.4$ on olmOCR-Bench, $90.77$ on OmniDocBench-v1.5, and $64.11$ mIoU on DocLayNet, and adding the spatial mIoU term improves every metric further, reaching $65.63$ mIoU, $83.8$ on olmOCR-Bench, and $90.88$ on OmniDocBench-v1.5 while lowering the Chinese edit distance from $0.157$ to $0.144$. The spatial term thus benefits text parsing as well as layout, which indicates that accurate localization feeds back into reading order and content fidelity, a gain that a text-only doc2md model cannot obtain. The final doc2json parser surpasses the doc2md SFT baseline on both olmOCR-Bench and OmniDocBench-v1.5, adds strong layout capability, and remains comparable on the OmniDocBench-v1.0 edit distances, so we adopt doc2json output trained with the mIoU + EDS reward as the default configuration.

\subsubsection{Ablation Study on Data Curation}
\label{sec:ablation_data_curation}

To assess the data iteration flywheel (Sec.~\ref{sec:flywheel}), we track benchmark performance across its successive rounds, grouped into two phases. The metric-improvement phase (Rounds 1--3) progressively enriches document-parsing and layout-analysis data to push the core parsing metrics, whereas the capability-expansion phase (Rounds 3--5) broadens task coverage by adding element-parsing data and then reasoning and generalization data, with Round 3 as the hinge between the two. Newly curated data are appended to the cumulative corpus at every round rather than replacing it, so each round isolates the marginal contribution of the data it introduces.

\begin{table*}[htbp]
\centering
\caption{Ablation on the metric-improvement rounds (Rounds 1--3), which progressively add document-parsing and layout-analysis data.}
\label{tab:ablation_data_curation_structure}
\resizebox{0.62\textwidth}{!}{
\begin{tabular}{llcc}
\toprule
\textbf{Round} & \textbf{Cumulative training data} & \textbf{olmOCR-Bench} & \textbf{DocLayNet} \\
\midrule
Round 1 & Public \& manually-labeled data & 28.4 & 49.03 \\
Round 2 & + Pseudo-labeled web data & 83.3 & 62.48 \\
Round 3 & + Synthesized data & 85.3 & 62.31 \\
\bottomrule
\end{tabular}
}
\end{table*}

\begin{table*}[htbp]
\centering
\caption{Ablation on the capability-expansion rounds (Rounds 3--5). From the Round 3 baseline, we add element-parsing data and then reasoning and generalization data. `-' denotes no valid output before the corresponding data is added.}
\label{tab:ablation_data_curation_element_reasoning}
\resizebox{\textwidth}{!}{
\begin{tabular}{llccccccc}
\toprule
\textbf{Round} & \textbf{Configuration} & \textbf{olmOCR-Bench} & \textbf{DocLayNet} & \textbf{UniMERNet} & \textbf{Chart2Table} & \textbf{CoSyn-Chemical} & \textbf{InfoVQA} & \textbf{MMBench-EN} \\
\midrule
Round 3 & Baseline & 85.3 & 62.31 & - & 17.39 & - & 21.71 & 62.11 \\
Round 4 & + element parsing datasets & 84.7 & 61.71 & 96.70 & 79.69 & 63.54 & 52.98 & 69.07 \\
Round 5 & + reasoning and generalization datasets & 84.3 & 60.81 & 96.79 & 79.96 & 64.25 & 75.73 & 77.66 \\
\bottomrule
\end{tabular}
}
\end{table*}

\textbf{Metric-improvement rounds.} Table~\ref{tab:ablation_data_curation_structure} reports the first phase. Trained only on public and manually-labeled data, Round 1 attains 28.4 on olmOCR-Bench~\cite{poznanski2025olmocr} and 49.03 on DocLayNet~\cite{pfitzmann2022doclaynet}, reflecting the limited scale and coverage of readily available supervision. Incorporating the pseudo-labeled web data mined and annotated by the flywheel (Round 2) produces the dominant gain, lifting olmOCR-Bench by 54.9 points to 83.3 and DocLayNet by 13.45 points to 62.48, which confirms that large-scale, weakness-targeted mining of real documents with expert-model annotation is the principal driver of core parsing accuracy. Adding synthesized documents (Round 3) yields a further 2.0-point improvement on olmOCR-Bench to 85.3 and leaves DocLayNet essentially unchanged (62.31), a modest headline effect consistent with synthesis targeting the rare layouts and long-tail elements that aggregate benchmarks under-represent.

\textbf{Capability-expansion rounds.} Table~\ref{tab:ablation_data_curation_element_reasoning} reports the second phase, starting from the Round 3 model. This baseline parses documents and analyzes layouts well (85.3 and 62.31) but lacks element-level and reasoning competence: it yields no valid output on UniMERNet~\cite{wang2024unimernet} or CoSyn-Chemical, scores only 17.39 on Chart2Table, and remains weak on InfoVQA~\cite{infovqa} (21.71) and MMBench-EN~\cite{liu2024mmbench} (62.11). Adding element-parsing data (Round 4) unlocks these capabilities from near-zero baselines, reaching 96.70 on UniMERNet and 63.54 on CoSyn-Chemical, raising Chart2Table more than fourfold to 79.69, and transferring to document-centric understanding (InfoVQA 52.98, MMBench-EN 69.07). Adding reasoning and generalization data (Round 5) then raises InfoVQA to 75.73 and MMBench-EN to 77.66 while element-parsing scores hold steady. Crucially, this expansion costs only 1.0 point on olmOCR-Bench (85.3 to 84.3) and 1.5 on DocLayNet (62.31 to 60.81), a minor trade-off under multi-task generalization that is far outweighed by the breadth of capabilities gained.

The two-phase progression validates the flywheel end to end. Each capability emerges precisely when its targeted data enters the corpus and is retained as later rounds add new objectives, so the pipeline first drives the core parsing metrics to a strong level through large-scale real-document mining and synthesis, then broadens Infinity-Parser2 into a general document parser covering formulas, charts, chemical formulas, and document reasoning, with negligible regression on its original strengths.

\subsubsection{Ablation Study on Training Strategies}

We first compare supervised fine-tuning (SFT) against the pretrained baseline. As shown in Table~\ref{tab:ablation_training_strategies}, SFT delivers large and consistent gains across both model sizes. For Infinity-Parser2-Flash, it raises olmOCR-Bench from $77.2$ to $85.5$, DocLayNet from $23.08$ to $63.29$, CoSyn-Chemical from $41.48$ to $65.48$, and Chart2Table from $78.70$ to $80.30$, while UniMERNet, InfoVQA, and MMBench-EN also rise to $96.75$, $74.94$, and $75.68$. For Infinity-Parser2-Pro, SFT lifts olmOCR-Bench from $69.8$ to $86.7$, DocLayNet from $53.62$ to $64.41$, and Chart2Table from $85.20$ to $88.43$, with only a modest drop on CoSyn-Chemical ($73.19$ to $71.17$) and an essentially flat MMBench-EN ($85.74$ to $85.65$). These results confirm that task-specific fine-tuning is essential for document parsing. Within SFT, opening the vision encoder rather than freezing it further improves document-centric parsing on Infinity-Parser2-Flash (olmOCR-Bench $84.3$ to $85.5$, DocLayNet $60.81$ to $63.29$) at a modest cost on general understanding (MMBench-EN $77.66$ to $75.68$), so we adopt the open-encoder configuration as the SFT default.

Building on the open-encoder SFT model, we then evaluate multi-task reinforcement learning (RL). For Infinity-Parser2-Flash, RL improves most benchmarks over SFT, lifting olmOCR-Bench to $86.0$, DocLayNet to $64.97$, and Chart2Table to $80.49$, and it recovers and surpasses the general multimodal performance lost during SFT, with InfoVQA and MMBench-EN rising to $75.94$ and $77.92$. The two exceptions are CoSyn-Chemical, which falls from $65.48$ to $63.34$, and a marginal decline on UniMERNet ($96.75$ to $96.51$), pointing to a trade-off on the most specialized recognition tasks at the smaller scale. This specialized-task trade-off disappears for the larger Infinity-Parser2-Pro, where RL improves both UniMERNet ($97.51$ to $97.66$) and CoSyn-Chemical ($71.17$ to $73.19$) over SFT, alongside olmOCR-Bench ($86.7$ to $87.6$), DocLayNet ($64.41$ to $64.93$), InfoVQA ($86.12$ to $86.26$), and MMBench-EN ($85.65$ to $87.54$), with the only regression on Chart2Table ($88.43$ to $86.50$). The net gains across both parsing and general multimodal understanding at the two scales lead us to adopt multi-task RL as the final training stage for both model sizes.

\begin{table*}[htbp]
\centering
\caption{Ablation on training strategies, comparing SFT (frozen or open ViT) and multi-task RL across seven benchmarks. The best result within each model group is in \textbf{bold}. The baselines for Infinity-Parser2-Flash and Infinity-Parser2-Pro are Qwen3.5-2B and Qwen3.5-35B-A3B, respectively.}
\label{tab:ablation_training_strategies}
\resizebox{\textwidth}{!}{
\begin{tabular}{llccccccc}
\toprule
\textbf{Model} & \textbf{Configuration} & \textbf{olmOCR-Bench} & \textbf{DocLayNet} & \textbf{UniMERNet} & \textbf{Chart2Table} & \textbf{CoSyn-Chemical} & \textbf{InfoVQA} & \textbf{MMBench-EN} \\
\midrule
\multirow{4}{*}{Infinity-Parser2-Flash} & Baseline & 77.2 & 23.08 & 95.70 & 78.70 & 41.48 & 72.37 & 71.30 \\
 & SFT (Freeze ViT) & 84.3 & 60.81 & \textbf{96.79} & 79.96 & 64.25 & 75.73 & 77.66 \\
 & SFT (Open ViT) & 85.5 & 63.29 & 96.75 & 80.30 & \textbf{65.48} & 74.94 & 75.68 \\
 & Multi-Task RL & \textbf{86.0} & \textbf{64.97} & 96.51 & \textbf{80.49} & 63.34 & \textbf{75.94} & \textbf{77.92} \\
\midrule
\multirow{3}{*}{Infinity-Parser2-Pro} & Baseline & 69.8 & 53.62 & 97.10 & 85.20 & \textbf{73.19} & 84.51 & 85.74 \\
 & SFT (Open ViT) & 86.7 & 64.41 & 97.51 & \textbf{88.43} & 71.17 & 86.12 & 85.65 \\
 & Multi-Task RL & \textbf{87.6} & \textbf{64.93} & \textbf{97.66} & 86.50 & \textbf{73.19} & \textbf{86.26} & \textbf{87.54} \\
\bottomrule
\end{tabular}
}
\end{table*}

\subsection{Inference Speed}

\begin{table}[htbp]
\centering
\small
\caption{Inference performance comparison of specialized VLMs and Infinity-Parser2 across different backends. `TP' denotes tensor-parallel, and `C' means concurrency.}
\label{tab:inference_speed}
\begin{tabular}{lccccccc}
\toprule
\textbf{Model} & \textbf{Size} & \textbf{Hardware} & \textbf{Task} & \textbf{Config} & \textbf{Tokens / Page} & \textbf{Tokens / Sec.} & \textbf{Sec. / Page} \\
\midrule
GPT-5.2 & - & API & doc2md & C=1 & 1330.49 & 35.12 & 37.51 \\
Gemini-3.1-Pro & - & API & doc2md & C=1 & 1038.08 & 93.84 & 10.73 \\
\midrule
MinerU2.5 & 1.2B & H100 & doc2md & TP=1, C=1 & 1042.11 & 296.38 & 3.45 \\
PaddleOCR-VL-1.5 & 0.9B & H100 & doc2md & TP=1, C=1 & 1842.04 & 700.04 & 2.48 \\
DeepSeek-OCR-2 & 3B & H100 & doc2md & TP=1, C=1 & 1056.67 & 439.67 & 2.35 \\
\midrule
\multirow{2}{*}{Infinity-Parser-7B} & \multirow{2}{*}{7B} & \multirow{2}{*}{H100} & doc2md & TP=1, C=1 & 885.22 & 114.89 & 7.44 \\
 &  &  & doc2md & TP=2, C=8 & 882.24 & 441.12 & 2.18 \\
\midrule
\multirow{2}{*}{Infinity-Parser2-Flash} & \multirow{2}{*}{2B} & \multirow{2}{*}{H100} & doc2json & TP=1, C=1 & 1510.33 & 244.30 & 6.06 \\
 & & & doc2json & TP=2, C=8 & 1546.99 & 1624.22 & 0.95 \\
\midrule
\multirow{2}{*}{Infinity-Parser2-Pro} & \multirow{2}{*}{35B-A3B} & \multirow{2}{*}{H100} & doc2json & TP=2, C=1 & 1498.72 & 173.58 & 8.48 \\
 & & & doc2json & TP=2, C=8 & 1500.48 & 703.71 & 2.13 \\
\bottomrule
\end{tabular}
\end{table}

We assess the inference efficiency of the Infinity-Parser2 family, comprising the 2B Infinity-Parser2-Flash and the 35B-A3B Infinity-Parser2-Pro, against three groups of baselines: the proprietary GPT-5.2 and Gemini-3.1-Pro APIs, the open specialized parsers MinerU2.5, PaddleOCR-VL-1.5, and DeepSeek-OCR-2, and our previous-generation Infinity-Parser-7B. All open-weight models are served with vLLM on H100 GPUs and benchmarked on an in-house set of business financial documents, with server startup overhead excluded. The external baselines are profiled under single-stream serving (TP=1, C=1), whereas the Infinity-Parser and Infinity-Parser2 models are also measured under a batched configuration (TP=2, C=8) that reflects concurrent production load. Note that Infinity-Parser2's doc2json output, which serializes bounding-box coordinates and structured fields alongside text, reaches roughly 1,510 tokens per page, compared with nearly 1,050 for doc2md models such as MinerU2.5 and DeepSeek-OCR-2. PaddleOCR-VL-1.5 emits even more, about 1,840 tokens per page, owing mainly to the styling tokens such as border and alignment in its HTML table output.

As reported in Table~\ref{tab:inference_speed}, under the batched configuration (TP=2, C=8), Infinity-Parser2-Flash reaches 1,624 tokens/sec and 0.95 sec/page despite its longer doc2json output. This represents a 3.68x throughput gain over the previous-generation Infinity-Parser-7B (441 tokens/sec, 2.18 sec/page) at the same configuration. The larger Infinity-Parser2-Pro (35B-A3B) sustains 704 tokens/sec and 2.13 sec/page under this configuration, a higher-capacity option kept efficient by its 3B active parameters. The proprietary APIs are far slower, at 10.73 sec/page (Gemini-3.1-Pro) and 37.51 sec/page (GPT-5.2), leaving them impractical for high-volume document parsing.

\subsection{Evaluation on Real-World Document Information Extraction}

To assess the practical efficacy of Infinity-Parser2 in real-world downstream applications, we evaluate it on Financial Information Extraction (FinIE), the task of transcribing the three primary financial statements in full from unstructured annual reports, capturing every line item (e.g., \textit{Total Assets}, \textit{Net Profit}) with its attributes. Financial reports mainly drawn from the stock market form a particularly demanding setting for document intelligence, owing to their frameless, complex layouts, and high density of tabular data. Because annual reports often span hundreds of pages, we adopt a modular parse-locate-extract pipeline instead of feeding entire documents to a language model. To isolate the contribution of document parsing, we fix the locator (Inf-Retriever-v1~\cite{infly-ai_2025}) and the extractor (DeepSeek-V4-Flash~\cite{deepseekai2026deepseekv4}), and vary only the parser that converts each page into structured text. The full task definition and the three-stage pipeline are detailed in Appendix~\ref{appendix:fie}.

\begin{table*}[htbp]
\centering
\small
\caption{Performance comparison on the Financial Information Extraction (FinIE) task. All systems share the same page locator and information extractor, and differ only in the document parser. The \textbf{best} reported result is shown in bold. Note that `*' indicates that Infinity-Parser2-Flash is further finetuned on our internal financial table parsing datasets.}
\label{tab:finie}
\resizebox{\textwidth}{!}{
\begin{tabular}{l ccc ccc ccc ccc}
\toprule
\multirow{2}{*}{\textbf{Method}} & \multicolumn{3}{c}{\textbf{Overall}} & \multicolumn{3}{c}{\textbf{Income Statement}} & \multicolumn{3}{c}{\textbf{Balance Sheet}} & \multicolumn{3}{c}{\textbf{Cash Flow Statement}} \\
\cmidrule(lr){2-4} \cmidrule(lr){5-7} \cmidrule(lr){8-10} \cmidrule(lr){11-13}
 & Precision & Recall & F1 & Precision & Recall & F1 & Precision & Recall & F1 & Precision & Recall & F1 \\
\midrule
MinerU2.5 & 91.20 & 92.10 & 91.50 & 80.90 & 84.90 & 82.60 & 96.10 & 95.00 & 95.50 & 96.70 & 96.20 & 96.50 \\
PaddleOCR-VL-1.5 & 93.07 & 94.83 & 93.73 & 86.83 & 93.10 & 89.40 & 96.13 & 95.33 & 95.67 & 96.23 & 95.93 & 96.13 \\
DeepSeek-OCR-2 & 89.77 & 91.03 & 90.23 & 80.07 & 84.87 & 82.03 & 94.10 & 93.47 & 93.73 & 95.10 & 94.73 & 94.90 \\
\midrule
Infinity-Parser2-Flash* & \textbf{95.70} & \textbf{96.05} & \textbf{95.75} & \textbf{88.55} & \textbf{91.35} & \textbf{89.50} & \textbf{99.25} & \textbf{97.55} & \textbf{98.35} & \textbf{99.40} &
  \textbf{99.20} & \textbf{99.30} \\
\bottomrule
\end{tabular}
}
\end{table*}

Table~\ref{tab:finie} reports the end-to-end results. Infinity-Parser2-Flash*, our Flash variant further finetuned on in-house financial tables, attains the highest overall F1 of $95.75\%$ and ranks first on every reported metric. The strongest baseline is PaddleOCR-VL-1.5~\cite{cui2025paddleocr} ($93.73\%$ overall F1), followed by MinerU2.5~\cite{niu2025mineru2} ($91.50\%$) and DeepSeek-OCR-2~\cite{wei2026deepseek2} ($90.23\%$). The Income Statement is the most challenging category for every parser, with the lowest scores throughout: Infinity-Parser2-Flash* leads at $89.50\%$ F1, against $89.40\%$ for PaddleOCR-VL-1.5 and roughly $82\%$ for both MinerU2.5 and DeepSeek-OCR-2. On the more regular Balance Sheet and Cash Flow Statement it approaches near-perfect extraction, reaching $98.35\%$ and $99.30\%$ F1. Accurate extraction of these line items hinges on faithful table structure recognition: a parser must align each numerical value with its row header (the financial concept) and column header (the reporting date). Trained with a structure-aware doc2json reward (Sec.~\ref{sec:str}), Infinity-Parser2-Flash* preserves high structural fidelity in dense and frameless tables, so that the retriever locates the correct context and the extractor returns the precise value. Baseline parsers more often merge cells or break frameless tables, producing spurious associations between values and dates that propagate to the final output. These results indicate that Infinity-Parser2 not only excels on standard parsing benchmarks but also serves as a robust front end for complex, high-stakes downstream applications.

\section{Limitations}

Although Infinity-Parser2 achieves strong performance across diverse benchmarks, several limitations remain and point to directions for future work. First, its training data are predominantly Chinese and English, mirroring the bilingual composition of Infinity-Doc2-5M, so parsing fidelity declines on documents written in other languages or containing substantial multilingual mixing. These data also incorporate pseudo-labels produced by expert models, and despite confidence-based and rule-based filtering, a degree of residual annotation noise inevitably remains. Second, accuracy degrades on visually complex charts with dense or heavily overlapping series, and on pages containing multi-oriented elements such as tables rotated at arbitrary angles, where structural and recognition errors become more frequent. Third, although the model faithfully reconstructs textual content and document structure, it does not preserve fine-grained inline formatting such as bold, italic, and strikethrough. Finally, its multimodal instruction-following remains limited, and the model may not reliably execute complex, multi-step visual instructions.

\section{Conclusion}

In this paper, we present Infinity-Parser2, an open-source large multimodal model that couples controllable data synthesis with multi-task reinforcement learning for end-to-end document parsing. Our contributions are threefold. First, we build a scalable synthesis engine that pairs a controllable rendering framework with an iterative refinement loop, and use it to construct and publicly release Infinity-Doc2-5M, a 5-million-sample bilingual corpus annotated with element bounding boxes, canonical content forms (Markdown, HTML, LaTeX, SMILES, and structured charts), and full-page reading order. Second, we introduce a verifiable, multi-task reward system that drives Joint Reinforcement Learning across eight co-trained objectives, spanning document parsing, layout analysis, table parsing, math formula parsing, chart parsing, chemical formula parsing, document VQA, and general multimodal understanding, unifying perception, structure, and reasoning within a single optimization signal. Third, we release two variants under a shared architecture: Infinity-Parser2-Flash, which attains a 3.68$\times$ throughput gain over Infinity-Parser-7B for low-latency inference, and Infinity-Parser2-Pro, engineered for precision-critical settings. Extensive experiments confirm the effectiveness of this design: Infinity-Parser2-Pro reaches state-of-the-art results of 87.6\% on olmOCR-Bench and 74.3\% on ParseBench, surpassing DeepSeek-OCR-2, PaddleOCR-VL-1.5, and MinerU2.5, while generalizing strongly to charts, chemical formulas, and document VQA.

Looking ahead, beyond the specific limitations of the current system, we see a broader trajectory opened by this work. The data iteration flywheel and the extensible synthesis engine together form a self-improving data foundation: because new modalities and notations are added simply by registering a renderer, the same pipeline can grow to cover richer document elements and steadily close the long-tail gaps surfaced in each iteration. We further aim to advance from page-level parsing toward document-level and cross-document understanding, coupling faithful structural extraction with downstream reasoning so that Infinity-Parser2 can serve as a reliable perception layer for agentic workflows over real-world documents. Together with broader language and layout coverage, these directions define our path toward increasingly general and dependable document understanding.

\section{Contributions}

\textbf{Contributors:} Zuming Huang, Jun Huang, Kexuan Ren, Baode Wang, Weizhen Li, Jianming Feng, Yu Wang, Yichen Yao, Shijun Lin, Yige Tang, Cheng Peng, Weidi Xu.

\textbf{Supervisors:} Wei Chu, Yinghui Xu, Yuan Qi.

\bibliographystyle{unsrt}  
\bibliography{latex/custom}

\appendix
\newpage
\section{Appendix}

\subsection{Evaluation Benchmarks}

\subsubsection*{Document Structure Tasks}

\begin{itemize}[
  leftmargin=1.0em,
  labelsep=0.5em
]
    \item \textbf{olmOCR-Bench}~\cite{poznanski2025olmocr} evaluates OCR and document parsing systems through simple, deterministic binary unit tests, analogous to software testing, enabling fair and transparent comparison across methods. It comprises 1,400 PDF pages and over 7,000 unit-test cases that span a broad range of document types and layout structures, providing a rigorous testbed for the robustness and generalization of document extraction systems.

    \item \textbf{ParseBench}~\cite{zhang2026parsebenchdocumentparsingbenchmark} targets document parsing for agentic workflows, emphasizing semantic correctness over surface text similarity. It contains roughly 2,000 human-verified pages from enterprise documents in insurance, finance, and government, organized along five capability dimensions: table structure, chart data, content faithfulness, semantic formatting, and visual grounding, each assessed with task-specific metrics. The benchmark therefore measures whether parsed outputs preserve the structure and meaning required for downstream automated decisions.

    \item \textbf{OmniDocBench}~\cite{ouyang2025omnidocbench} is a benchmark for comprehensive document parsing evaluation, with high-quality annotations across nine document sources ranging from academic papers and textbooks to more challenging handwritten notes and densely typeset newspapers. It supports multi-level evaluation, from end-to-end parsing to task-specific and attribute-level analysis, and covers 19 layout categories and 15 attribute labels for fine-grained comparison. Across this work we report three released versions of this benchmark: end-to-end document parsing uses OmniDocBench-v1.6, while the remaining evaluations use OmniDocBench-v1.5 and OmniDocBench-v1.0. These versions differ in document coverage and annotation refinements, so scores are comparable only within the same version.

    \item \textbf{DocLayNet}~\cite{pfitzmann2022doclaynet} is a large-scale benchmark for document layout analysis that addresses the limited layout diversity of earlier datasets. It provides over 80,000 manually annotated PDF pages from diverse sources, each labeled with bounding boxes across 11 layout categories, and includes a multiply annotated subset that establishes a human-agreement upper bound on performance.

    \item \textbf{D$^4$LA}~\cite{da2023vgt} is a dataset for document layout analysis covering 12 document categories and 27 layout types. It comprises 11,092 document images in total, of which 2,224 form the test set.

    \item \textbf{OmniDocBench Layout Subset}~\cite{ouyang2025omnidocbench} reuses the layout-region annotations of OmniDocBench-v1.5 to evaluate layout detection. Models predict bounding boxes for elements such as text, titles, tables, figures, and formulas, and accuracy is measured by mean intersection-over-union (mIoU) against the ground-truth regions.

\end{itemize}

\subsubsection*{Element-level Parsing Tasks}

\begin{itemize}[
  leftmargin=1.0em,
  labelsep=0.5em
]
    \item \textbf{OmniDocBench Text Subset}~\cite{ouyang2025omnidocbench} evaluates block-level text recognition. We crop all text-related regions from the 1,355 images of OmniDocBench-v1.5 according to their layout labels and discard null annotations, yielding 17,148 block-level images whose recognition quality is measured by edit distance similarity (EDS).

    \item \textbf{PubTabNet}~\cite{pubtabnet} is a large-scale benchmark for image-based table recognition, automatically built by aligning the XML and PDF representations of scientific articles from PubMed Central. It provides 500,777 training, 9,115 validation, and 9,138 test table images paired with structured HTML annotations of table structure and cell content. We evaluate on the validation split, measured by the Tree-Edit-Distance-based Similarity (TEDS) metric, which compares predicted and ground-truth tables as trees to jointly account for structural and content errors.

    \item \textbf{FinTabNet}~\cite{fintabnet} is a benchmark for table detection and cell-structure recognition in complex financial documents, drawn from the annual reports of S\&P 500 companies with HTML annotations of hierarchical cell structure. Its 112,887 tables are split into 91,596 training, 10,635 validation, and 10,656 test instances, and we evaluate on the validation split. It captures table styles that are common in practice yet underrepresented in scientific datasets.

    \item \textbf{UniMERNet}~\cite{wang2024unimernet} is a benchmark for mathematical expression recognition under real-world conditions. Its training set UniMER-1M offers over one million image-LaTeX pairs, while the evaluation set UniMER-Test contains 23,789 expressions across four categories: simple printed expressions (SPE), complex printed expressions (CPE), screen-capture expressions (SCE), and handwritten expressions (HWE).

    \item \textbf{ChartQA}~\cite{masry2022chartqa} is a chart-understanding benchmark of real-world charts, comprising 9.6K human-written and 23.1K machine-generated question-answer pairs that emphasize visual and arithmetic reasoning over chart content. In our evaluation it serves as the \textbf{Chart2Table} benchmark, where a model extracts the underlying data table from a chart image, scored by relative-mapping-similarity F1 (RMS-F1).

    \item \textbf{ChartX-SE}~\cite{chen2024onechart} is the structural-extraction split of the ChartX test set, used to evaluate \textbf{Chart2JSON} parsing on conventionally rendered bar, line, and pie charts. Models must recover the structured numerical content of each chart, evaluated by average precision at strict, slight, and high tolerance levels (AP@strict, AP@slight, AP@high).

    \item \textbf{ChartMimic~\cite{chartmimic}} is a \textbf{Chart2Code} generation benchmark that pairs 4,800 publication-style charts from scientific papers with the Python code that renders them, spanning 22 chart types and 201 subcategories. We evaluate on its Direct Mimic subset of 600 charts, reporting execution rate together with low-level scores (text, layout, type, and color) and a high-level visual-similarity score.

    \item \textbf{CoSyn-Chemical~\cite{cosyn}} is the chemical formula subset of CoSyn, a code-guided synthetic corpus in which images are rendered from programmatically generated code. Molecular structures are produced with RDKit alongside paired SMILES ground truth. We evaluate on its validation split of 128 image-text pairs, measuring optical chemical formula recognition by InChI exact-match accuracy, Tanimoto similarity, and valid SMILES rate.

    \item \textbf{ChemDraw-Bench} is an in-house business evaluation set for optical chemical formula recognition. It consists of roughly 200 challenging peptide and small-molecule SMILES sequences whose images are synthesized with the commercial software ChemDraw, probing generalization to realistic, publication-quality structures beyond synthetically rendered molecules. It adopts the same metrics as CoSyn-Chemical: InChI exact-match accuracy, Tanimoto similarity, and valid SMILES rate.

\end{itemize}

\subsubsection*{Reasoning and Generalization Tasks}

\begin{itemize}[
  leftmargin=1.0em,
  labelsep=0.5em
]
    \item \textbf{DocVQA}~\cite{docvqa} is a visual question answering benchmark over document images, comprising 39,463 training, 5,349 validation, and 5,188 test questions over 12,767 real industry-document images. We evaluate on the validation split. Answering requires jointly reading printed text and interpreting layout, tables, and figures, and is measured by average normalized Levenshtein similarity (ANLS).

    \item \textbf{InfoVQA}~\cite{infovqa} (InfographicVQA) is a visual question answering benchmark over infographics, comprising 23,946 training, 2,801 validation, and 3,288 test questions over 5,485 web-sourced images. We evaluate on the validation split. Its questions stress joint reasoning over text, graphical elements, and layout, frequently involving counting, sorting, and arithmetic, and are scored by ANLS.

    \item \textbf{AI2D}~\cite{kembhavi2016diagram} is a diagram-understanding benchmark built from over 5,000 grade-school science diagrams paired with more than 15,000 multiple-choice questions. It evaluates the ability to interpret diagrammatic elements such as labels, arrows, and icons together with their relationships, grounding language in structured visual layouts.

    \item \textbf{MathVista}~\cite{lu2023mathvista} is a benchmark for mathematical reasoning in visual contexts, aggregating 6,141 examples that pair figures, plots, and geometry diagrams with problems requiring both perception and multi-step reasoning. We evaluate on the testmini split of 1,000 examples. It stress-tests the full see-then-reason pipeline, where errors may stem from either visual extraction or symbolic computation.

    \item \textbf{MMBench}~\cite{liu2024mmbench} is a general-purpose benchmark for large vision-language models, with single-choice questions across 20 ability dimensions and parallel English and Chinese versions, split into 1,164 development and 1,784 test questions. We evaluate on the development split. It covers diverse question types, including recognition, attribute and relation reasoning, and OCR, under standardized prompts for fair cross-model comparison.

    \item \textbf{MMMU}~\cite{yue2024mmmu} is a benchmark for expert-level multimodal understanding, comprising 11.5K college-level questions across six disciplines, 30 subjects, and 30 image types such as charts, diagrams, maps, and chemical formulas. It emphasizes knowledge-intensive, reasoning-heavy problems that require integrating domain knowledge with visual evidence.

    \item \textbf{MMStar}~\cite{chen2024we} is a benchmark of 1,500 human-curated samples spanning six core capabilities and 18 detailed axes, designed to reduce visual redundancy and data leakage in existing evaluations. It offers a balanced testbed for fine-grained assessment of multimodal perception and reasoning.

    \item \textbf{OCRBench}~\cite{liu2024ocrbench} is a benchmark for OCR and text-centric visual understanding, consisting of 1,000 manually verified question-answer pairs across five tasks: text recognition, scene-text VQA, document-oriented VQA, key information extraction, and handwritten mathematical expression recognition. It assesses both accurate transcription and downstream comprehension of text in images.

\end{itemize}

\subsection{Evaluation Prompts}

To facilitate reproduction of our evaluation results, we provide the evaluation prompt for each task. By default, the \texttt{<image>} tag is prepended to the prompt text. For the reasoning and generalization tasks, we directly adopt the original prompts from the evaluation dataset.

\subsubsection*{Document Structure Tasks}

\begin{tcolorbox}[
  colback=gray!5,
  colframe=gray!60,
  title={Prompt for doc2json},
  fonttitle=\bfseries,
  breakable
]
\begin{lstlisting}[style=promptlisting, breakautoindent=false, breakindent=0pt]
- Extract layout information from the provided PDF image.
- For each layout element, output its bbox, category, and the text content within the bbox.
- Bbox format: [x1, y1, x2, y2].
- Allowed layout categories: ['header', 'title', 'text', 'figure', 'table', 'formula', 'figure_caption', 'table_caption', 'formula_caption', 'figure_footnote', 'table_footnote', 'page_footnote', 'footer'].
- Text extraction and formatting:
  1) For 'figure', the text field must be an empty string.
  2) For 'formula', format text as LaTeX.
  3) For 'table', format text as HTML.
  4) For all other categories (e.g., text, title), format text as Markdown.
- The output text must be exactly the original text from the image, with no translation or rewriting.
- Sort all layout elements in human reading order.
- Final output must be a single JSON object.
\end{lstlisting}
\end{tcolorbox}

\begin{tcolorbox}[
  colback=gray!5,
  colframe=gray!60,
  title={Prompt for doc2md},
  fonttitle=\bfseries,
  breakable
]
\begin{lstlisting}[style=promptlisting, breakautoindent=false, breakindent=0pt]
You are an AI assistant specialized in converting PDF images to Markdown format. Please follow these instructions for the conversion:

1. Text Processing:
- Accurately recognize all text content in the PDF image without guessing or inferring.
- Convert the recognized text into Markdown format.
- Maintain the original document structure, including headings, paragraphs, lists, etc.

2. Mathematical Formula Processing:
- Convert all mathematical formulas to LaTeX format.
- Enclose inline formulas with $ $. For example: This is an inline formula $E = mc^2$
- Enclose block formulas with $$ $$. For example: $$\\frac{-b \\pm \\sqrt{b^2 - 4ac}}{2a}$$

3. Table Processing:
- Convert tables to HTML format.

4. Figure Handling:
- Ignore figures content in the PDF image. Do not attempt to describe or convert images.

5. Output Format:
- Ensure the output Markdown document has a clear structure with appropriate line breaks between elements.
- For complex layouts, try to maintain the original document's structure and format as closely as possible.

Please strictly follow these guidelines to ensure accuracy and consistency in the conversion. Your task is to accurately convert the content of the PDF image into Markdown format without adding any extra explanations or comments.
\end{lstlisting}
\end{tcolorbox}

\begin{tcolorbox}[
  colback=gray!5,
  colframe=gray!60,
  title={Prompt for layout analysis},
  fonttitle=\bfseries,
  breakable
]
\begin{lstlisting}[style=promptlisting, breakautoindent=false, breakindent=0pt]
Please output the layout information from the PDF image, including each layout element's bbox and its category.
1. Bbox format: [x1, y1, x2, y2]
2. Layout Categories: ['header', 'title', 'text', 'figure', 'table', 'formula', 'figure_caption', 'table_caption', 'formula_caption', 'figure_footnote', 'table_footnote', 'page_footnote', 'footer']
3. Constraint: All layout elements must be sorted in human reading order.
4. Final Output: The entire output must be a single JSON object.
\end{lstlisting}
\end{tcolorbox}

\subsubsection*{Element-level Parsing Tasks}

\begin{tcolorbox}[
  colback=gray!5,
  colframe=gray!60,
  title={Prompt for table2html},
  fonttitle=\bfseries,
  breakable
]
\textbf{Randomly select any one of the following prompts:} \\
\begin{lstlisting}[style=promptlisting, breakautoindent=false, breakindent=0pt]
1. Please encode the table from the image into HTML format.
2. Convert the table found in the image into HTML format.
3. Render the table in the image as HTML code, please.
4. Examine the table in the provided image and generate an HTML text representation of the table.
5. Watch this table and show an HTML-style reconstructed table in text. Only preserve the basic table structure and content.
\end{lstlisting}
\end{tcolorbox}

\begin{tcolorbox}[
  colback=gray!5,
  colframe=gray!60,
  title={Prompt for table2md},
  fonttitle=\bfseries,
  breakable
]
\textbf{Randomly select any one of the following prompts:} \\
\begin{lstlisting}[style=promptlisting, breakautoindent=false, breakindent=0pt]
1. Convert the table found in the image into Markdown format.  
2. Please encode the table from the image into Markdown format.  
3. Transform the image's table into the Markdown format, please.  
4. Render the table in the image as Markdown code, please.  
5. Convert the image's table data into the Markdown structure.  
\end{lstlisting}
\end{tcolorbox}

\begin{tcolorbox}[
  colback=gray!5,
  colframe=gray!60,
  title={Prompt for formula2latex},
  fonttitle=\bfseries,
  breakable
]
\textbf{Randomly select any one of the following prompts:} \\
\begin{lstlisting}[style=promptlisting, breakautoindent=false, breakindent=0pt]
1. Convert formulas in images to LaTeX format.
2. Convert the equations found in the images into LaTeX format.
3. Transform the formulas in images into LaTeX format, please.
4. Render the formulas in images as LaTeX equations, please.
5. Please transcribe the formulas from images into LaTeX format.
\end{lstlisting}
\end{tcolorbox}

\begin{tcolorbox}[
  colback=gray!5,
  colframe=gray!60,
  title={Prompt for chart2table},
  fonttitle=\bfseries,
  breakable
]
\textbf{Randomly select any one of the following prompts:} \\
\begin{lstlisting}[style=promptlisting, breakautoindent=false, breakindent=0pt]
[Option1]
You are an expert data analyst. Your task is to extract data from the provided chart image and strictly convert it into a plain Markdown table format.

[Option2]
Observe the following strict rules:
1. Headers: Use the actual chart title, legend, or axis labels as column headers. DO NOT use generic headers like "Characteristic" or "Value".
2. Values: Extract the exact raw numerical values. DO NOT append units (like %) or years to the data cells if they are already described in the headers.
3. Format: Use standard Markdown table syntax with `|` to separate columns, including a separator row (e.g., `|---|---|`). Ensure every line starts and ends with the `|` character.
4. Output: Provide ONLY the table. Do not include any introductory text, concluding remarks, or repetitions. STOP generating immediately after the last row of the table.

### Example 1
User: <image>
please convert the image to a markdown table

Assistant:
| Entity | 1991 | 1995 | 2000 | 2005 | 2010 | 2015 | 2017 |
|---|---|---|---|---|---|---|---|
| Mongolia | 46.867000579834 | 54.5810012817383 | 53.0040016174316 | 45.6650009155273 | 33.5270004272461 | 28.4529991149902 | 30.4220008850098 |

### Current Task
User: <image>
please convert the image to a markdown table

Assistant:
\end{lstlisting}
\end{tcolorbox}

\begin{tcolorbox}[
  colback=gray!5,
  colframe=gray!60,
  title={Prompt for chart2json},
  fonttitle=\bfseries,
  breakable
]
\textbf{Randomly select any one of the following prompts:} \\
\begin{lstlisting}[style=promptlisting, breakautoindent=false, breakindent=0pt]
1. You are an expert at analyzing charts and extracting their key information into a structured JSON format.
2. Act as a chart analysis expert: analyze charts, distill key insights, and deliver them as structured JSON.
3. You're great at reading charts-grab the essentials and put them into clean, structured JSON.
4. Role: Chart analysis specialist. Responsibilities: chart parsing; key info extraction; output: structured JSON.
5. I need you to analyze charts like an expert and output the key information in structured JSON.
\end{lstlisting}
\end{tcolorbox}

\begin{tcolorbox}[
  colback=gray!5,
  colframe=gray!60,
  title={Prompt for chart2code},
  fonttitle=\bfseries,
  breakable
]
\begin{lstlisting}[style=promptlisting, breakautoindent=false, breakindent=0pt]
You are an expert Python developer who specializes in writing matplotlib code based on a given picture. I need your help to generate the Python code that can reproduce the picture based on the picture I provide. Now, please give me the matplotlib code that reproduces the picture.
\end{lstlisting}
\end{tcolorbox}

\begin{tcolorbox}[
  colback=gray!5,
  colframe=gray!60,
  title={Prompt for chem2smiles},
  fonttitle=\bfseries,
  breakable
]
\textbf{Randomly select any one of the following prompts:} \\
\begin{lstlisting}[style=promptlisting, breakautoindent=false, breakindent=0pt]
1. Please convert this chemical structure to its SMILES string representation.
2. Convert the chemical structure shown to standard SMILES notation.
3. Generate the SMILES string for the chemical structure displayed in the image.
4. What is the SMILES representation of the chemical compound shown?
5. Convert all chemical diagrams in the image to their SMILES format and separate them with newlines.
\end{lstlisting}
\end{tcolorbox}

\subsection{Financial Information Extraction: Task Definition and Pipeline}
\label{appendix:fie}

This appendix details the Financial Information Extraction (FinIE) task used as a downstream evaluation of Infinity-Parser2 (Table~\ref{tab:finie}). Annual reports, drawn primarily from the stock market in our benchmark, represent a challenging domain for document intelligence: a single report often spans several hundred pages and couples frameless, multi-column layouts with a high density of tabular data. Our task targets the three primary statements, the balance sheet, the income statement, and the cash flow statement, which the pipeline must first locate and then transcribe in full with high structural fidelity.

\subsubsection*{Task Definition}
The objective of FinIE is to extract the three primary financial statements, the balance sheet, the income statement, and the cash flow statement, in their entirety from an unstructured PDF report. Rather than querying individual indicators, the task transcribes every line item of each primary statement in a single pass. Note references are recorded verbatim but not expanded, so the detailed note tables are left untouched. We formulate it as a structured extraction task with the following inputs and outputs.

\begin{tcolorbox}[
  colback=gray!5,
  colframe=gray!60,
  title={Definition of FinIE},
  fonttitle=\bfseries,
  breakable
]
\textbf{Input:} \\
\begin{enumerate}[
  leftmargin=1.0em,
  labelsep=0.5em
]
\item PDF Document $\mathcal{D}$.
\item Extraction directive $\mathcal{I}$: extract the balance sheet, the income statement, and the cash flow statement, without expanding note references.
\end{enumerate}

\medskip
\textbf{Output:} A JSON object with one entry per primary statement. Each entry carries the statement name, its reporting \texttt{periods} (given as \texttt{cutoff\_date}s) and their \texttt{period\_type}, the reporting \texttt{currency} and \texttt{unit}, and an array of line-item records. Each record holds the source \texttt{page}, the original line-item label, the amounts for the current and prior periods (\texttt{amount\_cy}, \texttt{amount\_py}), and the unexpanded \texttt{notes} reference.
\end{tcolorbox}

\subsubsection*{Parse-Locate-Extract Pipeline}
Processing an entire annual report at once is infeasible, given context-window limits and the noise introduced by hundreds of irrelevant pages. We therefore adopt a modular pipeline with three stages: full-document parsing, three-statement page localization, and LLM-based statement extraction with intra-statement validation. To isolate the contribution of document parsing, we fix the localization and extraction modules and vary only the parser, so that any change in end-to-end accuracy is attributable to parsing quality.

\begin{itemize}[
  leftmargin=1.0em,
  labelsep=0.5em
]
    \item \textbf{Stage 1: Full-Document Parsing.} Each page of the raw PDF is converted into structured text by the candidate parser, namely Infinity-Parser2-Flash*, MinerU2.5~\cite{niu2025mineru2}, PaddleOCR-VL-1.5~\cite{cui2025paddleocr}, or DeepSeek-OCR-2~\cite{wei2026deepseek2}. Tables are serialized to HTML and running text to Markdown, while element coordinates and reading order are preserved. This stage is decisive: a single merged cell or a broken reading order corrupts the data irrecoverably and propagates to every later stage.

    \item \textbf{Stage 2: Three-Statement Page Localization.} Because the target line items reside almost entirely in the balance sheet, the income statement, and the cash flow statement, we locate the pages of these three statements rather than scan the whole report. We combine complementary localization signals, principally keyword search and RAG-based semantic search. Keyword search matches canonical statement headings and their bilingual variants (for example, ``Consolidated Balance Sheet'' or its Chinese equivalent) against the parsed text to nominate candidate pages. RAG-based semantic search then embeds each parsed page and the canonical descriptions of the three statements with Inf-Retriever-v1~\cite{infly-ai_2025} and ranks pages by semantic similarity, recovering statements whose headings are abbreviated or missing. The candidate sets are merged and deduplicated to yield the located statement pages, whose parsed content forms the context $\mathcal{C}$.

    \item \textbf{Stage 3: LLM-based Statement Extraction and Validation.} The context $\mathcal{C}$ is passed to DeepSeek-V4-Flash~\cite{deepseekai2026deepseekv4}, a large language model prompted to transcribe every line item of the three primary statements into the structured JSON defined above. Note references are copied verbatim without expansion, and consistency checks are performed within each primary statement, verifying that subtotals reconcile with their component rows and that the balance-sheet identity holds, with total assets equal to the sum of total liabilities, mezzanine equity, and total equity. Operating on the localized statement pages rather than the full report keeps the input within the context window and suppresses spurious matches from unrelated sections.
\end{itemize}

\subsubsection*{Example Output}
Applying the pipeline to a representative annual report (amounts in RMB millions) yields, for each primary statement, an array of line-item records. An abridged excerpt appears below. The two amount columns correspond to the current and prior periods declared in \texttt{periods}, and the \texttt{notes} reference is retained verbatim without expansion.

\begin{tcolorbox}[
  colback=gray!5,
  colframe=gray!60,
  title={Example FinIE Output},
  fonttitle=\bfseries,
  breakable
]
\begin{lstlisting}[basicstyle=\ttfamily\small, breaklines=true]
{
  "balance_sheet": {
    "statement": "Consolidated Balance Sheet",
    "period_type": "as_of",
    "periods": ["20240331", "20230331"],
    "currency": "RMB",
    "unit": "million",
    "line_items": [
      {"page": 256, "line_item": "Cash and cash equivalents", "amount_cy": "248,125", "amount_py": "193,086", "notes": "Note 2(p)"},
      {"page": 256, "line_item": "Total current assets", "amount_cy": "752,864", "amount_py": "697,966", "notes": null},
      {"page": 256, "line_item": "Total assets", "amount_cy": "1,764,829", "amount_py": "1,753,044", "notes": null},
      {"page": 257, "line_item": "Total liabilities, mezzanine equity and equity", "amount_cy": "1,764,829", "amount_py": "1,753,044", "notes": null}
    ]
  },
  "income_statement": {
    "statement": "Consolidated Income Statement",
    "period_type": "year_ended",
    "periods": ["20240331", "20230331"],
    "currency": "RMB",
    "unit": "million",
    "line_items": [
      {"page": 254, "line_item": "Revenue", "amount_cy": "941,168", "amount_py": "868,687", "notes": "Note 5, 22"},
      {"page": 254, "line_item": "Income from operations", "amount_cy": "113,350", "amount_py": "100,351", "notes": null},
      {"page": 254, "line_item": "Net income", "amount_cy": "71,332", "amount_py": "65,573", "notes": null}
    ]
  },
  "cash_flow_statement": {
    "statement": "Consolidated Cash Flow Statement",
    "period_type": "year_ended",
    "periods": ["20240331", "20230331"],
    "currency": "RMB",
    "unit": "million",
    "line_items": [
      {"page": 261, "line_item": "Net cash provided by operating activities", "amount_cy": "182,593", "amount_py": "199,752", "notes": null},
      {"page": 263, "line_item": "Cash and cash equivalents, restricted cash and escrow receivables at end of year", "amount_cy": "286,424", "amount_py": "229,510", "notes": null}
    ]
  }
}
\end{lstlisting}
\end{tcolorbox}

\subsection{Visualization Results}

\begin{figure*}[htbp]
    \centering
    \includegraphics[width=\textwidth]{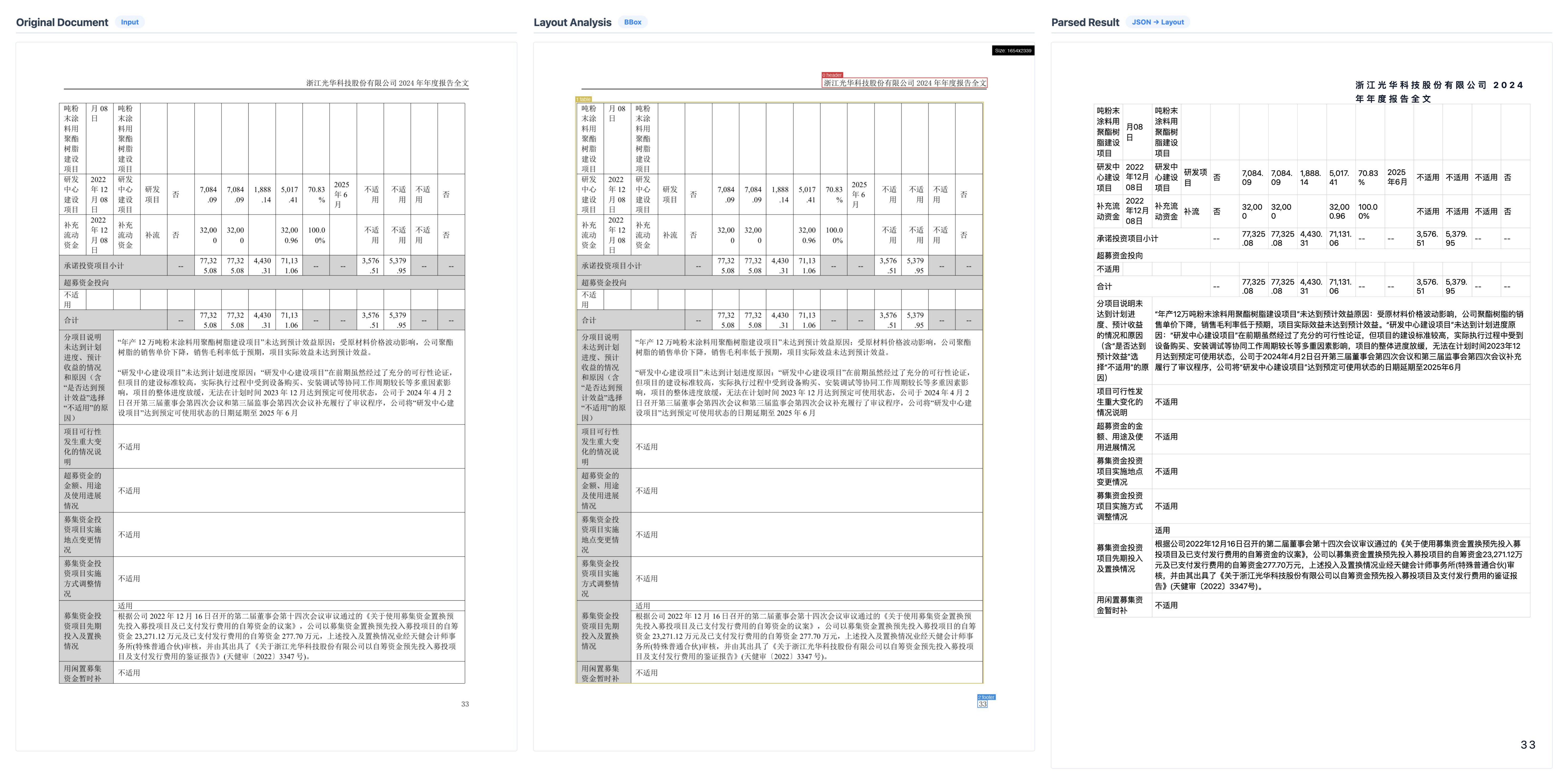}
    \caption{Visualization results of the A-stock image. The challenge is accurately determining column spans (colspans) in tables to prevent miscounting.}
    \label{fig:visualization_results_a_stock}
\end{figure*}

\begin{figure*}[htbp]
    \centering
    \includegraphics[width=\textwidth]{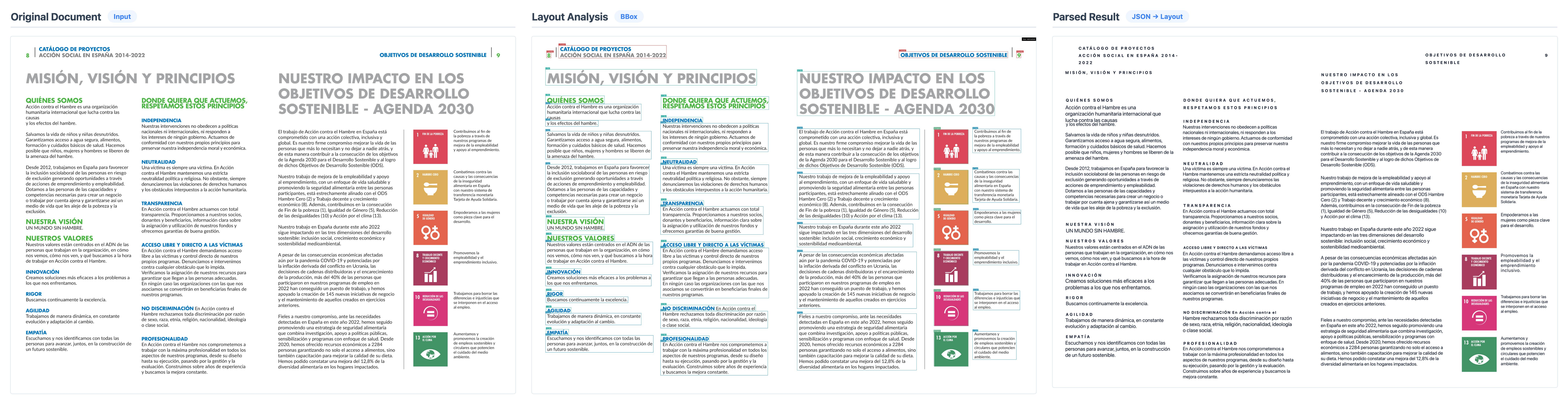}
    \caption{Visualization results of the multi-column image. The difficulty lies in executing complex layout analysis and accurately recovering the reading order.}
    \label{fig:visualization_results_muti_column}
\end{figure*}

\begin{figure*}[htbp]
    \centering
    \includegraphics[width=\textwidth]{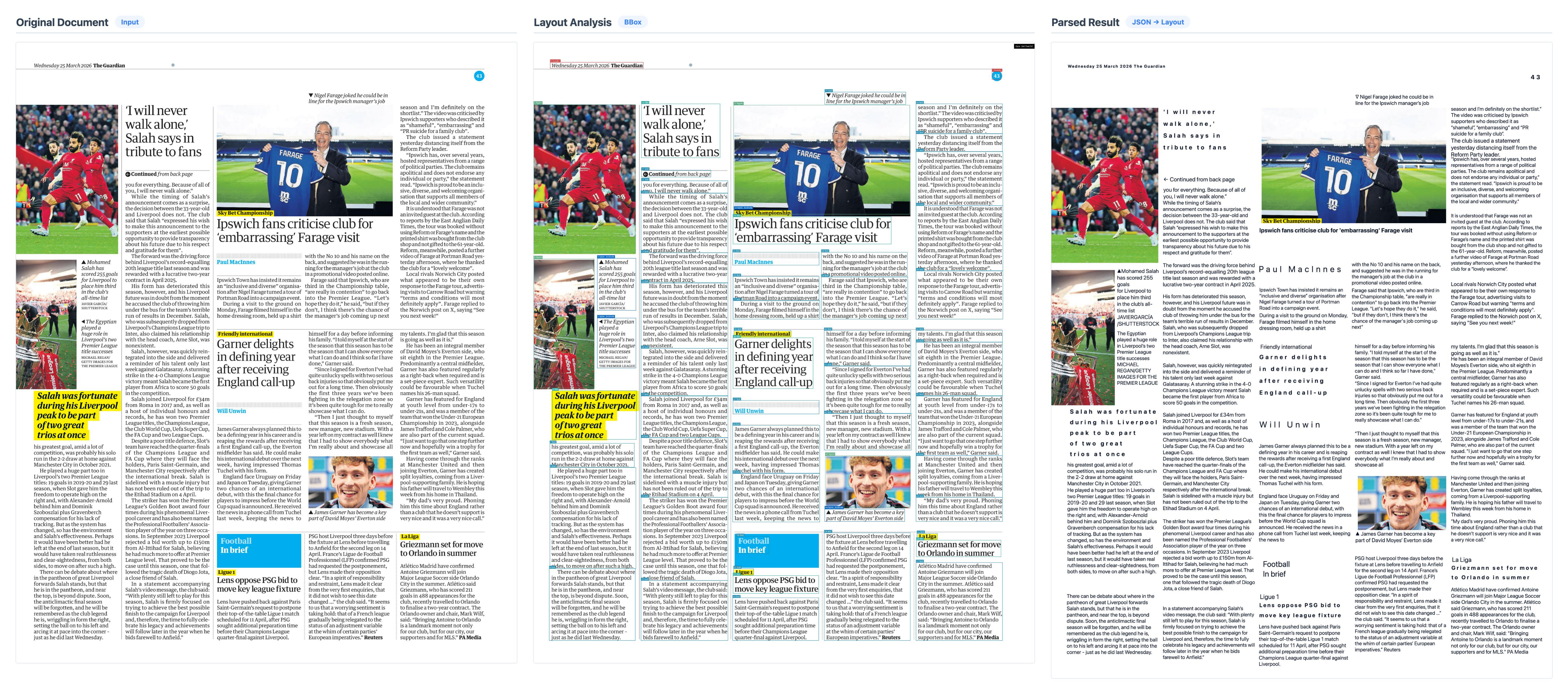}
    \caption{Visualization results of the newspaper image. The primary challenge is avoiding bounding box omissions caused by ultra-dense text distribution, narrow column margins, and microscopic fonts.}
    \label{fig:visualization_results_newspaper_1}
\end{figure*}

\begin{figure*}[htbp]
    \centering
    \includegraphics[width=\textwidth]{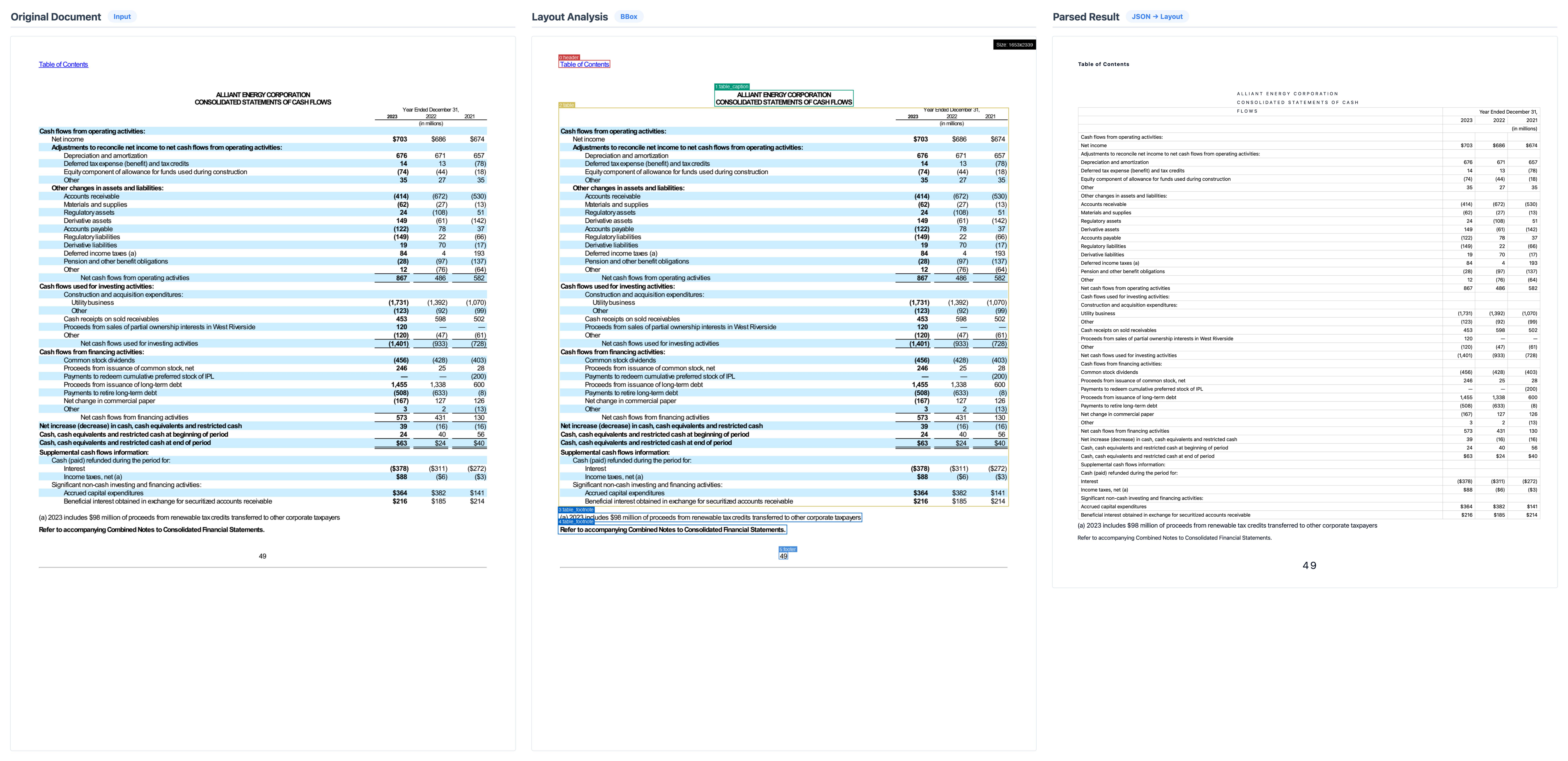}
    \caption{Visualization results of the US-stock image. The difficulty involves achieving precise row alignment across wide frameless spaces and capturing the hierarchical semantics of indented headers.}
    \label{fig:visualization_results_us_stock}
\end{figure*}

\begin{figure*}[htbp]
    \centering
    \includegraphics[width=\textwidth]{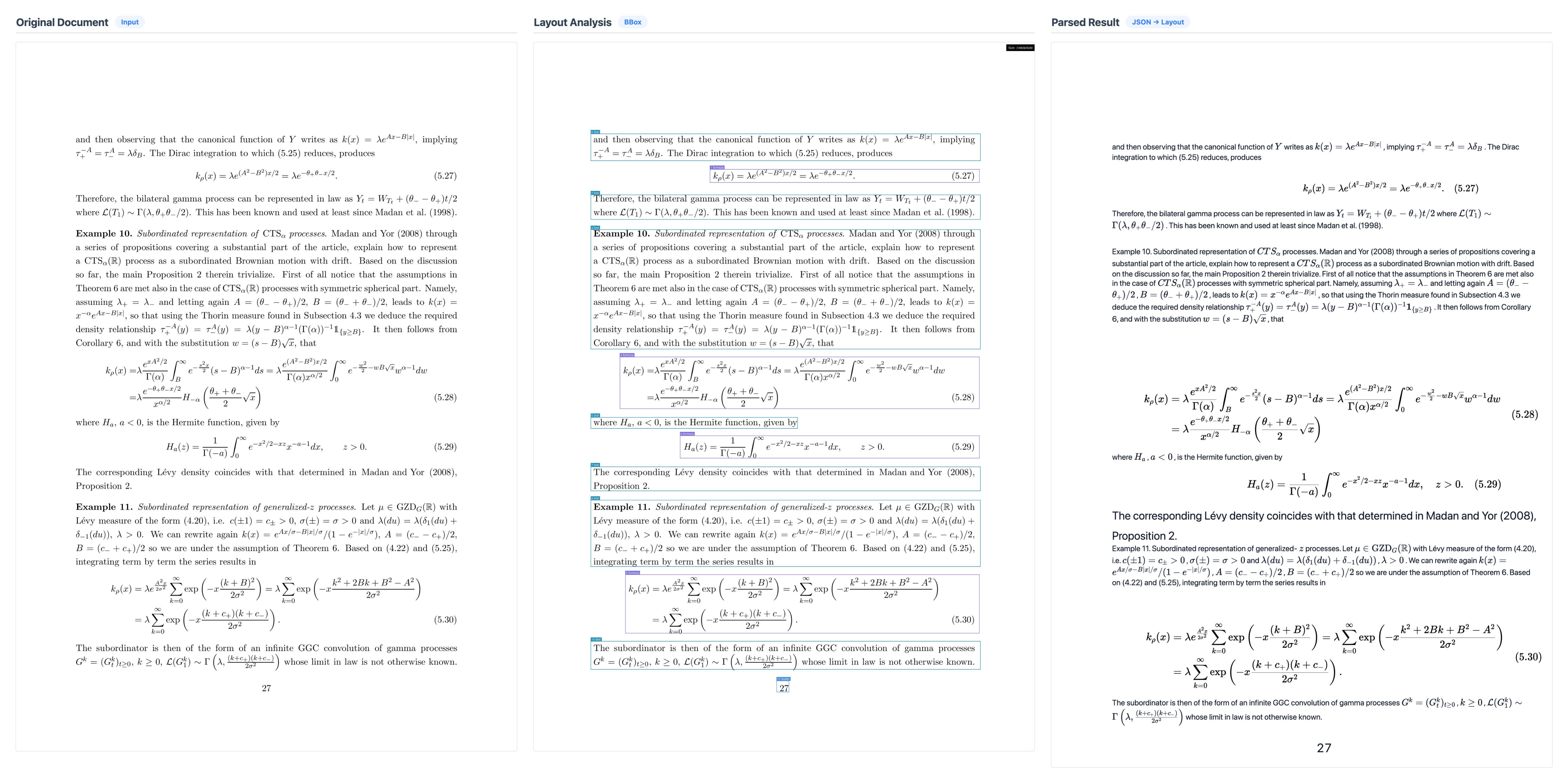}
    \caption{Visualization results of the arXiv paper image. The challenge is accurately preserving the structure of complex multi-line mathematical formulas, dense inline notations, and deeply nested subscripts/superscripts.}
    \label{fig:visualization_results_arxiv}
\end{figure*}

\begin{figure*}[htbp]
    \centering
    \includegraphics[width=\textwidth]{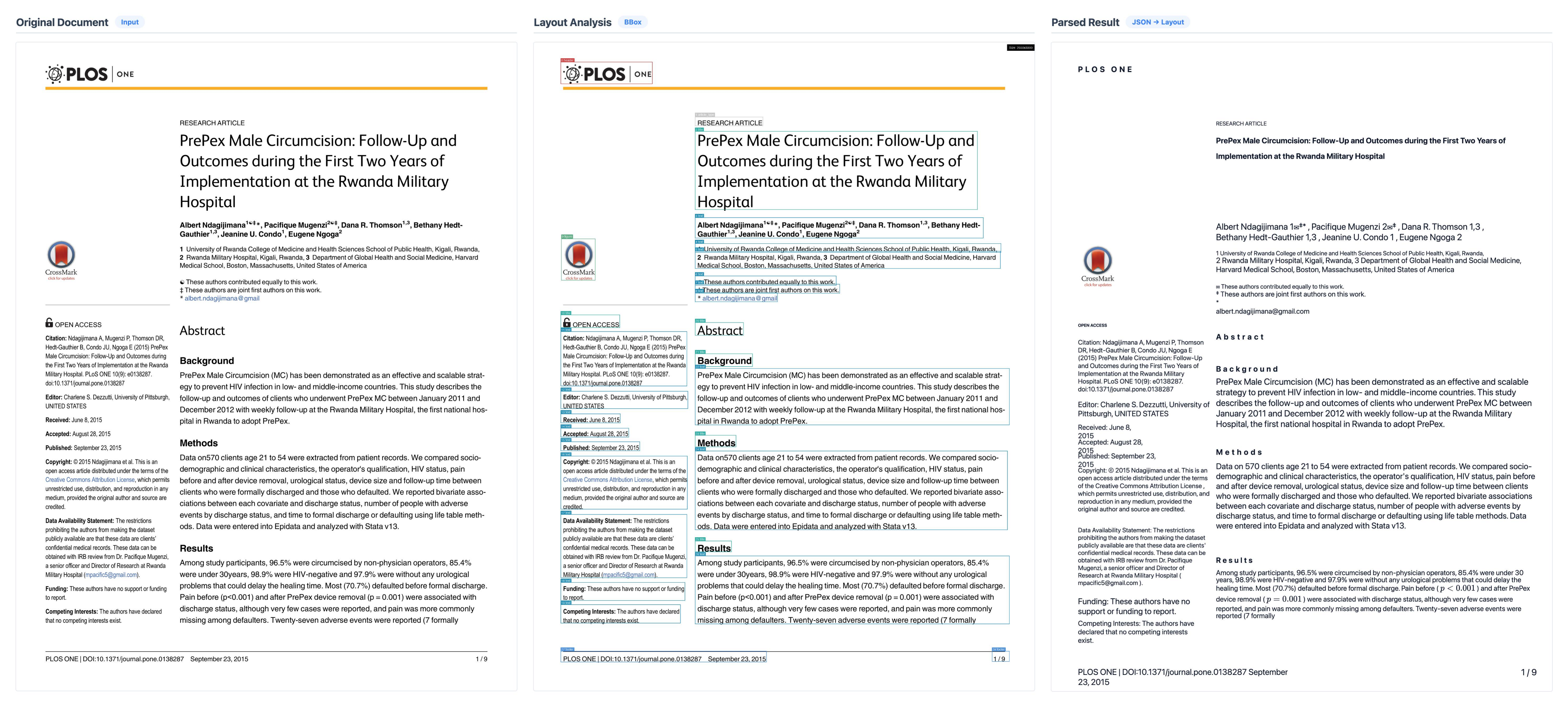}
    \caption{Visualization results of the magazine image. The main difficulty is recovering the correct reading order within an asymmetric multi-column layout.}
    \label{fig:visualization_results_magazine}
\end{figure*}

\begin{figure*}[htbp]
    \centering
    \includegraphics[width=\textwidth]{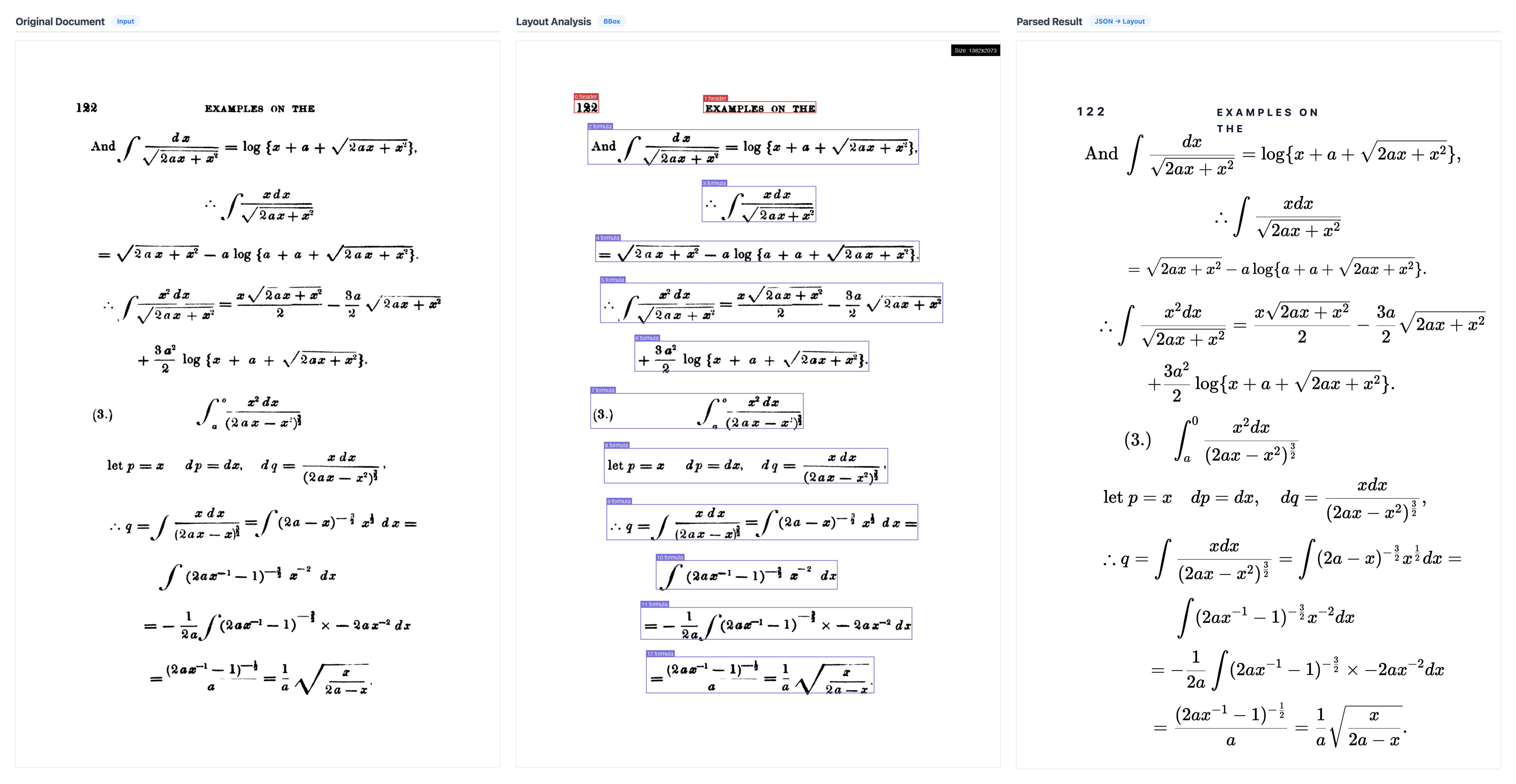}
    \caption{Visualization results of the old scanned math image. The challenge lies in recognizing text and symbols from severely degraded and blurred print.}
    \label{fig:visualization_results_old_scan_math}
\end{figure*}

\begin{figure*}[htbp]
    \centering
    \includegraphics[width=\textwidth]{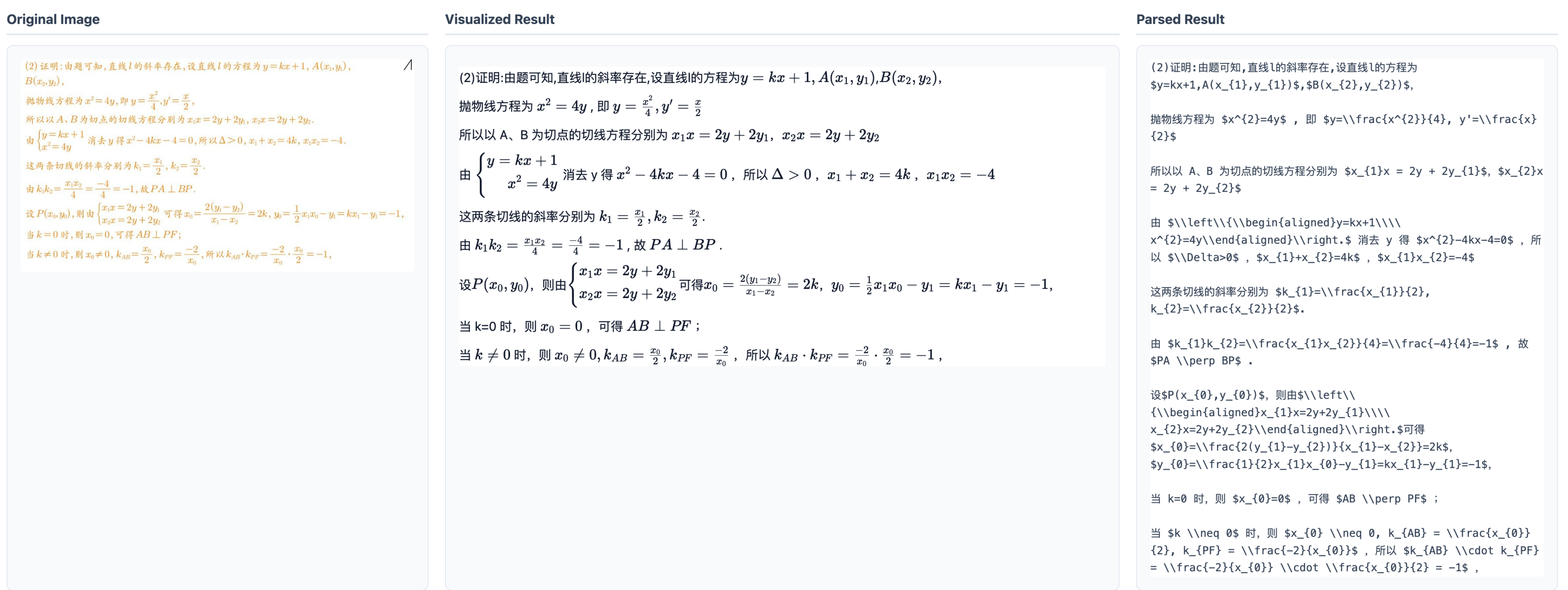}
    \caption{Visualization results of the text parsing task. The challenge lies in faithfully transcribing densely written text interleaved with inline and displayed mathematical expressions while preserving the correct reading order across the full page.}
    \label{fig:visualization_results_text_parsing}
\end{figure*}

\begin{figure*}[htbp]
    \centering
    \includegraphics[width=\textwidth]{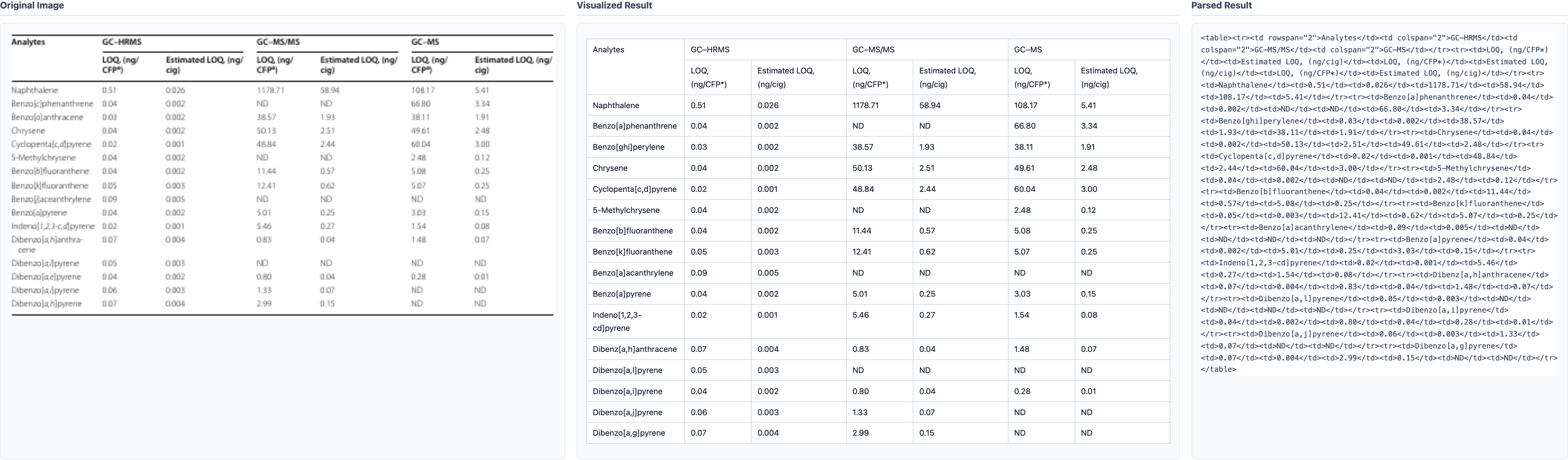}
    \caption{Visualization results of the table parsing task. The difficulty involves reconstructing multi-level grouped column headers and aligning numerous numeric cells precisely within a dense scientific table.}
    \label{fig:visualization_results_table_parsing}
\end{figure*}

\begin{figure*}[htbp]
    \centering
    \includegraphics[width=\textwidth]{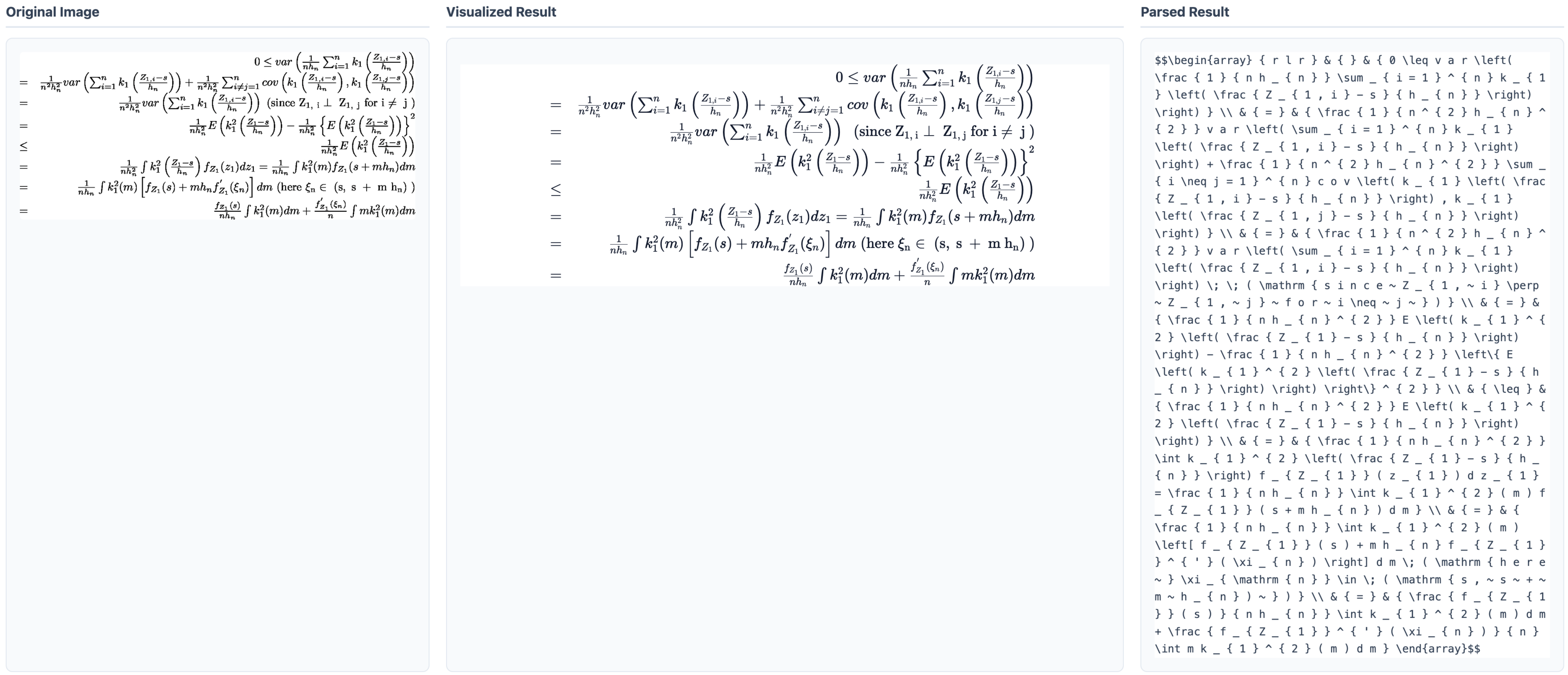}
    \caption{Visualization results of the formula parsing task. The challenge is accurately transcribing a multi-line derivation with nested fractions, summation and expectation operators, integrals, and densely stacked subscripts and superscripts into valid LaTeX.}
    \label{fig:visualization_results_formula_parsing}
\end{figure*}

\begin{figure*}[htbp]
    \centering
    \includegraphics[width=\textwidth]{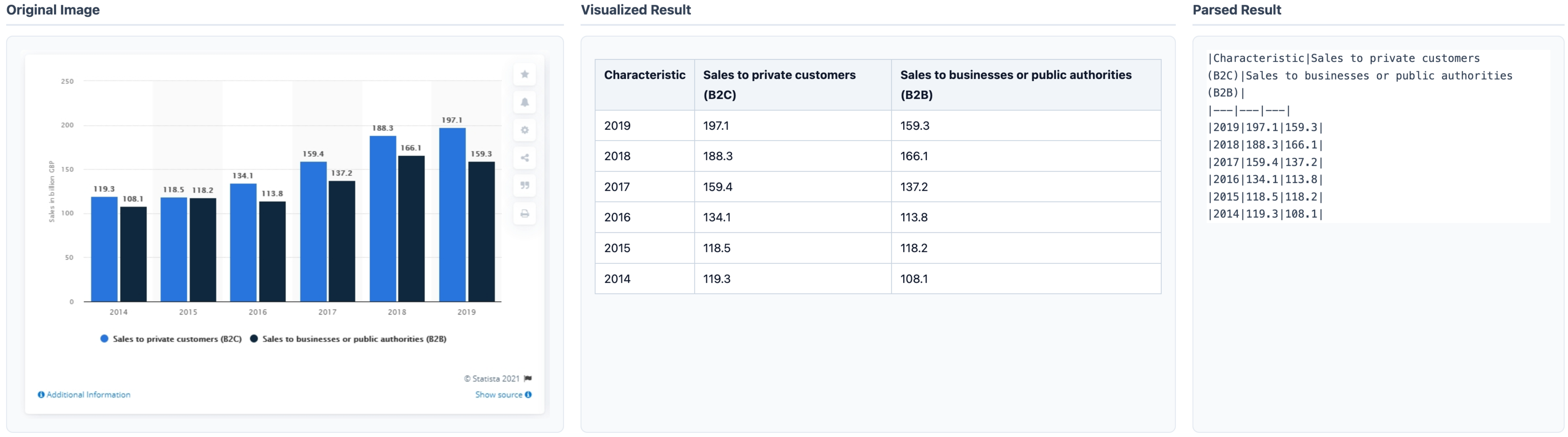}
    \caption{Visualization results of the Chart-to-Table task. The difficulty lies in reading the numeric value of each grouped bar and associating it with the correct category and data series to recover a structured table.}
    \label{fig:visualization_results_chart2table}
\end{figure*}

\begin{figure*}[htbp]
    \centering
    \includegraphics[width=\textwidth]{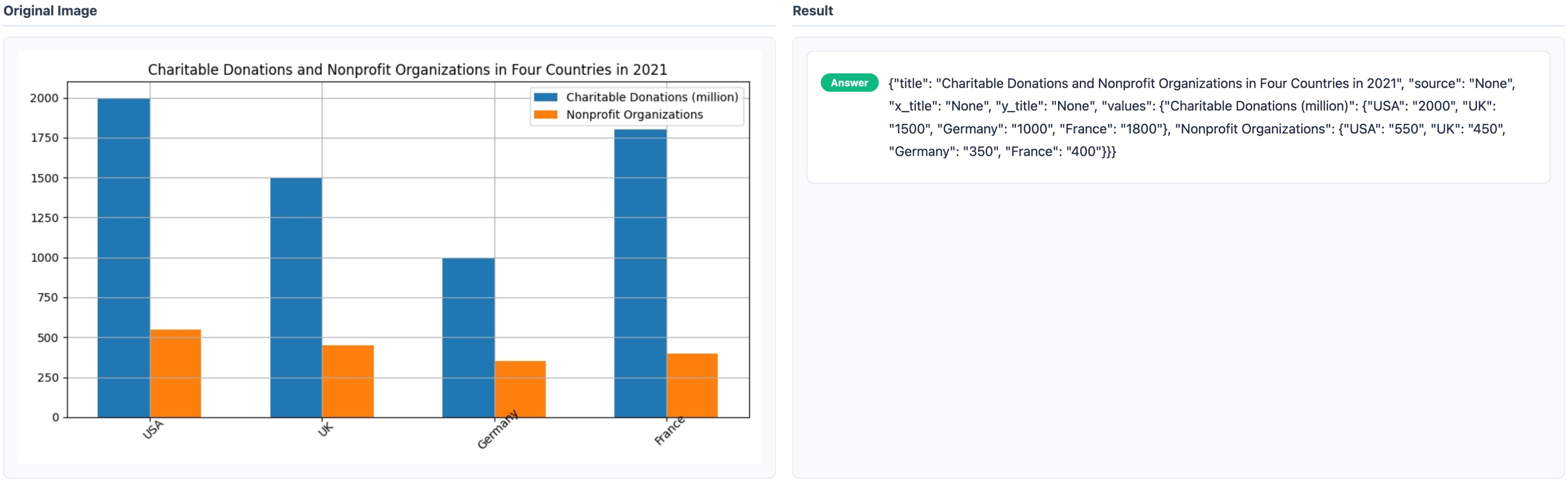}
    \caption{Visualization results of the Chart-to-JSON task. The challenge is extracting the title, axis labels, and hierarchically grouped data values into a well-formed, schema-consistent JSON representation.}
    \label{fig:visualization_results_chart2json}
\end{figure*}

\begin{figure*}[htbp]
    \centering
    \includegraphics[width=\textwidth]{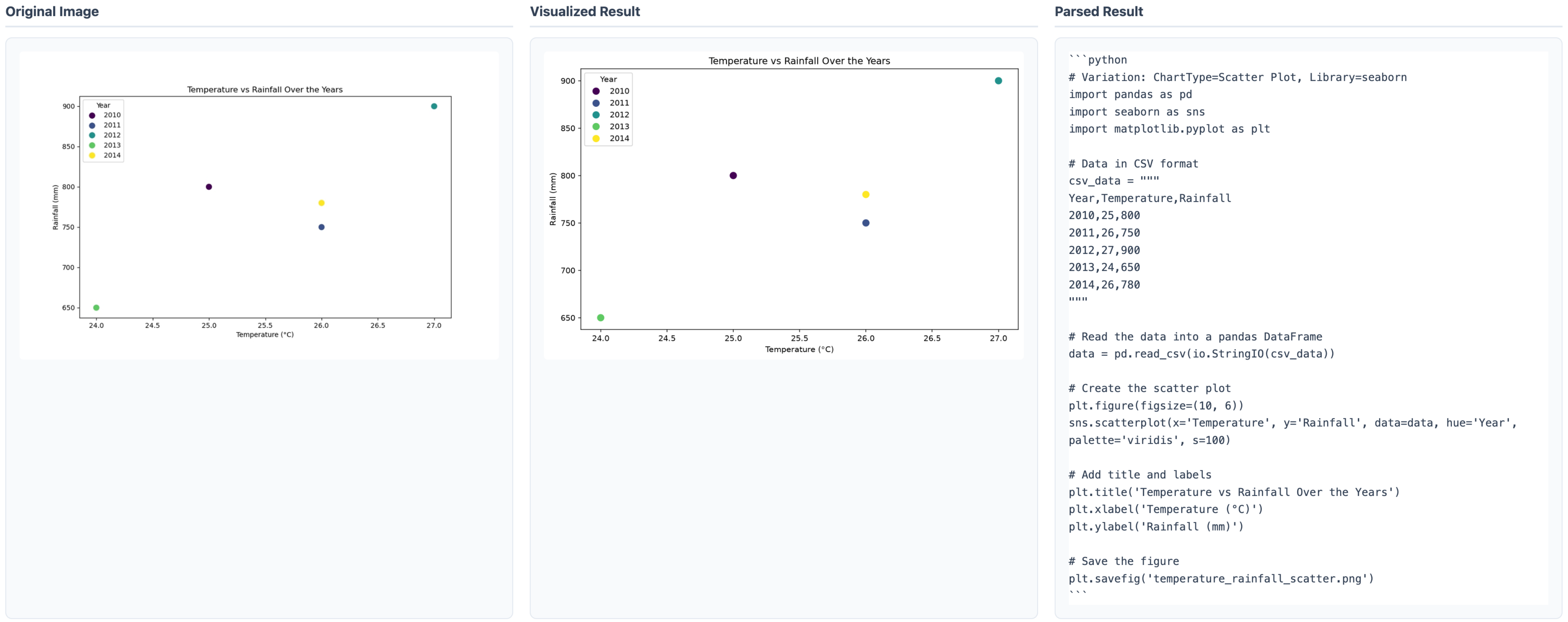}
    \caption{Visualization results of the Chart-to-Code task. The difficulty involves inferring the chart type, underlying data, and visual styling to generate executable plotting code that faithfully reproduces the original figure.}
    \label{fig:visualization_results_chart2code}
\end{figure*}

\begin{figure*}[htbp]
    \centering
    \includegraphics[width=\textwidth]{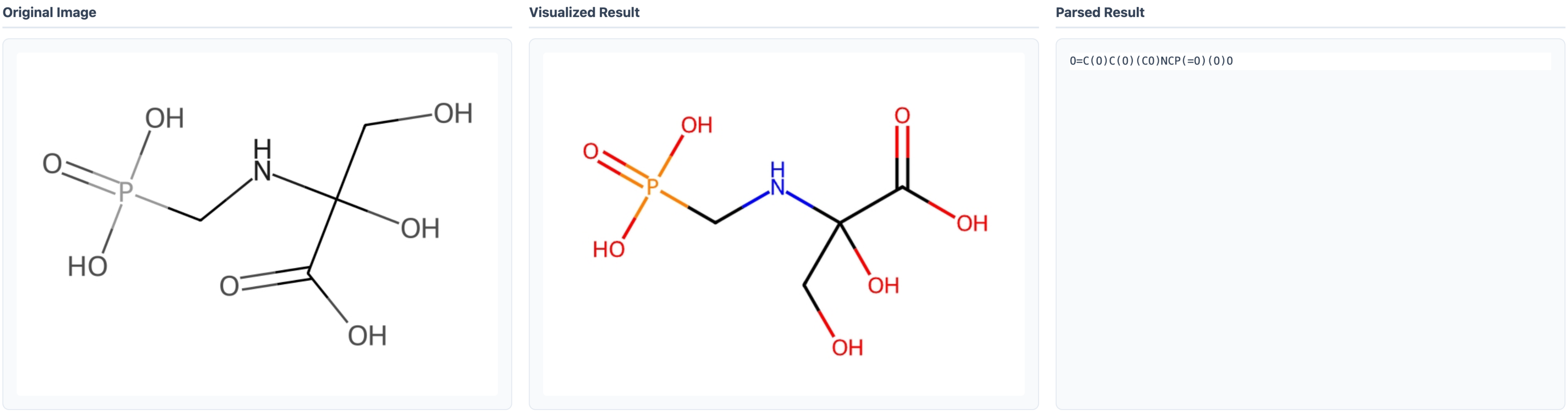}
    \caption{Visualization results of the Chemical-to-SMILES task. The challenge is recognizing atoms, bonds, and functional groups in the molecular diagram and converting the two-dimensional structure into a valid linear SMILES string.}
    \label{fig:visualization_results_chem2smiles}
\end{figure*}

\begin{figure*}[htbp]
    \centering
    \includegraphics[width=\textwidth]{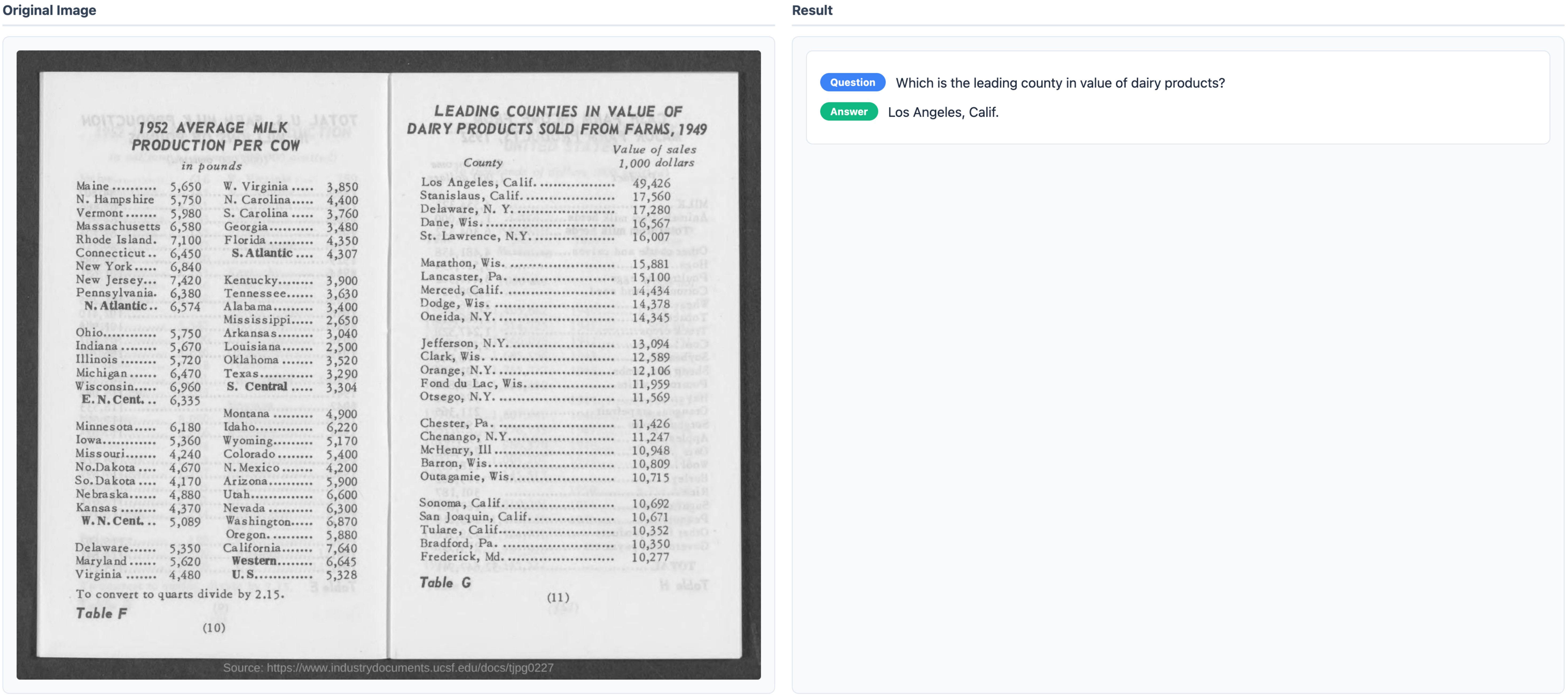}
    \caption{Visualization results of the DocVQA task. The challenge lies in locating the relevant table within a dense multi-table document page and reading the top-ranked entry to answer the question.}
    \label{fig:visualization_results_docvqa}
\end{figure*}

\begin{figure*}[htbp]
    \centering
    \includegraphics[width=\textwidth]{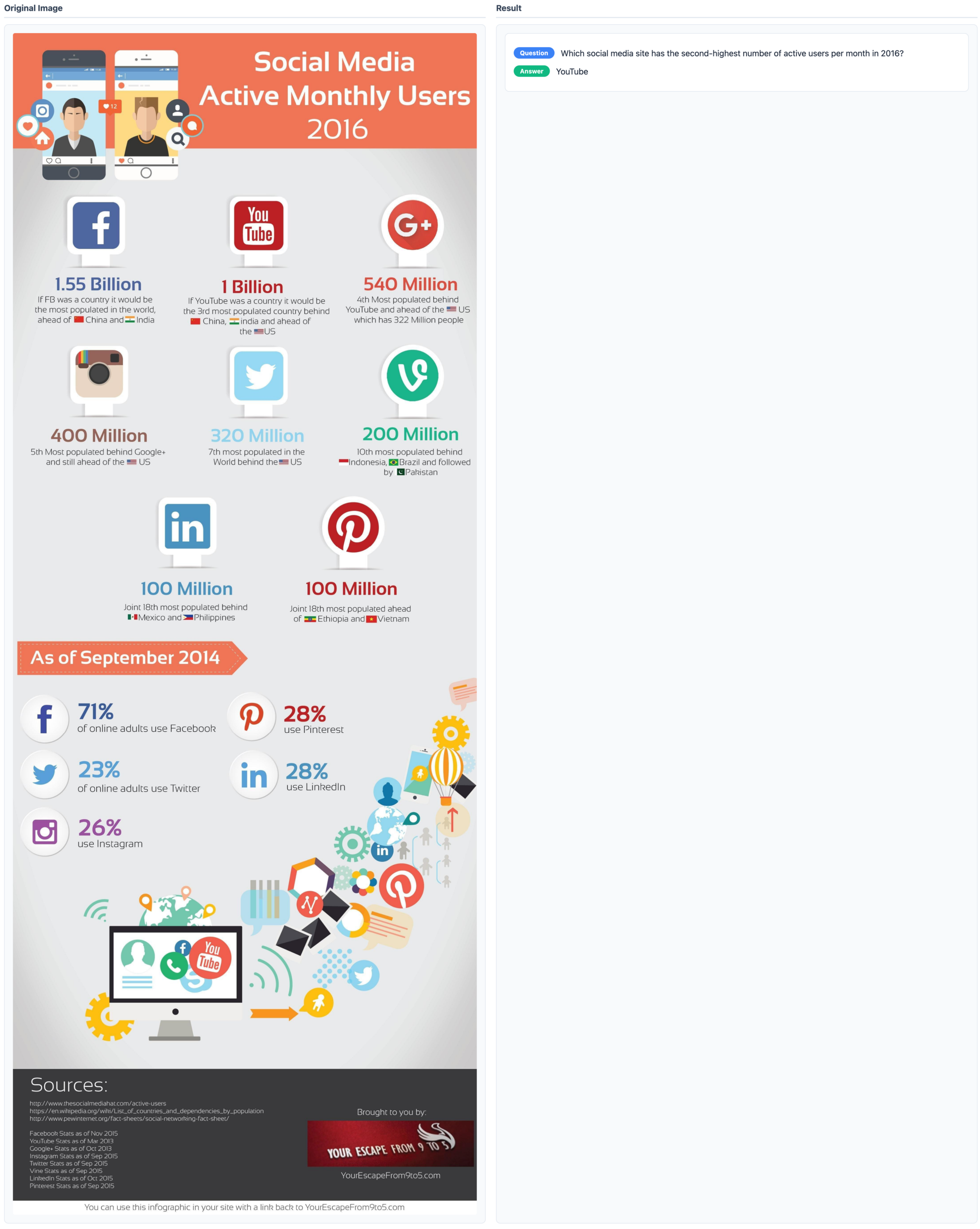}
    \caption{Visualization results of the InfoVQA task. The difficulty lies in reasoning over a visually complex infographic with scattered icons, stylized numbers, and a non-linear layout to compare and rank quantities.}
    \label{fig:visualization_results_infovqa}
\end{figure*}

\begin{figure*}[htbp]
    \centering
    \includegraphics[width=\textwidth]{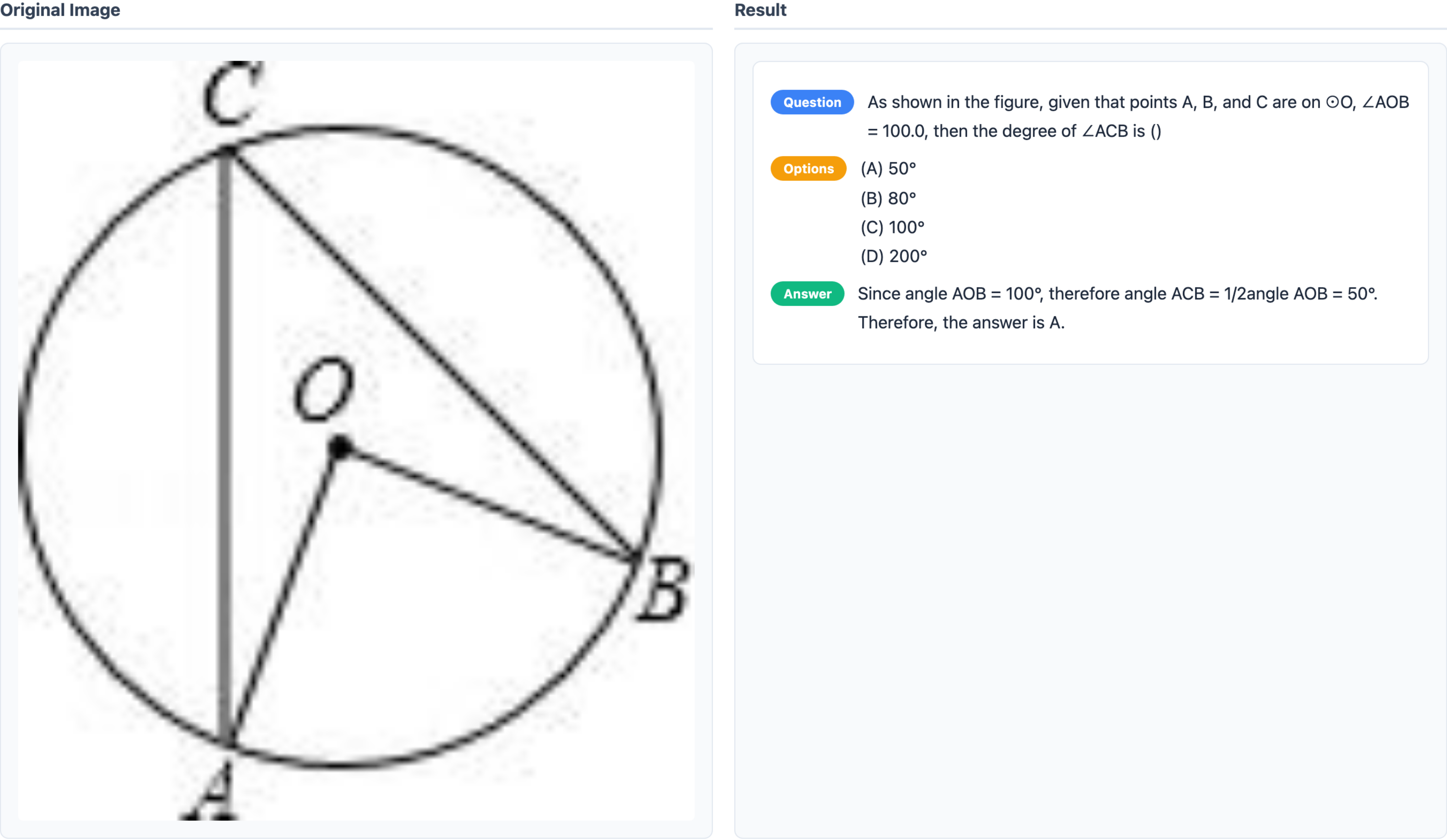}
    \caption{Visualization results of the MathVista task. The challenge is jointly interpreting the geometric figure and applying mathematical reasoning to derive the correct answer.}
    \label{fig:visualization_results_mathvista}
\end{figure*}

\begin{figure*}[htbp]
    \centering
    \includegraphics[width=\textwidth]{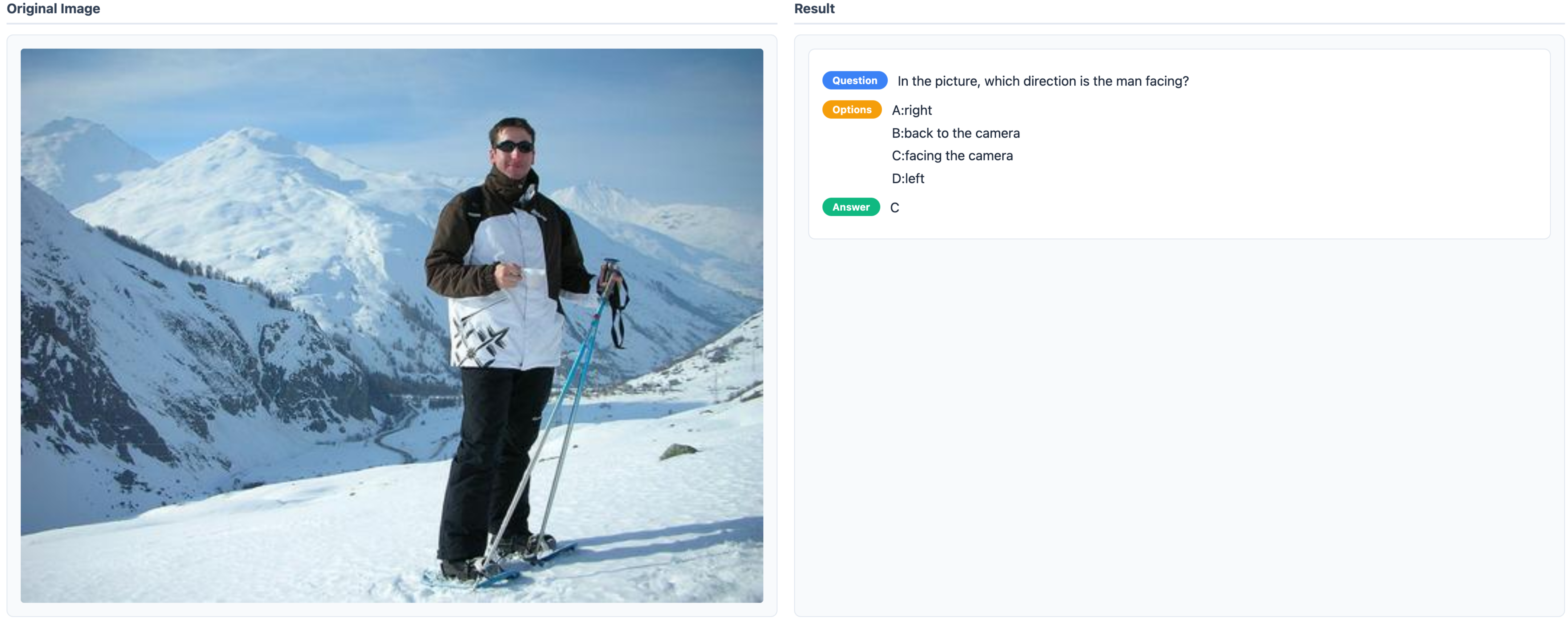}
    \caption{Visualization results of the MMBench task. The difficulty lies in fine-grained spatial perception and commonsense reasoning to determine the subject's orientation within a natural scene.}
    \label{fig:visualization_results_mmbench}
\end{figure*}

\end{document}